\documentclass{article} 
\usepackage{iclr2027_conference,times}

\usepackage{amsmath,amsfonts,bm}

\def\eqref#1{equation~\ref{#1}}

\def\1{\bm{1}}

\DeclareMathAlphabet{\mathsfit}{\encodingdefault}{\sfdefault}{m}{sl}
\SetMathAlphabet{\mathsfit}{bold}{\encodingdefault}{\sfdefault}{bx}{n}

\usepackage[utf8]{inputenc} 
\usepackage[T1]{fontenc}    
\usepackage{hyperref}       
\usepackage{url}            
\usepackage{booktabs}       
\usepackage{amsfonts}       
\usepackage{nicefrac}       
\usepackage{microtype}      
\usepackage{xcolor}         
\usepackage{graphicx}
\usepackage{colortbl}
\usepackage{makecell}
\usepackage{multirow}
\usepackage{xspace}
\usepackage{listings}
\usepackage{amsmath}
\usepackage{cleveref}

\definecolor{pastelblue}{HTML}{EEF6F2} 
\definecolor{tablebanner}{HTML}{ECE9F2} 
\definecolor{preferredcell}{gray}{0.95}
\definecolor{promptbg}{HTML}{EEE9F8} 
\definecolor{promptrule}{HTML}{B5A6D4}
\usepackage{pifont}

\newcommand{\benchmark}{\texttt{FindIt\xspace}}

\title{FindIt: A Format-Informed Visual Detection Benchmark for Generalist Multimodal LLMs}

\author{Eshika Khandelwal\thanks{Correspondance: \texttt{eshika.khandelwal@uni-tuebingen.de}} \\ Tuebingen AI Center\\ University of Tuebingen  \And Jingjing Pan \\ Woven by Toyota, Inc. \\ Tokyo, Japan \And Mingfang Zhang \\ Woven by Toyota, Inc. \\ Tokyo, Japan  \And  Quan Kong \\ Woven by Toyota, Inc. \\ Tokyo, Japan \And Lorenzo Garattoni \\ Toyota Motor Europe \\ Brussels, Belgium \And Hilde Kuehne \\ Tuebingen AI Center\\ University of Tuebingen}

\iclrfinalcopy 

\begin{document}
\maketitle

\begin{abstract}
    Multimodal large language models (MLLMs) are predominantly evaluated on free-form vision-language tasks such as visual question answering, captioning, and summarization. 
    However, their practical use is rapidly expanding to more structured computer vision settings, where users prompt models to perform localization-centric tasks such as object detection, often within larger agentic or decision-making systems. 
    Despite this shift, there is currently no standardized benchmark that systematically evaluates these capabilities at scale.
    In this work, we introduce the first comprehensive benchmark specifically designed to assess the promptable localization abilities of generalist MLLMs. 
    Our benchmark spans four core task categories: object detection, referring expression detection, instance-level detection, and video-based detection. 
    To enable consistent and fair evaluation, we develop a unified framework that standardizes inputs, enforces parsable bounding box outputs, and defines transparent evaluation protocols across tasks.
    Using this suite, we evaluate a diverse set of open-source and proprietary MLLMs, providing an in-depth analysis of their performance and limitations.
    Beyond accuracy, we examine models' ability to adhere to output format specifications, showing that current systems are highly sensitive to formatting constraints and often fail to generalize even to minor variations.
    Our results highlight both the strengths and shortcomings of state-of-the-art MLLMs in localization settings, and point toward important directions for improving multimodal model design and evaluation. \footnote{Code available at: \href{https://github.com/esh04/FindIt}{https://github.com/esh04/FindIt}}
    \end{abstract}

\section{Introduction}

Generalist multimodal large language models (MLLMs) have evolved from generative systems for free-form text output to versatile tools capable of a wide range of tasks. A particularly important use case is object localization as users increasingly prompt these models to produce spatial outputs such as bounding boxes, e.g. for agentic or VLA systems~\citep{zhang2026vlm4vlarevisitingvisionlanguagemodelsvisionlanguageaction} or reasoning~\citep{Liao2025Can}. Despite this growing practical relevance, current evaluation protocols have yet to address localization as a benchmark category, leaving an important aspect of real-world performance underexplored and making it difficult to compare model performance for scenarios that require localization.

Localization tasks such as object detection are a long-lasting subject in computer vision which led to specialized methods and model series such as 
YOLO~\citep{sapkota2026ultralyticsyoloevolutionoverview} and DETR~\citep{Carion2020DETR, Robinson2026rf-detr,huang2026ledetrrevisitingrealtimedetection,huang2024deim}. Those task-specific models are designed to localize objects via learned visual representations, relying on architectures such as convolutional neural networks or vision transformers, and produce a bounding box and class label via specialized detection heads. As a result, they allow for high-precision detection within the predefined scope but have limited flexibility beyond it.
In contrast, generalist MLLMs achieve localization via model architectures where vision and text features are usually jointly processed by a fine-tuned language model, generating bounding boxes as part of the LLM output based on a prompt that specifies the task. 
The spatial prediction is therefore strongly guided by the model's internal representations as localization is not enforced by an explicit pixel-wise objective but rather inferred through alignment between vision and language. 
While this can diminish performance, it gives these models much higher flexibility with respect to various tasks, open-world detection, and flexible output formats (e.g., coordinate tuples embedded in text), practically enabling a single model to handle multiple tasks without retraining. But, such generalist MLLM localization also raises new challenges for evaluation as the output is no longer based on algorithmic constraints.

\begin{figure}
    \centering
\includegraphics[width=\textwidth]{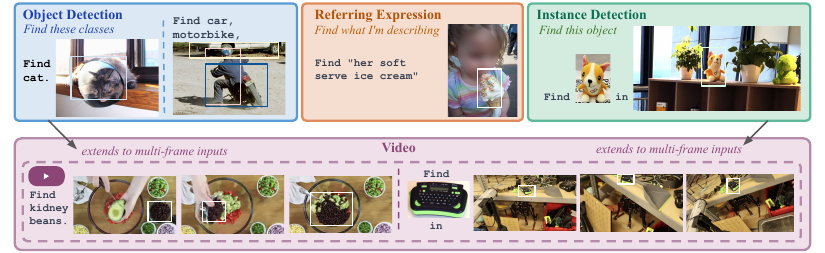}
    \vspace{-2mm}
    \caption{Overview of the four tasks in \benchmark{}: (1) object detection, (2) referring expression detection, (3) instance localization, and (4) video object detection. Video object detection contains both object detection and instance localization datasets and extends those two to multi-frame inputs.
    }    
    \label{fig:tasks}
    \vspace{-4mm}
\end{figure}

This work addresses these challenges by introducing a first benchmark specifically designed to evaluate localization abilities in generalist MLLMs. 
Namely, it focuses on generalist MLLMs that can follow free-form prompts and directly generate bounding-box outputs in a parsable format.
However, promptable bounding box generation introduces new ambiguities as models can differ in how they express coordinates and different formats can influence localization performance.
As a result, the output can often not be directly compared to the ground truth definitions of existing datasets. 
%
%
The following benchmark addresses this problem by proposing a framework that allows for direct comparison of generalist MLLMs for the task of visual localization. To this end, we leverage classical datasets for four common localization tasks: (1) object detection, (2) referring expression detection, (3) instance localization, and (4) video object detection as shown in \Cref{fig:tasks}. We consider different bounding box as well as structured output formats to identify the best format for each model. 
We further analyze the models' overall ability to follow different bounding box formats, e.g., upper-left and lower-right corners vs. center-point-width-height information. We finally consider their localization capabilities with respect to different structured output formats, namely comparing text vs. JSON-formatted output. 
Our evaluation comprises open-source and closed-source MLLMs, showing that the frontier closed-source models (GPT-6 Astra and Claude Opus 5.5) lead on every task family. 
Yet, the open-source models can have better localization abilities than previous-generation closed-source models. 
However, they struggle to deviate from a specific output format, while closed-source models are more robust to format variations.
Importantly, we observe that while prompting for different bounding box formats, the format itself can be correct, while localization results can still be wrong. 
This is based on the fact that models will usually adhere to the format they are trained on and most bounding box formats are quadruples. Thus, models will output four values but those values will not represent the information as requested in the prompt, but rather follow the training format.

We summarize the contributions of this work as follows: 
(1) We propose a first systematic benchmark for visual detection performance of MLLMs. 
(2) We consider 4 task categories: object detection, referring expression detection, instance localization and video object detection. 
(3) We evaluate the format instruction following capabilities with respect to bounding box definitions as well as text vs. JSON-formatted outputs, showing that current MLLMs are highly attuned towards a specific format and cannot interpolate.
(4) We leverage an extensive benchmark suite to assess current open-source and closed-source models' detection capabilities at scale.


\section{Related Work}
\label{sec:related}
\vspace{-2mm}
\noindent\textbf{Datasets and benchmarks for localization and detection.}
Detection evaluates a model's ability to localize image regions corresponding to a natural-language query.
Standard object detection benchmarks built on a fixed taxonomy~\citep{lvis,objects365,openimages,ilsvrc, coco, pascal}, using category names as queries, require all instances of each category to be localized.
Compared to that, referring expression datasets~\citep{referitgame,refcocog,refcoco_unc} replace the category names with a natural-language description that usually identifies a specific object in the image.
Related to this are also phrase grounding datasets such as 
\citet{flickr30kentities} and 
\citet{phrasecut}, which ground free-form noun phrases against one or more regions.
\citet{vg} and its denser extension
\citet{svg} provide region descriptions and scene-graph annotations at scale.
Instance detection benchmarks~\citep{hrinsdet,robotools,perseg,pdm} require localization of one specific object instance, conditioned on a reference image of that instance. In this case, the query is not based on text but is usually given in the form of another image, asking the model to relate between visual features. 
Other benchmarks vary the task further: descriptions may match zero, one, or multiple instances~\citep{d3,omnilabel,rod,mcbench}, target longer expressions and broader category coverage~\citep{refl4,hcrefloco}, focus on small or fine-grained objects~\citep{sorec}, or require external world knowledge~\citep{knowdr}. 
The proposed benchmark builds upon those ideas, and uses them for MLLM localization assessment by considering the variable output formats of MLLMs and parsing them to allow them to match the format of specific datasets.

\vspace{-2mm}
\noindent\textbf{Generalist multimodal LLMs.} Generalist MLLMs take an image and a text prompt as input and return a free-form text response.
Most MLLMs combine a vision encoder with a large language model through a learned projection.
The LLaVA family~\citep{llava,llavanext,improvedllava} pairs a CLIP encoder with a pretrained language model via a trainable connector trained on image-text instruction data.
\citet{internvl3} aligns a large vision encoder with the language model via large-scale contrastive pre-training and supports image and video inputs.
The Qwen-VL family~\citep{qwen2.5vl,qwen3vl,qwen3.5} introduces native dynamic resolution and document understanding.
\citet{molmo} trains on a fully open and human-annotated instruction corpus and emphasizes pointing as a primary localization output.
\citet{cogvlm} inserts trainable visual experts into each transformer layer of the language model.
Gemma-4~\citep{gemma4} releases an open-weight multimodal family, and \citet{glm4.6v} applies scalable reinforcement learning to multimodal reasoning.
Proprietary models from \citet{gpt5,gpt6astra}, \citet{gemini2.5}, and \citet{sonnet4.5,opus5.5} are accessed through APIs without publishing weights or architectural details. 
Generalist MLLMs perform well on a range of vision-language tasks and are widely used across research and everyday use, including for visual grounding.
However, an MLLM's output format is set by the prompt, unlike specialist models that produce one fixed format.
A fixed-prompt benchmark therefore evaluates each MLLM at only one of the many formats it can produce.
\benchmark{} extends visual grounding evaluation to MLLMs by varying the bounding-box representation and the output format in the prompt.
\section{Benchmark}
\label{sec:benchmark}
\vspace{-2mm}
Visual detection performance for generalist MLLMs depends on three factors: first, the performance on the visual task itself; second, the optimal bounding-box representation; and third, the preferred output format of a specific model.
As a result, a model whose preferred format differs from the benchmark's scores poorly for reasons unrelated to grounding. Compared to standard object detection benchmarks, which assume a fixed, matching format, 
\benchmark{} is designed to vary all these factors, so the model's grounding ability is measured at the format where it performs the best.
We design the benchmark as a grid over three axes: (i) \emph{task and data}, spanning four task families and thirteen datasets (\Cref{subsec:tasks}); (ii) \emph{bounding-box representation}, comprising corner-based, center-with-size, and four-corner formats (\Cref{subsec:bbox}); and (iii) \emph{output format}, covering plain text and JSON variants with different keys (\Cref{subsec:format}).
Finally, we measure both how well a model localizes and how strongly its score depends on the format.

\vspace{-2mm}
\subsection{Task families and datasets}
\label{subsec:tasks}
\vspace{-2mm}
We consider four task families which all address the same problem, producing a bounding box for localization, in different contexts: 
\emph{Object detection} uses one or multiple class labels to indicate which objects to localize, \emph{referring expressions} uses a free-form natural-language description for the object, testing also the model's language understanding capabilities, and \emph{instance detection} uses a visual example as reference. Finally, \emph{video} extends object and instance detection to multi-frame input. Note that for each dataset, we sample up to 1,000 queries from each evaluation split with a fixed seed (42) for reproducibility. 
\Cref{tab:datasets} summarizes all datasets and their per-subset statistics. The detailed prompts for each task can be found in the supplementary~\Cref{app:prompts}. 

\begin{table}[t]
    \centering
    \resizebox{0.8\textwidth}{!}{
    \setlength{\tabcolsep}{4pt}
    \begin{tabular}{lllrr}
    \toprule
    Dataset & Source & Labels & Queries & Boxes/query \\
    \midrule
    \multicolumn{5}{l}{\textit{Object detection}} \\
    Pascal VOC          & curated images                              & 20 classes                       & 1{,}000             & 2.03 / 3.05 \\
    OpenImages V7       & web images                                  & 184 classes                      & 1{,}000             & 2.30 / 8.47 \\
    \midrule
    \multicolumn{5}{l}{\textit{Referring expressions}} \\
    RefCOCO             & COCO                                        & free-form                        & 3 $\times$ 1{,}000  & 1.00 \\
    RefCOCO+            & COCO                                        & free-form, no spatial words      & 2 $\times$ 1{,}000  & 1.00 \\
    RefCOCO-g           & COCO                                        & free-form, longer descriptions   & 1{,}000             & 1.00 \\
    RefL4               & COCO                                        & free-form, cleaner re-annotation & 1{,}000             & 1.00 \\
    D3                  & curated                                     & free-form, with negatives        & 1{,}000             & 0.47 \\
    PhraseCut           & Visual Genome                               & free-form                        & 1{,}000             & 1.71 \\
    Flickr30k Entities  & Flickr                                      & free-form                        & 1{,}000             & 1.58 \\
    Synthetic VG        & scene graphs                                & free-form, scene-graph nodes     & 1{,}000             & 1.00 \\
    \midrule
    \multicolumn{5}{l}{\textit{Instance detection}} \\
    HR-InsDet (easy)    & high-res scenes ($\sim$8K $\times$ 6K)      & 98 instances                     & 1{,}000             & 1.01 \\
    HR-InsDet (hard)    & high-res scenes ($\sim$8K $\times$ 6K)      & 61 instances                     & 963                 & 1.00 \\
    \midrule
    \multicolumn{5}{l}{\textit{Video}} \\
    iGround             & instructional video, 21--71 frames/clip     & free-form                        & 1{,}000             & 42.7 / 118.8 \\
    RoboTools           & robot video, 35--129 frames/clip            & 20 instances                     & 161                 & 56.6 \\
    \bottomrule
    \end{tabular}
    }
    \vspace{-2mm}
    \caption{Datasets in \benchmark{}, grouped by task family. The benchmark is based on a subset of up to 1{,}000 queries of each dataset. \emph{Boxes/query} values report \emph{single-label/multi-label} averages. $N \times 1{,}000$ indicates $N$ splits of 1{,}000 sampled with a fixed seed. 
    }
    \label{tab:datasets}
    \vspace{-4mm}
\end{table}

\noindent\textbf{Object detection.}
We consider the most common task, object detection, using three datasets: \emph{Pascal VOC}~\citep{pascal}, \emph{OpenImages V7}~\citep{openimages}, and \emph{iGround}~\citep{iground}.
\emph{Pascal VOC} covers 20 classes.
\emph{OpenImages} contains 184 classes (in our subset) localized in web images.
\emph{iGround} is a manually annotated grounded video-caption dataset, adapted for object detection with each grounded noun as a class label.
We consider two scenarios, \textit{single-label object detection}, where the model is given only one class label per query, as well as \textit{multi-label detection} where several labels are given and the model must localize every instance of each label.

\noindent\textbf{Referring expressions.}
The model is given a natural-language description and must localize every object that matches it in the image.
We use \emph{RefCOCO/+/g}~\citep{referitgame,refcocog, refcoco_unc}, \emph{RefL4}~\citep{refl4}, \emph{D3}~\citep{d3}, \emph{PhraseCut}~\citep{phrasecut}, \emph{Flickr30k Entities}~\citep{flickr30kentities}, and \emph{Synthetic Visual Genome}~\citep{svg}.
For most datasets, exactly one object matches; in \emph{D3}, \emph{PhraseCut}, and \emph{Flickr30k Entities} a query may match several objects, and in \emph{D3} it may also match none.
All these datasets do not only test the models' localization capabilities but also require a strong vision-language alignment to translate the respective language cues to a specific region in the image.

\noindent\textbf{Instance detection.} Different from referring expressions, instance detection actually tests the model's ability to relate visual information between two images. Here, the model is given a support image of a target instance and must localize that instance in the image.
We include this task specifically as it reflects a scenario where models need to relate between visual information, which can be a relevant skill for reasoning in high-resolution visual space as well as for practical robotic scenarios. 
Consequently, we leverage two datasets for this task: \emph{HR-InsDet}~\citep{hrinsdet} and \emph{RoboTools}~\citep{robotools}.
\emph{HR-InsDet} contains high-resolution ($\sim$8K $\times$ 6K) scenes and comes in easy and hard variants.
\emph{RoboTools} pairs videos of robotic manipulation with a support image of the tool to detect.

\noindent\textbf{Video detection.} Finally, we want to assess the localization capabilities if the model is given more than one image, as e.g. in case of video sequences. 
Here, the model receives multiple frames from a clip and must localize the queried target in each frame.
We extend object detection to multi-frame input on \emph{iGround}, and instance detection to multi-frame input on \emph{RoboTools}.
Different from the single-frame use of these datasets, here we sample two or eight frames uniformly per clip.

\subsection{Bounding-box representation}
\label{subsec:bbox}
\begin{figure}[t]
    \centering
    \includegraphics[width=\textwidth]{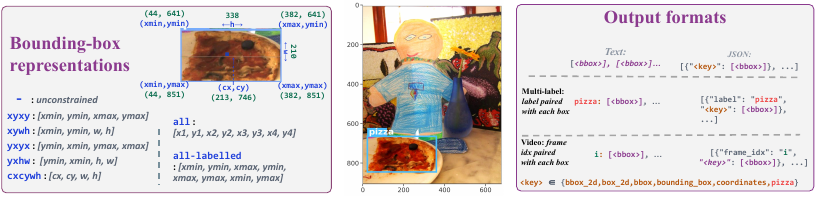}
    \vspace{-8mm}
    \caption{Overview of bounding box and structured format output variations: With respect to bounding boxes, we consider seven variations. For text and JSON output, we consider different options for single-label, multi-label, multi-frame input and also evaluate different JSON key variations. 
    }
    \label{fig:format}
    \vspace{-4mm}
\end{figure}

\vspace{-2mm}
While bounding boxes are the go-to representation for object detection, the same bounding box information can be expressed in various terms. This was not a problem as long as the bounding box was created by a fixed algorithm and evaluated by a fixed framework. 
We found that MLLMs change this situation, because the output is no longer algorithmically determined, but based on free-form text. 
This raises the problem that the user needs to know which format the model outputs by default (e.g., with respect to coordinate order and scale), which differs between model families and is documented in different places, if at all.
Further, a downstream system may require a specific convention regardless of the model's default. 
While this seems addressable via prompting (the model can be prompted for different formats), it shows that, in practice, the model will often not adhere to the requested format~\citep{ifbench}, and, second, even if it does, it might not maintain the same level of accuracy. 
Note that those problems are inherent to generalist MLLM detection and do not occur in the case of specialized models.
To assess the performance of different models with respect to different output bbox structures, we vary the bounding-box representation across seven types as shown in~\Cref{fig:format}:
(1) \emph{Two-corner} lists the top-left and bottom-right corners as $[x_{\min}, y_{\min}, x_{\max}, y_{\max}]$ (\texttt{xyxy}).
(2) \emph{Corner-with-size} lists the top-left corner with width and height as $[x_{\min}, y_{\min}, w, h]$ (\texttt{xywh}).
(3+4) For two-corner and corner-with-size, we additionally test variants with the $y$-coordinate listed first: \texttt{yxyx} ($[y_{\min}, x_{\min}, y_{\max}, x_{\max}]$) and \texttt{yxhw} ($[y_{\min}, x_{\min}, h, w]$).
(5) \emph{Center-with-size} lists the box center with width and height as $[c_x, c_y, w, h]$ (\texttt{cxcywh}).
(6) \emph{Four-corner} lists all four corners.
The \texttt{all} variant gives them as a flat sequence $[x_1, y_1, x_2, y_2, x_3, y_3, x_4, y_4]$.
(7) The \texttt{all-labelled} variant names each corner: $[x_{\min}, y_{\min}, x_{\max}, y_{\min}, x_{\max}, y_{\max}, x_{\min}, y_{\max}]$.
These follow conventions in practical use, e.g., \texttt{xyxy} in \citet{pascal}, \texttt{xywh} in \citet{coco}, \texttt{yxyx} in \citet{gemini2.5}, \texttt{cxcywh} in \citet{sapkota2026ultralyticsyoloevolutionoverview} and \citet{Carion2020DETR}, and four-corner boxes in \citet{dota}.
Formats on which most models fail are kept deliberately, as they test whether a model follows the requested convention or falls back to its trained one.
We finally also consider an \emph{unconstrained} prompt that omits any format specification from the prompt. To compare different models, we test the models' performance under all those variations and choose the prompt that gives the highest score out of all.

\vspace{-2mm}
\subsection{Output format}
\label{subsec:format}
\vspace{-2mm}
Beyond the problem of bounding box formatting, there is also the problem of what format the structured output is supposed to have. Some models will perform better with structured text, while others will perform better with JSON-formatted outputs, presumably depending on their training. 
We therefore consider the performance of generalist models in both scenarios, one where the bounding box should be returned as plain text (e.g.,\ \texttt{[x\_min, y\_min, x\_max, y\_max]}) and one where the bounding box should be returned as JSON.
Further, in JSON mode, each box is a key-value pair whose value is the coordinate list.
For the key, we test five candidates that MLLMs commonly default to (as observed in the unconstrained settings): \texttt{bbox\_2d}, \texttt{box\_2d}, \texttt{bbox}, \texttt{bounding\_box}, and \texttt{coordinates} as shown in~\Cref{fig:format}.
In multi-label queries (where each query covers multiple classes), we additionally test \texttt{class\_name}, which uses the actual class name as the key (e.g.,\ \texttt{\{"cat": [\ldots]\}}).
Again, we test the models' performance for both cases and different JSON keys and choose the configuration that gives the highest score out of all.
\section{Implementation Details}
\label{sec:implementation}
\vspace{-2mm}
\noindent\textbf{Models.}
We evaluate six open-source models: Qwen2.5-VL-7B~\citep{qwen2.5vl}, Qwen3-VL-8B~\citep{qwen3vl}, Qwen3.5-9B~\citep{qwen3.5}, InternVL3-8B~\citep{internvl3}, Gemma-4-E4B~\citep{gemma4}, and GLM-4.6V-Flash~\citep{glm4.6v}; and five proprietary models: GPT-5.4~\citep{gpt5}, Claude Sonnet~4.5~\citep{sonnet4.5},
Gemini~2.5 Flash~\citep{gemini2.5}, GPT-6 Astra~\citep{gpt6astra}, and Claude Opus~5.5~\citep{opus5.5}.
We route all proprietary-model requests through OpenRouter.
Qwen3.5 is also run with reasoning (Qwen3.5-Thinking).
GLM-4.6V, GPT-5.4, and Gemini 2.5 Flash are run without reasoning.
GPT-6 Astra and Opus 5.5 are run with the providers' default reasoning settings.
We do not know if this provider-side reasoning involves additional steps, e.g., zooming into the image.

\noindent\textbf{Output parsing.}
One challenge in parsing the outputs of those models is that the coordinate spaces used during training (and thus also for predicted outputs) differ across models.
We assume for the Qwen3-VL family, InternVL3, GLM-4.6V, Gemma-4 and Gemini to return coordinates in a $1000\times1000$ normalized grid and for Qwen2.5-VL, GPT, and Claude to return pixel coordinates of the input image.
Additionally, for \emph{HR-InsDet}, the pixel-space coordinates can be subject to model-specific resizing (see~\Cref{app:resizing} for details).
We rescale every prediction to pixel space before scoring.
We further noticed that GLM-4.6V wraps every box prediction in \texttt{<|begin\_of\_box|>...<|end\_of\_box|>} markers and repeats the same block multiple times in a single response. 
The repeated boxes still overlap with the ground truth.
We therefore read only the content of the first \texttt{<|begin\_of\_box|>...<|end\_of\_box|>} block.
Note that we do not apply any further post-processing e.g. in the form of non-maximum suppression.

\noindent\textbf{Metrics.}
To compute the final performance, we extract bounding boxes from the response according to the prompted format, rescale to pixel coordinates if necessary, and finally map predictions to ground-truth boxes by Hungarian matching.
The matching cost between a prediction and a ground-truth box is $1 - \text{IoU}$.
For multi-label queries, the cost becomes $1 - \text{IoU} - \mathbf{1}[\text{label matches}]$, so label agreement always dominates IoU when assigning a prediction. 
We report metrics computed on Hungarian-matched pairs.
\textbf{mIoU} averages the IoU over every matched pair.
\textbf{F1} (F1@0.5) counts a prediction as correct when its IoU with the matched ground-truth box is at least $0.5$; in multi-label queries the predicted and ground-truth labels must also match. 
F1 at stricter IoU thresholds is reported in \Cref{app:f1avg}.
\textbf{Format Adherence} (FA) is the fraction of responses that parse as the prompted format, regardless of whether the parsed boxes match the ground truth. Note that format adherence is a precondition for all other metrics. If the format is not parsable, the output will not be considered for the evaluation. Thus, a low FA can significantly impact performance, as an unparsable output will be treated as an empty list and thus result in a higher false negative count.

\noindent\textbf{Multi-stage Format Search.}
To find the format that elicits each model's best score, we run a per-model probe before the full evaluation as a multi-stage sweep: Stage 1 selects the bounding-box representation based on the model's text vs. JSON output.
For the JSON output, we set the default key to \texttt{bbox\_2d}. 
Stage 2 then explores the respective best JSON key based on the bounding box configuration chosen in Stage 1. 
We choose Pascal VOC~\citep{pascal} as the reference dataset for the format search, assuming that this should be the dataset that most models should be most familiar with.
Each combination is run on 50 queries for open-source models and 20 for proprietary models to limit API cost. The single steps for this selection are displayed in~\Cref{tab:format_ablation_mini}. 
Finally, we use the best selected text format, the best JSON format, and the unformatted prompt of each model and apply it on each task independently and report the setting that gives the best average F1@0.5 score for this task. 
 A detailed description of the format search can also be found in~\Cref{app:sandbox}.
\section{Evaluation}
\label{sec:results}
\vspace{-2mm}

\subsection{Comparison of State-of-the-art}
\vspace{-2mm}
We compare all models on the four tasks. We report the best-performing setup for each model based on its average F1@0.5 score, and detail the format of the best configuration. This is an upper bound on performance; \Cref{app:robustness} evaluates robustness to format, prompt wording, and IoU threshold.

\begin{table*}[t]
    \centering
    \resizebox{1.0\textwidth}{!}{
    \setlength{\tabcolsep}{3pt}
    \begin{tabular}{lccc ccc ccc ccc | ccc}
    \toprule
    \multicolumn{16}{c}{Single-label Object Detection} \\
     & & & & \multicolumn{3}{c}{Pascal} & \multicolumn{3}{c}{OpenImages} & \multicolumn{3}{c}{iGround} & \multicolumn{3}{|c}{Average} \\
    \cmidrule(lr){5-7} \cmidrule(lr){8-10} \cmidrule(lr){11-13} \cmidrule(lr){14-16}
    Model & Bbox & Format & JSON key & F1@0.5 & mIoU & FA (\%) & F1@0.5 & mIoU & FA (\%) & F1@0.5 & mIoU & FA (\%) & F1@0.5 & mIoU & FA (\%) \\
    \rowcolor{tablebanner} \multicolumn{16}{l}{\textit{Open Source Models}} \\
    Gemma-4 & \texttt{yxyx} & Text & -- & 68.7 & 56.2 & 99.8 & 34.9 & 26.4 & 94.7 & 53.5 & 42.6 & 92.1 & 52.4 & 41.7 & 95.5 \\
    Qwen2.5VL & \emph{-- (as xyxy)} & Text \emph{(as JSON)} & \emph{-- (as bbox\_2d)} & 70.1 & 52.6 & 98.9 & 49.0 & 37.6 & 98.0 & 42.5 & 40.6 & 90.0 & 53.9 & 43.6 & 95.6 \\
    InternVL3 & \texttt{xyxy} & Text & -- & 77.0 & 62.5 & 99.3 & 34.9 & 27.6 & 95.8 & 53.1 & 53.5 & 96.7 & 55.0 & 47.9 & 97.3 \\
    Qwen3.5 & \texttt{xyxy} & JSON & \texttt{bbox\_2d} & 77.3 & 67.2 & 98.4 & 53.9 & 48.9 & 97.4 & 55.3 & 59.3 & 99.5 & 62.2 & 58.5 & 98.4 \\
    Qwen3.5-th. & \emph{-- (as xyxy)} & Text \emph{(as JSON)} & \emph{-- (as bbox\_2d)} & 75.3 & 58.5 & 94.9 & 52.1 & 41.1 & 87.5 & 62.3 & 59.1 & 83.2 & 63.2 & 52.9 & 88.5 \\
    GLM4.6V & \texttt{xyxy} & Text & -- & 80.5 & 64.9 & 99.4 & 55.9 & 50.0 & 98.2 & 60.3 & 61.1 & 99.5 & \underline{65.5} & 58.7 & 99.0 \\
    Qwen3VL & \emph{-- (as xyxy)} & Text \emph{(as JSON)} & \emph{-- (as bbox\_2d)} & 77.5 & 66.5 & 97.1 & 62.8 & 54.6 & 97.3 & 60.4 & 57.5 & 85.3 & \textbf{66.9} & 59.6 & 93.2 \\
    \rowcolor{tablebanner} \multicolumn{16}{l}{\textit{Closed Source Models}} \\
    Sonnet 4.5 & \texttt{xywh} & JSON & \texttt{bbox} & 27.3 & 31.7 & 99.9 & 24.1 & 25.6 & 99.8 & 22.2 & 30.2 & 100.0 & 24.5 & 29.1 & 99.9 \\
    GPT5.4 & \texttt{xyxy} & Text & -- & 48.8 & 43.3 & 100.0 & 41.3 & 36.1 & 100.0 & 37.3 & 39.9 & 100.0 & 42.5 & 39.7 & 100.0 \\
    Gemini 2.5 Flash & \texttt{yxyx} & JSON & \texttt{bbox} & 74.5 & 64.4 & 94.9 & 46.8 & 42.6 & 83.6 & 53.2 & 57.7 & 74.4 & 58.2 & 54.9 & 84.3 \\
    GPT-6 Astra & \texttt{xywh} & Text & -- & 81.4 & 79.6 & 99.9 & 46.2 & 66.2 & 99.9 & 54.7 & 70.5 & 100.0 & \underline{60.8} & 72.1 & 99.9 \\
    Opus 5.5 & \texttt{yxhw} & Text & -- & 86.3 & 82.0 & 100.0 & 56.7 & 66.6 & 99.7 & 60.2 & 71.8 & 96.7 & \textbf{67.7} & 73.5 & 98.8 \\
    \bottomrule
    \end{tabular}
    }
    \vspace{-2mm}
    \caption{Comparison of single-label object detection: We consider the best performing output for each model based on its average F1 score and list the respective configuration.}
    \label{tab:objdet_single_sota}
    \vspace{-4mm}
\end{table*}

\begin{table*}[t]
    \centering
    \resizebox{1.0\textwidth}{!}{
    \setlength{\tabcolsep}{3pt}
    \begin{tabular}{lccc ccc ccc ccc | ccc}
    \toprule
    \multicolumn{16}{c}{Multi-label Object Detection} \\
     & & & & \multicolumn{3}{c}{Pascal} & \multicolumn{3}{c}{OpenImages} & \multicolumn{3}{c}{iGround} & \multicolumn{3}{|c}{Average} \\
    \cmidrule(lr){5-7} \cmidrule(lr){8-10} \cmidrule(lr){11-13} \cmidrule(lr){14-16}
    Model & Bbox & Format & JSON key & F1@0.5 & mIoU & FA (\%) & F1@0.5 & mIoU & FA (\%) & F1@0.5 & mIoU & FA (\%) & F1@0.5 & mIoU & FA (\%) \\
    \rowcolor{tablebanner} \multicolumn{16}{l}{\textit{Open Source Models}} \\
    Qwen2.5VL & \emph{-- (as xyxy)} & Text \emph{(as JSON)} & \emph{-- (as bbox\_2d)} & 80.5 & 66.8 & 99.2 & 41.5 & 30.8 & 96.4 & 37.6 & 38.1 & 95.2 & 53.2 & 45.3 & 96.9 \\
    Qwen3.5-th. & \texttt{xyxy} & Text & -- & 78.4 & 60.6 & 94.1 & 23.9 & 13.2 & 58.4 & 60.1 & 49.0 & 83.4 & 54.1 & 41.0 & 78.6 \\
    InternVL3 & \texttt{xyxy} & JSON & \texttt{coordinates} & 82.0 & 69.2 & 97.4 & 34.8 & 25.4 & 95.2 & 54.8 & 52.2 & 97.3 & 57.2 & 49.0 & 96.6 \\
    Gemma-4 & \texttt{yxyx} & JSON & \texttt{box\_2d} & 75.1 & 62.6 & 99.9 & 40.0 & 32.1 & 95.4 & 58.7 & 53.7 & 100.0 & 58.0 & 49.5 & 98.4 \\
    Qwen3.5 & \emph{-- (as xyxy)} & Text \emph{(as JSON)} & \emph{-- (as bbox\_2d)} & 80.4 & 70.0 & 97.2 & 45.0 & 32.8 & 87.0 & 57.1 & 57.9 & 96.7 & 60.9 & 53.6 & 93.6 \\
    GLM4.6V & \texttt{xyxy} & Text & -- & 82.2 & 65.3 & 91.0 & 45.8 & 35.5 & 95.8 & 59.7 & 58.3 & 98.4 & \underline{62.5} & 53.1 & 95.1 \\
    Qwen3VL & \emph{-- (as xyxy)} & Text \emph{(as JSON)} & \emph{-- (as bbox\_2d)} & 81.2 & 70.9 & 96.5 & 51.5 & 37.8 & 89.3 & 58.8 & 58.4 & 93.6 & \textbf{63.8} & 55.7 & 93.1 \\
    \rowcolor{tablebanner} \multicolumn{16}{l}{\textit{Closed Source Models}} \\
    Sonnet 4.5 & \texttt{xywh} & JSON & \texttt{coordinates} & 32.3 & 34.5 & 99.9 & 22.9 & 21.1 & 96.1 & 25.9 & 31.6 & 100.0 & 27.0 & 29.1 & 98.7 \\
    GPT5.4 & \texttt{xyxy} & Text & -- & 55.6 & 48.1 & 100.0 & 38.5 & 32.7 & 99.9 & 39.5 & 40.2 & 99.7 & 44.5 & 40.4 & 99.9 \\
    Gemini 2.5 Flash & \texttt{yxyx} & JSON & \texttt{class\_name} & 78.6 & 66.7 & 94.8 & 42.4 & 31.5 & 78.3 & 59.9 & 60.6 & 92.1 & 60.3 & 52.9 & 88.4 \\
    GPT-6 Astra & \texttt{xywh} & Text & -- & 82.8 & 81.4 & 100.0 & 47.3 & 62.0 & 100.0 & 64.4 & 72.3 & 98.7 & \underline{64.8} & 71.9 & 99.6 \\
    Opus 5.5 & \texttt{yxhw} & Text & -- & 85.6 & 81.7 & 100.0 & 50.4 & 54.1 & 99.9 & 65.1 & 71.9 & 97.8 & \textbf{67.0} & 69.2 & 99.2 \\
    \bottomrule
    \end{tabular}
    }
    \vspace{-3mm}
    \caption{Comparison of multi-label object detection: We consider the best performing output for each model based on its average F1 score and list the respective configuration.}
    \label{tab:objdet_multi_sota}
    \vspace{-2mm}
    \end{table*}

\noindent\textbf{Object detection.}
For the case of object detection, we consider single and multi-object detection scenarios in \Cref{tab:objdet_single_sota} and \Cref{tab:objdet_multi_sota}, respectively. 
Overall, it shows that while most models perform well in both categories, the best open-source models are on par with the frontier proprietary models and outperform the previous proprietary generation. 
Opus 5.5 performs best on average in both cases, while Qwen3-VL and GLM-4.6V are the best open-source models.
Qwen3.5-Thinking struggles to keep up on multi-object detection, mainly on OpenImages, while Gemma-4 performs better on multi-object than on single-object detection.

\begin{table*}[t]
    \centering
    \resizebox{1.0\textwidth}{!}{
    \setlength{\tabcolsep}{3pt}
    \begin{tabular}{lccc ccc ccc ccc ccc ccc ccc |ccc}
    \toprule
    \multicolumn{25}{c}{Referring Expression Detection}\\
     & & & & \multicolumn{3}{c}{RefCOCO-Avg} & \multicolumn{3}{c}{RefL4} & \multicolumn{3}{c}{Flickr30k-Entities} & \multicolumn{3}{c}{D3} & \multicolumn{3}{c}{PhraseCut} & \multicolumn{3}{c}{SVG} & \multicolumn{3}{|c}{Average} \\
    \cmidrule(lr){5-7} \cmidrule(lr){8-10} \cmidrule(lr){11-13} \cmidrule(lr){14-16} \cmidrule(lr){17-19} \cmidrule(lr){20-22} \cmidrule(lr){23-25}
    Model & Bbox & Format & JSON key & F1@0.5 & mIoU & FA (\%) & F1@0.5 & mIoU & FA (\%) & F1@0.5 & mIoU & FA (\%) & F1@0.5 & mIoU & FA (\%) & F1@0.5 & mIoU & FA (\%) & F1@0.5 & mIoU & FA (\%) & F1@0.5 & mIoU & FA (\%) \\
    
    \rowcolor{tablebanner} \multicolumn{25}{l}{\textit{Open Source Models}} \\
    Gemma-4 & \texttt{yxyx} & JSON & \texttt{bounding\_box} & 71.2 & 65.9 & 99.8 & 75.8 & 69.0 & 100.0 & 57.9 & 49.9 & 99.7 & 46.8 & 55.0 & 99.9 & 39.5 & 36.8 & 99.9 & 59.8 & 56.9 & 100.0 & 58.5 & 55.6 & 99.9 \\
    InternVL3 & \texttt{xyxy} & Text & -- & 78.1 & 70.5 & 100.0 & 68.8 & 61.4 & 100.0 & 61.6 & 51.1 & 97.9 & 47.1 & 53.8 & 100.0 & 37.8 & 32.8 & 99.6 & 72.3 & 66.5 & 100.0 & 61.0 & 56.0 & 99.6 \\
    Qwen3.5 & \texttt{xyxy} & JSON & \texttt{bbox\_2d} & 86.5 & 77.6 & 100.0 & 82.4 & 72.6 & 100.0 & 62.0 & 47.8 & 99.9 & 36.2 & 53.7 & 100.0 & 43.9 & 36.2 & 98.8 & 75.1 & 65.2 & 100.0 & 64.4 & 58.8 & 99.8 \\
    Qwen2.5VL & \emph{-- (as xyxy)} & Text \emph{(as JSON)} & \emph{-- (as bbox\_2d)} & 85.5 & 78.7 & 100.0 & 82.3 & 74.4 & 99.9 & 58.1 & 46.8 & 99.8 & 51.7 & 48.9 & 97.9 & 49.4 & 40.1 & 99.4 & 74.1 & 66.2 & 100.0 & 66.9 & 59.2 & 99.5 \\
    Qwen3.5 - th & \texttt{xyxy} & Text & -- & 87.2 & 77.4 & 99.3 & 83.2 & 72.4 & 98.8 & 63.1 & 51.8 & 96.9 & 47.9 & 60.4 & 98.1 & 48.5 & 39.2 & 96.7 & 73.3 & 64.4 & 99.3 & 67.2 & 60.9 & 98.2 \\
    Qwen3VL & \texttt{xyxy} & JSON & \texttt{bbox\_2d} & 87.3 & 79.9 & 99.9 & 88.4 & 80.1 & 100.0 & 57.5 & 45.0 & 100.0 & 46.9 & 53.3 & 99.8 & 48.4 & 39.7 & 99.2 & 77.4 & 68.6 & 100.0 & \underline{67.7} & 61.1 & 99.8 \\
    GLM4.6V & \texttt{xyxy} & Text & -- & 84.3 & 78.3 & 99.1 & 86.7 & 79.3 & 98.9 & 74.5 & 64.6 & 99.5 & 39.6 & 59.0 & 98.2 & 49.7 & 41.2 & 98.7 & 77.0 & 68.5 & 99.2 & \textbf{68.6} & 65.2 & 98.9 \\
    
    \rowcolor{tablebanner} \multicolumn{25}{l}{\textit{Closed Source Models}} \\
    Sonnet 4.5 & \texttt{xywh} & JSON & \texttt{bbox} & 31.3 & 37.5 & 99.4 & 26.3 & 31.9 & 92.6 & 21.6 & 24.3 & 100.0 & 25.0 & 35.9 & 99.9 & 16.6 & 23.6 & 99.9 & 18.8 & 27.6 & 100.0 & 23.3 & 30.1 & 98.6 \\
    GPT5.4 & \texttt{xyxy} & Text & -- & 62.5 & 52.7 & 100.0 & 60.9 & 52.6 & 100.0 & 40.1 & 36.7 & 100.0 & 41.2 & 41.5 & 100.0 & 28.1 & 30.0 & 99.9 & 36.3 & 38.4 & 100.0 & 44.8 & 42.0 & 100.0 \\
    Gemini 2.5 Flash & \texttt{yxyx} & JSON & \texttt{bbox} & 74.7 & 70.7 & 97.7 & 74.4 & 69.0 & 97.8 & 57.4 & 55.3 & 90.3 & 50.8 & 60.4 & 95.6 & 36.9 & 39.7 & 83.2 & 66.8 & 65.1 & 96.4 & 60.2 & 60.0 & 93.5 \\
    GPT-6 Astra & \texttt{xywh} & Text & -- & 92.4 & 85.4 & 100.0 & 93.3 & 85.5 & 100.0 & 59.0 & 68.7 & 99.8 & 53.3 & 81.2 & 99.9 & 45.8 & 55.4 & 99.9 & 77.8 & 70.3 & 100.0 & \underline{70.3} & 74.4 & 99.9 \\
    Opus 5.5 & \texttt{yxhw} & Text & -- & 90.2 & 83.9 & 100.0 & 91.4 & 84.2 & 100.0 & 63.0 & 61.9 & 99.9 & 62.2 & 78.0 & 100.0 & 51.9 & 55.7 & 100.0 & 81.2 & 74.1 & 100.0 & \textbf{73.3} & 73.0 & 100.0 \\
    \bottomrule
    \end{tabular}
    }
    \vspace{-3mm}
    \caption{Comparison of referring expression results across six datasets: We consider the best performing output for each model based on its average F1 score and list the respective configuration. }
    \label{tab:refex_sota_main}
    \vspace{-3mm}
    \end{table*}

\noindent\textbf{Referring expressions.}
Second, we evaluate all models for the task of referring expression detection as shown in~\Cref{tab:refex_sota_main}. We observe that the models achieve even higher absolute average performance here than in classical object detection tasks. 
The frontier models lead again, with Opus 5.5 and GPT-6 Astra ahead of GLM-4.6V, which reaches the best performance among all open-source models here.
Opus 5.5 gains most on D3, where 70\% of the queries have no matching object, while GLM-4.6V remains the best model on Flickr30k-Entities.
Reasoning helps on this task: Qwen3.5-Thinking beats Qwen3.5 and, unlike in multi-object detection, almost matches Qwen3-VL.

\begin{table*}[t]
    \centering
    \resizebox{1.0\textwidth}{!}{
    \setlength{\tabcolsep}{3pt}
    \begin{tabular}{lccc ccc ccc ccc |ccc}
    \toprule
    \multicolumn{16}{c}{Instance Detection} \\
     & & & & \multicolumn{3}{c}{HR-InsDet {\small\textit{easy}}} & \multicolumn{3}{c}{HR-InsDet {\small\textit{hard}}} & \multicolumn{3}{c}{RoboTools} & \multicolumn{3}{|c}{Average} \\
    \cmidrule(lr){5-7} \cmidrule(lr){8-10} \cmidrule(lr){11-13} \cmidrule(lr){14-16}
    Model & Bbox & Format & JSON key & F1@0.5 & mIoU & FA (\%) & F1@0.5 & mIoU & FA (\%) & F1@0.5 & mIoU & FA (\%) & F1@0.5 & mIoU & FA (\%) \\
    
    \rowcolor{tablebanner} \multicolumn{16}{l}{\textit{Open Source Models}} \\
    Gemma-4 & \texttt{yxyx} & Text & -- & 0.2 & 0.8 & 100.0 & 0.0 & 0.4 & 100.0 & 0.6 & 6.4 & 100.0 & 0.3 & 2.5 & 100.0 \\
    InternVL3 & \texttt{xyxy} & JSON & \texttt{coordinates} & 0.0 & 0.6 & 98.1 & 0.0 & 0.3 & 99.0 & 1.5 & 7.2 & 99.4 & 0.5 & 2.7 & 98.8 \\
    Qwen2.5VL & \emph{-- (as xyxy)} & Text \emph{(as JSON)} & \emph{-- (as bbox\_2d)} & 14.5 & 18.8 & 99.8 & 16.1 & 18.7 & 99.8 & 55.3 & 49.0 & 93.8 & 28.7 & 28.8 & 97.8\\

    Qwen3.5-th. & \emph{-- (as xyxy)} & Text \emph{(as JSON)} & \emph{-- (as bbox\_2d)} & 47.6 & 43.3 & 95.3 & 24.8 & 30.1 & 91.3 & 32.1 & 32.7 & 91.9 & 34.8 & 35.4 & 92.8 \\
   Qwen3.5 &  \texttt{xyxy} & Text & -- & 52.8 & 45.3 & 99.6 & 31.8 & 33.6 & 99.1 & 28.1 & 32.8 & 100.0  & 37.6 & 37.3 & 99.6 \\
    GLM4.6V & \texttt{xyxy} & Text & -- & 64.1 & 54.5 & 99.3 & 41.1 & 37.6 & 99.7 & 74.8 & 72.0 & 99.4 & \underline{60.0} & 54.7 & 99.5 \\
      Qwen3VL& \texttt{xyxy} & JSON & \texttt{bbox\_2d} & 62.3 &  51.1 &  99.4  &  42.8 &  38.6 &  99.6 &  82.0 &  72.4 &  99.4 & \textbf{62.3} & 54.0 & 99.5 \\
    
    \rowcolor{tablebanner} \multicolumn{16}{l}{\textit{Closed Source Models}} \\
    Sonnet 4.5 & \texttt{xywh} & JSON & \texttt{bbox} & 12.6 & 18.3 & 99.8 & 7.8 & 11.3 & 92.0 & 15.0 & 21.1 & 100.0 & 11.8 & 16.9 & 97.3 \\
    Gemini 2.5 Flash & \texttt{yxyx} & JSON & \texttt{box\_2d} & 17.2 & 18.0 & 86.6 & 4.5 & 7.8 & 94.2 & 32.3 & 34.7 & 96.3 & 18.0 & 20.2 & 92.4 \\
    GPT5.4 & \texttt{xywh} & JSON & \texttt{coordinates} & 20.0 & 25.3 & 100.0 & 8.1 & 12.6 & 100.0 & 47.9 & 45.7 & 100.0 & 25.3 & 27.8 & 100.0 \\
    Opus 5.5 & \texttt{yxhw} & Text & -- & 81.6 & 76.0 & 99.9 & 70.3 & 66.0 & 99.6 & 60.4 & 57.4 & 95.0 & \underline{70.8} & 66.4 & 98.2 \\
    GPT-6 Astra & \texttt{cxcywh} & JSON & \texttt{bounding\_box} & 93.3 & 89.6 & 100.0 & 82.8 & 79.4 & 100.0 & 93.8 & 90.5 & 100.0 & \textbf{90.0} & 86.5 & 100.0 \\
    \bottomrule
    \end{tabular}
    }
    \vspace{-3mm}
    \caption{Comparison of instance detection results: We consider the best
performing output for each model based on its average F1 score and list the respective configuration.}
    \label{tab:insdet_sota}
    \vspace{-3mm}
    \end{table*}

\noindent\textbf{Instance detection.}
Unlike the two previous tasks, instance detection focuses less on language and more on visual capabilities, a point most models seem to struggle on as shown in \Cref{tab:insdet_sota}.
GPT-6 Astra leads by a wide margin, followed by Opus 5.5 and the best open-source models, Qwen3-VL and GLM-4.6V.
Behind them, Qwen3.5 and Qwen3.5-Thinking perform almost on par, with Qwen3.5 showing a slight advantage on high-resolution instance detection and Qwen3.5-Thinking showing a slightly better performance on RoboTools, while InternVL3 and Gemma-4 fail almost completely.
Notably, none of the previous proprietary models can keep up with the best-performing open-source models, and Sonnet 4.5 reaches only 11.8 F1.

\begin{table*}[t]
    \centering
    \resizebox{1.0\textwidth}{!}{
    \setlength{\tabcolsep}{3pt}
    \begin{tabular}{lccc ccc ccc ccc ccc | ccc}
    \toprule
    \multicolumn{19}{c}{Video Detection} \\
    \midrule
     & & & & \multicolumn{6}{c}{RoboTools (Inst. Det.)} & \multicolumn{6}{c}{iGround (Single Obj. Det.)} & \multicolumn{3}{|c}{\multirow{2}{*}{Average}} \\
    \cmidrule(lr){5-10} \cmidrule(lr){11-16}
     & & & & \multicolumn{3}{c}{2 frames} & \multicolumn{3}{c}{8 frames} & \multicolumn{3}{c}{2 frames} & \multicolumn{3}{c|}{8 frames} & & & \\
    \cmidrule(lr){5-7} \cmidrule(lr){8-10} \cmidrule(lr){11-13} \cmidrule(lr){14-16}
    Model & Bbox & Format & JSON key & F1@0.5 & mIoU & FA (\%) & F1@0.5 & mIoU & FA (\%) & F1@0.5 & mIoU & FA (\%) & F1@0.5 & mIoU & FA (\%) & F1@0.5 & mIoU & FA (\%) \\
    
    \rowcolor{tablebanner} \multicolumn{19}{l}{\textit{Open Source Models}} \\
    Qwen3.5 & \texttt{xyxy} & Text & -- & 1.8 & 1.2 & 12.4 & 3.0 & 5.1 & 13.7 & 16.2 & 9.1 & 22.0 & 16.7 & 9.7 & 25.4 & 9.4 & 6.3 & 18.4 \\
    Qwen2.5VL & \texttt{xyxy} & JSON & \texttt{bbox\_2d} &  9.3 &  5.6 &  14.3 & 2.4  &  6.0  &  99.4 & 31.9 & 33.5 & 94.5 & 24.4 & 25.1 & 74.5 & 17.0 & 17.5 & 70.7 \\
    Gemma-4 & \texttt{yxyx} & JSON & \texttt{bounding\_box} & 0.0 & 3.0 & 42.9 & 0.7 & 2.9 & 73.9 & 39.0 & 40.0 & 86.5 & 28.9 & 28.7 & 85.3 & 17.2 & 18.6 & 72.1 \\
    InternVL3 & \texttt{xyxy} & Text & -- & 0.3 & 2.6 & 100.0 & 0.0 & 2.2 & 4.3 & 44.5 & 45.1 & 99.7 & 27.1 & 29.9 & 99.3 & 18.0 & 20.0 & 75.8 \\
    Qwen3.5 - th & \texttt{xyxy} & JSON & \texttt{bbox\_2d} & 16.0 & 17.0 & 66.5 & 5.3 & 8.7 & 82.0 & 51.5 & 43.2 & 83.8 & 48.2 & 40.0 & 88.7 & 30.2 & 27.2 & 80.2 \\
    Qwen3VL & \texttt{xyxy}  & JSON & \texttt{bbox\_2d} &  59.9 &  54.2 &  100.0  & 18.5 &  21.7 &  91.9  & 47.3 & 45.7 & 94.7 & 39.0 & 38.8 & 95.6 & \underline{41.1} & 40.1 & 95.6 \\
    GLM4.6V & \texttt{xyxy} & Text & -- & 69.9 & 61.2 & 66.5 & 35.4 & 32.5 & 82.6 & 53.7 & 49.4 & 90.1 & 44.7 & 39.2 & 88.1 & \textbf{50.9} & 45.6 & 81.8 \\
    
    \rowcolor{tablebanner} \multicolumn{19}{l}{\textit{Closed Source Models}} \\
    Sonnet 4.5 & \texttt{xyxy} & Text & -- & 13.6 & 17.8 & 100.0 & 4.2 & 8.0 & 100.0 & 22.8 & 28.2 & 99.7 & 22.9 & 27.3 & 99.5 & 15.9 & 20.3 & 99.8 \\
    Gemini 2.5 Flash & \texttt{yxyx} & Text & -- & 18.8 & 20.8 & 85.7 & 6.5 & 10.9 & 95.0 & 53.5 & 48.3 & 96.0 & 36.9 & 36.9 & 98.9 & 28.9 & 29.2 & 93.9 \\
    GPT5.4 & \texttt{xyxy} & Text & -- & 31.7 & 34.4 & 100.0 & 17.8 & 22.8 & 100.0 & 39.7 & 38.4 & 99.3 & 36.8 & 37.0 & 98.9 & 31.5 & 33.2 & 99.6 \\
    Opus 5.5 & \texttt{yxhw} & Text & -- & 93.2 & 88.2 & 98.8 & 95.8 & 88.6 & 98.8 & 61.1 & 59.0 & 97.9 & 63.1 & 58.4 & 99.9 & \underline{78.3} & 73.6 & 98.8 \\
    GPT-6 Astra & \texttt{cxcywh} & JSON & \texttt{bounding\_box} & 96.1 & 91.8 & 93.8 & 98.5 & 92.6 & 77.0 & 60.1 & 60.2 & 91.9 & 68.0 & 62.1 & 77.5 & \textbf{80.7} & 76.7 & 85.1 \\
    \bottomrule
    \end{tabular}
    }
    \vspace{-3mm}
    \caption{Comparison of video detection results: We consider the best performing output for each model based on its average F1 score.}
    \label{tab:video_sota}
    \vspace{-3mm}
    \end{table*}

\noindent\textbf{Video detection.}
Finally, we assess the models' multi-frame detection capabilities in \Cref{tab:video_sota}.
Note that while some MLLMs accept a video file as input directly (e.g., Qwen's video chat template), our preliminary tests show that video input performs worse than passing pre-sampled frames as a multi-image prompt, and not every evaluated model accepts video input (e.g., the Claude API accepts only images). We therefore use multi-image prompting (2 or 8 uniformly sampled frames) across all models. 
Overall, GPT-6 Astra and Opus 5.5 lead by a wide margin over GLM-4.6V, the best open-source model, mainly driven by video instance detection, a task that most other models fail. 
On video object detection (iGround, single-object mode), the margin is smaller, and with two frames GLM-4.6V and Qwen3.5-Thinking come close to the frontier models. 
%
Note that because the video task leverages datasets from object and instance detection, it also allows for a direct comparison of how model performance changes when given multiple input images. 
Overall, we observe that models struggle with longer visual inputs and, in most cases, show a drastic performance drop compared to the single-image baseline, as well as when extending the task from two to eight frames. 
The frontier models are the exception, as their performance does not drop from two to eight frames.


\begin{figure}
    \vspace{-4mm}
    \centering
    \includegraphics[width=\textwidth]{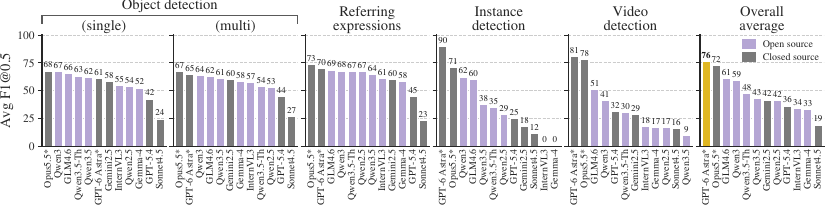}
    \vspace{-8mm}
    \caption{Overview of cross-task performance given the best configuration for each model and each task. The overall average weights the four task families equally, with object detection as the mean of the single- and multi-object settings. GPT-6 Astra and Opus 5.5 lead overall, mainly through instance and video detection, and GLM-4.6V is the best open-source model. $^*$API-side reasoning.}
    \label{fig:cross-task}
    \vspace{-4mm}
\end{figure}

\noindent\textbf{Cross-task performance.}
We summarize the performance of all models across all tasks in~\Cref{fig:cross-task}. 
Averaging the F1 performance over the four task families, GPT-6 Astra and Opus 5.5 perform best, followed by the best open-source models GLM-4.6V and Qwen3-VL.
The lead of the frontier models comes mostly from instance and video detection, whereas on object detection and referring expression the best open-source models are close.

\vspace{-2mm}
\subsection{Bounding-box representation}
\vspace{-2mm}
\begin{table*}[t]
    \centering
    \resizebox{1.0\textwidth}{!}{
    \setlength{\tabcolsep}{4pt}
    \renewcommand{\arraystretch}{1.05}
    \newcolumntype{N}{>{\centering\arraybackslash}p{2.4em}}
    \begin{tabular}{ll  | NN | NN NN NN NN NN || NN NN NN NN NN NN}
    \toprule
     & & \multicolumn{12}{c|}{Stage 1: Bounding-box representation} & \multicolumn{12}{c}{Stage 2: JSON key} \\
    \cmidrule(lr){3-14} \cmidrule(lr){15-26}
     & & \multicolumn{2}{c}{\texttt{--}}
        & \multicolumn{2}{c}{\texttt{xyxy}}
        & \multicolumn{2}{c}{\texttt{xywh}}
        & \multicolumn{2}{c}{\texttt{yxhw}}
        & \multicolumn{2}{c}{\makecell{\texttt{all-}\\\texttt{lab.}}}
        & \multicolumn{2}{c|}{\texttt{cxcywh}+\,\emph{def}}
        & \multicolumn{2}{c}{\texttt{bbox\_2d}}
        & \multicolumn{2}{c}{\texttt{box\_2d}}
        & \multicolumn{2}{c}{\texttt{bbox}}
        & \multicolumn{2}{c}{\makecell{\texttt{bounding}\\\texttt{\_box}}}
        & \multicolumn{2}{c}{\makecell{\texttt{coor-}\\\texttt{dinates}}}
        & \multicolumn{2}{c}{\texttt{class\_name}} \\
    \cmidrule(lr){3-4} \cmidrule(lr){5-6} \cmidrule(lr){7-8}
    \cmidrule(lr){9-10} \cmidrule(lr){11-12} \cmidrule(lr){13-14}
    \cmidrule(lr){15-16} \cmidrule(lr){17-18} \cmidrule(lr){19-20}
    \cmidrule(lr){21-22} \cmidrule(lr){23-24} \cmidrule(lr){25-26}
    Model & Output
      & F1 & FA & F1 & FA & F1 & FA & F1 & FA & F1 & FA & F1 & FA
      & F1 & FA & F1 & FA & F1 & FA & F1 & FA & F1 & FA & F1 & FA \\
    \rowcolor{tablebanner} \multicolumn{26}{l}{\textit{Single-label}} \\
    \multirow{2}{*}{GLM4.6V} & Text
      & \textbf{79.4} & 98.0 & 78.8 & 98.0 & 16.7 & 100.0 & 7.4 & 98.0 & 0.0 & 2.0 & 0.0 & 100.0
      & -- & -- & -- & -- & -- & -- & -- & -- & -- & -- & -- & -- \\
                              & JSON
      & 0.0 & 0.0 & \textbf{44.1} & 16.0 & 4.0 & 20.0 & 0.0 & 22.0 & 0.0 & 0.0 & 0.0 & 14.0
      & 44.1 & 16.0 & 57.9 & 26.0 & 53.3 & 22.0 & 50.3 & 20.0 & \textbf{58.9} & 30.0 & -- & -- \\
    \cmidrule(lr){1-26}
    \multirow{2}{*}{Qwen3VL} & Text
      & 67.9 & 100.0 & \textbf{69.9} & 100.0 & 19.5 & 100.0 & 8.5 & 100.0 & 0.0 & 6.0 & 0.0 & 100.0
      & -- & -- & -- & -- & -- & -- & -- & -- & -- & -- & -- & -- \\
                              & JSON
      & \textbf{75.5} & 98.0 & 74.6 & 96.0 & 17.2 & 98.0 & 6.7 & 98.0 & 0.0 & 0.0 & 0.0 & 98.0
      & \textbf{74.6} & 96.0 & 65.0 & 72.0 & 74.0 & 94.0 & 60.0 & 62.0 & 62.4 & 66.0 & -- & -- \\
    \cmidrule(lr){1-26}
    \multirow{2}{*}{Opus 5.5} & Text
      & 83.9 & 100.0 & \textbf{84.7} & 100.0 & 83.2 & 100.0 & 82.1 & 100.0 & 83.2 & 100.0 & 81.7 & 100.0
      & -- & -- & -- & -- & -- & -- & -- & -- & -- & -- & -- & -- \\
                              & JSON
      & 82.5 & 100.0 & 80.3 & 100.0 & \textbf{83.2} & 100.0 & \textbf{83.2} & 100.0 & 73.7 & 100.0 & 80.0 & 100.0
      & 83.2 & 100.0 & 7.1 & 100.0 & 81.0 & 100.0 & \textbf{83.9} & 100.0 & 82.1 & 100.0 & -- & -- \\
    \cmidrule(lr){1-26}
    \multirow{2}{*}{GPT-6 Astra} & Text
      & 75.4 & 100.0 & 75.4 & 100.0 & \textbf{76.8} & 100.0 & 66.1 & 100.0 & 62.0 & 95.0 & 76.1 & 100.0
      & -- & -- & -- & -- & -- & -- & -- & -- & -- & -- & -- & -- \\
                              & JSON
      & 0.0 & 0.0 & 75.2 & 100.0 & \textbf{76.9} & 100.0 & 69.0 & 100.0 & 62.4 & 90.0 & 76.5 & 100.0
      & 76.5 & 100.0 & 74.8 & 100.0 & 71.8 & 100.0 & \textbf{77.2} & 100.0 & \textbf{77.2} & 100.0 & -- & -- \\
    \rowcolor{tablebanner} \multicolumn{26}{l}{\textit{Multi-label}} \\
    \multirow{2}{*}{GLM4.6V} & Text
      & 0.0 & 0.0 & \textbf{80.3} & 96.0 & 19.0 & 88.0 & 6.8 & 86.0 & 0.0 & 16.0 & 0.8 & 98.0
      & -- & -- & -- & -- & -- & -- & -- & -- & -- & -- & -- & -- \\
                              & JSON
      & 0.0 & 0.0 & \textbf{63.7} & 44.0 & 12.6 & 48.0 & 1.8 & 40.0 & 0.0 & 0.0 & 0.0 & 42.0
      & 63.7 & 44.0 & \textbf{66.4} & 48.0 & 60.4 & 38.0 & 64.6 & 44.0 & 59.5 & 38.0 & 45.0 & 28.0 \\
    \cmidrule(lr){1-26}
    \multirow{2}{*}{Qwen3VL} & Text
      & \textbf{79.4$^{\ast}$} & 88.0$^{\ast}$ & 68.8 & 100.0 & 24.5 & 100.0 & 13.0 & 100.0 & 0.0 & 0.0 & 0.0 & 100.0
      & -- & -- & -- & -- & -- & -- & -- & -- & -- & -- & -- & -- \\
                              & JSON
      & \textbf{82.0} & 96.0 & 74.1 & 96.0 & 17.8 & 96.0 & 8.8 & 96.0 & 0.0 & 0.0 & 0.7 & 94.0
      & 74.1 & 96.0 & 64.8 & 94.0 & 64.7 & 90.0 & 67.3 & 82.0 & \textbf{75.4} & 90.0 & 69.5 & 92.0 \\
    \cmidrule(lr){1-26}
    \multirow{2}{*}{Opus 5.5} & Text
      & 82.0 & 100.0 & 84.3 & 100.0 & 78.0 & 100.0 & \textbf{87.1} & 100.0 & 86.0 & 100.0 & 83.2 & 100.0
      & -- & -- & -- & -- & -- & -- & -- & -- & -- & -- & -- & -- \\
                              & JSON
      & 86.3 & 100.0 & 83.5 & 100.0 & 55.4 & 100.0 & \textbf{88.0} & 100.0 & 72.5 & 100.0 & 80.4 & 100.0
      & \textbf{88.0} & 100.0 & 17.8 & 100.0 & 80.4 & 100.0 & 83.2 & 100.0 & 84.0 & 100.0 & 84.3 & 100.0 \\
    \cmidrule(lr){1-26}
    \multirow{2}{*}{GPT-6 Astra} & Text
      & 41.2 & 25.0 & 80.4 & 100.0 & 82.7 & 100.0 & \textbf{84.6} & 100.0 & 66.0 & 100.0 & 80.0 & 100.0
      & -- & -- & -- & -- & -- & -- & -- & -- & -- & -- & -- & -- \\
                              & JSON
      & 0.0 & 0.0 & 80.0 & 100.0 & 78.1 & 100.0 & 75.0 & 100.0 & 72.5 & 95.0 & \textbf{82.7} & 100.0
      & 80.8 & 100.0 & 78.8 & 100.0 & \textbf{82.7} & 100.0 & 77.8 & 100.0 & 80.4 & 100.0 & \textbf{82.7} & 100.0 \\
    \bottomrule
    \end{tabular}
    }
    \vspace{-2mm}
\caption{Format ablation on a Pascal subset for the two best open-source and proprietary models (all models in \Cref{tab:lmms_stage1_pascal,tab:stage_2_pascal}). Stage~1 (left) sweeps the bounding-box representation; Stage~2 (right) sweeps the JSON key name. The \texttt{--} column reports each model's unconstrained output parsed under its default scheme. Bold marks the highest F1 per row and stage among the shown columns. $^{\ast}$ marks text-mode outputs that were parsed as JSON (Qwen3VL: \texttt{xyxy}+\texttt{bbox\_2d}).}
    \label{tab:format_ablation_mini}
    \vspace{-4mm}
\end{table*}

Second, we evaluate the models' behavior across different bounding box formats. For this evaluation, we prompt the models to output detection results in specific bounding box formats, as well as without any instructions (see prompts in \Cref{app:prompts}). We show the detailed results for the two best performing open and closed source models in \Cref{tab:format_ablation_mini} and report the results of all other models in the appendix \Cref{tab:lmms_stage1_pascal} and \Cref{tab:stage_2_pascal}.
The preferred bbox for most models is \texttt{xyxy}, \texttt{yxyx} for Gemma-4 and Gemini 2.5 Flash, and \texttt{xywh} for GPT-5.4 and Sonnet 4.5 in JSON.
The frontier models show no strong preference.
Switching from the best to the second-best format collapses both mIoU and F1 on open-source models, and \texttt{cxcywh}, \texttt{all}, and \texttt{all-labelled} fail, with most F1 below 5.

The evaluation shows that the format instruction in the prompt does not override the convention each model has internalized.
E.g. when prompted for \texttt{cxcywh}, every open-source model scores near-zero F1 using the prompted \texttt{cxcywh} during parsing.
But, parsing the same outputs as the model's preferred corner format recovers 32--79 F1 (\Cref{app:cxcywh}). 
This indicates that most models specialize in one preferred format and will output this format independent of the given prompt instructions.  
Namely, open-source models and Gemini 2.5 Flash specialize to one preferred format, where these models reach high F1. 
Switching to a non-preferred format collapses the score (\Cref{tab:lmms_stage1_pascal}).
GPT-6 Astra and Opus 5.5 generalize across most formats.
GPT-5.4 is similarly robust. Its best result is still below Qwen3-VL and Gemini 2.5 Flash at their preferred formats, but averaged over all seven representations it reaches about twice the F1 of any open-source model (\Cref{app:crossformat}).

Second, while models can produce parsable output in the requested syntax, they might use coordinates that do not conform to that format.
As a result, many outputs have a format adherence near 100\% with a very low F1 score, 
showing that a good format adherence does not usually correlate with a good F1@0.5 score.
This point also motivates the proposed multi-stage probing to select each model's preferred format, so the per-task numbers fairly reflect the localization abilities of each model.



\vspace{-2mm}
\subsection{Output format}
\vspace{-2mm}
Finally, we assess the models' ability to adapt to different structured output formats, namely text and JSON. The results are again shown for the two best performing open source as well as closed source models in \Cref{tab:format_ablation_mini}, and we show the results of all other models in the appendix \Cref{tab:lmms_stage1_pascal} and \Cref{tab:stage_2_pascal}. All those experiments were done on a subset of Pascal VOC. Note that the JSON key used for Stage 1 is \texttt{bbox\_2d}, and that subsequent keys are tested using the bounding box format that yielded the best result with this key. 
It shows that, first, while most models can adhere to both formats when prompted, some actually show a drastic drop in performance between the two. Namely, on object detection, JSON beats text on the F1 score for most open-source models and Gemini 2.5 Flash. GLM-4.6V scores much higher in text, and Qwen3.5-Thinking and GPT-5.4 slightly higher. 
For GLM-4.6V, text is also the preferred output format when no format instructions are given.
The two frontier models perform on par in both formats.
Second, it shows that while most models are able to handle different JSON keys (Qwen3.5 only returns boxes under \texttt{bbox\_2d}), the preferred JSON key itself varies across models, with sometimes strong variations as in the case of Gemini. 
Opus 5.5 collapses under the key \texttt{box\_2d}, because this key changes its coordinate space (\Cref{app:quirks}).

\vspace{-2mm}
\section{Conclusion}
\vspace{-2mm}
In this work, we introduced a comprehensive visual detection benchmark for generalist multimodal LLMs tailored to promptable localization. In this benchmark, we assess the ability of current models to produce structured, spatially grounded outputs such as bounding boxes across four different task categories.
Our empirical study reveals that although most models exhibit strong localization, their performance remains highly sensitive to specific output formats. 
As such, our results point to the need for models that can reliably generate standardized, interpretable outputs and generalize across tasks and prompting conditions. Addressing these challenges is critical to improve multimodal reasoning as well as reliability, e.g., in the context of agentic systems.
We hope that this benchmark serves as a foundation for future research on promptable, generalist localization. 
\subsection*{Acknowledgments}
This work was supported by Toyota Motor Europe and Woven by Toyota.

\subsection*{AI use statement}

In this work, we used generative AI tools (Claude Code and Cursor) as coding assistants when implementing the benchmark.
We have not used generative AI tools to design the benchmark or the experiments, or to interpret results.
Generating synthetic data, theoretical models, mathematical claims and proofs, translation, and qualitative data analysis are not applicable to this work.
Additionally, we used the same tools for plotting and visualization.
We have reviewed all AI-assisted work, tested the AI-assisted code, and checked all plots against the results they show.
We take responsibility for the final content of this work, including text, claims or artifacts produced with the aid of generative AI.




\subsection*{Reproducibility statement}

We release the code of \benchmark{} in the anonymous repository linked in the abstract.
It contains the prompt generation, output parsing, metrics, and model wrappers, together with the 1,000-query subset of every dataset, sampled with a fixed seed (\Cref{subsec:tasks}).
It also lists the software versions, decoding settings, and commands used to run the format search and the main evaluation.
All datasets are publicly available.
\Cref{app:prompts} lists all prompts verbatim.
\Cref{sec:implementation} specifies the evaluated models, their reasoning settings, the coordinate space assumed for each model, and the metrics.
\Cref{app:sandbox} details the multi-stage format search and reports every configuration it evaluates, not only the selected one.
\Cref{app:resizing} documents the image preprocessing and the effective input resolution of each model, and \Cref{app:quirks} documents per-model parsing issues.
We access the proprietary models through OpenRouter.



\bibliography{iclr2027_conference}

@String(IJCV = {Int. J. Comput. Vis.})

@String(CVPR= {IEEE Conf. Comput. Vis. Pattern Recog.})

@String(ICCV= {Int. Conf. Comput. Vis.})

@String(ECCV= {Eur. Conf. Comput. Vis.})

@String(NIPS= {Adv. Neural Inform. Process. Syst.})

@String(ICLR = {Int. Conf. Learn. Represent.})

@String(IJCV  = {IJCV})

@String(CVPR  = {CVPR})

@String(ICCV  = {ICCV})

@String(ECCV  = {ECCV})

@String(NIPS  = {NeurIPS})

@String(ICLR  = {ICLR})

@misc{pascal,
	author = "Everingham, M. and Van~Gool, L. and Williams, C. K. I. and Winn, J. and Zisserman, A.",
	title = "The {PASCAL} {V}isual {O}bject {C}lasses {C}hallenge 2007 {(VOC2007)} {R}esults",
	howpublished = "http://www.pascal-network.org/challenges/VOC/voc2007/workshop/index.html",
    year ={2007}
}

@inproceedings{iground,
      title={Large-scale Pre-training for Grounded Video Caption Generation}, 
      author={Evangelos Kazakos and Cordelia Schmid and Josef Sivic},
      year={2025},
      booktitle=ICCV
}

@article{openimages,
   title={The Open Images Dataset V4: Unified Image Classification, Object Detection, and Visual Relationship Detection at Scale},
   author={Kuznetsova, Alina and Rom, Hassan and Alldrin, Neil and Uijlings, Jasper and Krasin, Ivan and Pont-Tuset, Jordi and Kamali, Shahab and Popov, Stefan and Malloci, Matteo and Kolesnikov, Alexander and Duerig, Tom and Ferrari, Vittorio},
   year={2020},
   journal=IJCV
}

@inproceedings{refcocog,
  title={Generation and Comprehension of Unambiguous Object Descriptions},
  author={Mao, Junhua and Huang, Jonathan and Toshev, Alexander and Camburu, Oana and Yuille, Alan and Murphy, Kevin},
  booktitle={CVPR},
  year={2016}
}

@inproceedings{referitgame,
    title = "{R}efer{I}t{G}ame: Referring to Objects in Photographs of Natural Scenes",
    author = "Kazemzadeh, Sahar and Ordonez, Vicente and Matten, Mark and Berg, Tamara",
    year={2014},
    booktitle={EMNLP}
}

@inproceedings{svg,
  author    = {Park, Jae Sung and Ma, Zixian and Li, Linjie and Zheng, Chenhao and Hsieh, Cheng-Yu and Lu, Ximing and Chandu, Khyathi and Kong, Quan and Kobori, Norimasa and Farhadi, Ali and Choi, Yejin and Krishna, Ranjay},
  title     = {Synthetic Visual Genome: Dense Scene Graphs at Scale with Multimodal Language Models},
  booktitle = CVPR,
  year      = {2025}
}

@misc{refl4,
      title={Revisiting Referring Expression Comprehension Evaluation in the Era of Large Multimodal Models},
      author={Jierun Chen and Fangyun Wei and Jinjing Zhao and Sizhe Song and Bohuai Wu and Zhuoxuan Peng and S. -H. Gary Chan and Hongyang Zhang},
      year={2024},
      url={https://arxiv.org/abs/2406.16866}, 
}

@inproceedings{phrasecut,
      title={PhraseCut: Language-based Image Segmentation in the Wild}, 
      author={Chenyun Wu and Zhe Lin and Scott Cohen and Trung Bui and Subhransu Maji},
      year={2020},
      booktitle=CVPR
}

@inproceedings{d3,
      title={Described Object Detection: Liberating Object Detection with Flexible Expressions}, 
      author={Chi Xie and Zhao Zhang and Yixuan Wu and Feng Zhu and Rui Zhao and Shuang Liang},
      year={2023},
      booktitle=NIPS
}

@inproceedings{flickr30kentities,
      title={Flickr30k Entities: Collecting Region-to-Phrase Correspondences for Richer Image-to-Sentence Models}, 
      author={Bryan A. Plummer and Liwei Wang and Chris M. Cervantes and Juan C. Caicedo and Julia Hockenmaier and Svetlana Lazebnik},
      year={2016},
      booktitle=ICCV
}

@inproceedings{hrinsdet,
      title={Solving Instance Detection from an Open-World Perspective}, 
      author={Qianqian Shen and Yunhan Zhao and Nahyun Kwon and Jeeeun Kim and Yanan Li and Shu Kong},
      year={2025},
      booktitle=CVPR
}

@inproceedings{robotools,     
	author={Li, Bowen and Wang, Jiashun and Hu, Yaoyu and Wang, Chen and Scherer, Sebastian},   
	booktitle=NIPS, 
	title={VoxDet: Voxel Learning for Novel Instance Detection},
	year={2023}
}

@article{qwen3vl,
      title={Qwen3-VL Technical Report}, 
      author={{Qwen3-VL}},
	  journal={arXiv preprint arXiv:2511.21631},
      year={2025}
}

@article{qwen2.5vl,
  title={Qwen2.5-VL Technical Report},
  author={{Qwen2.5-VL Team}},
  journal={arXiv preprint arXiv:2502.13923},
  year={2025}
}

@misc{qwen3.5,
    title = {Qwen3.5: Accelerating Productivity with Native Multimodal Agents},
    url = {https://qwen.ai/blog?id=qwen3.5},
    author = {{Qwen Team}},
    month = {February},
    year = {2026}
}

@misc{internvl3,
      title={InternVL3: Exploring Advanced Training and Test-Time Recipes for Open-Source Multimodal Models}, 
      author={{InternVL3 Team}},
      year={2025},
      url={https://arxiv.org/abs/2504.10479}, 
}

@misc{glm4.6v,
      title={GLM-4.5V and GLM-4.1V-Thinking: Towards Versatile Multimodal Reasoning with Scalable Reinforcement Learning}, 
      author={{GLM-V Team}},
      year={2025},
      url={https://arxiv.org/abs/2507.01006}, 
}

@misc{gemma4,
    title = {Gemma 4},
    author = {{Google DeepMind}},
    year = {2025},
    url = {https://deepmind.google/models/gemma/gemma-4/}
}

@misc{sonnet4.5,
    title = {Claude Sonnet 4.5},
    author = {{Anthropic}},
    year = {2025},
    url = {https://www.anthropic.com/news/claude-sonnet-4-5}
}

@misc{gpt6astra,
    title = {GPT-6 Astra},
    author = {{OpenAI}},
    year = {2026},
      url = {https://openai.com/index/gpt-6-astra/}
}

@misc{opus5.5,
    title = {Claude Opus 5.5},
    author = {{Anthropic}},
    year = {2026},
    url = {https://www.anthropic.com/news/claude-opus-5-5}
}

@misc{gpt5,
    title = {GPT-5},               
    author = {{OpenAI}},
    year = {2025},
    url = {https://openai.com/index/gpt-5-system-card/}
}

@article{gemini2.5,
    title   = {Gemini 2.5: Pushing the Frontier with Advanced Reasoning, Multimodality, Long Context, and Next Generation Agentic Capabilities},
    author  = {Comanici, Gheorghe and Bieber, Eric and Schaekermann, Mike and others},
    journal = {arXiv preprint arXiv:2507.06261},
    year    = {2025}
}

@inproceedings{refcoco_unc,
      title={Modeling Context in Referring Expressions}, 
      author={Licheng Yu and Patrick Poirson and Shan Yang and Alexander C. Berg and Tamara L. Berg},
      year={2016},
      booktitle=ECCV
}

@inproceedings{lvis,
      title={LVIS: A Dataset for Large Vocabulary Instance Segmentation}, 
      author={Agrim Gupta and Piotr Dollár and Ross Girshick},
      year={2019},
      booktitle=CVPR
}

@inproceedings{objects365,
      title={Objects365: A Large-scale, High-quality Dataset for Object Detection}, 
      author={Shuai Shao and Zeming Li and Tianyuan Zhang and Chao Peng and Gang Yu and Jing Li and Xiangyu Zhang and Jian Sun},
      year={2019},
      booktitle=ICCV
}

@article{ilsvrc,
Author = {Olga Russakovsky and Jia Deng and Hao Su and Jonathan Krause and Sanjeev Satheesh and Sean Ma and Zhiheng Huang and Andrej Karpathy and Aditya Khosla and Michael Bernstein and Alexander C. Berg and Li Fei-Fei},
Title = {{ImageNet Large Scale Visual Recognition Challenge}},
Year = {2015},
journal   = {IJCV},
}

@article{coco,
      title={Microsoft COCO: Common Objects in Context},
      author={Tsung-Yi Lin and Michael Maire and Serge Belongie and Lubomir Bourdev and Ross Girshick and James Hays and Pietro Perona and Deva Ramanan and C. Lawrence Zitnick and Piotr Dollár},
      year={2015},
      journal={arXiv preprint arxiv:1405.0312}, 
}

@article{vg,
  title={Visual Genome: Connecting Language and Vision Using Crowdsourced Dense Image Annotations},
  author={Ranjay Krishna and Yuke Zhu and Oliver Groth and Justin Johnson and Kenji Hata and Joshua Kravitz and Stephanie Chen and Yannis Kalantidis and Li-Jia Li and David A. Shamma and Michael S. Bernstein and Li Fei-Fei},
  journal={International Journal of Computer Vision},
  year={2017},
  volume={123},
  pages={32-73},
  url={https://doi.org/10.1007/s11263-016-0981-7},
  doi={10.1007/s11263-016-0981-7}
}

@inproceedings{perseg,
  title={Personalize Segment Anything Model with One Shot},
  author={Zhang, Renrui and Jiang, Zhengkai and Guo, Ziyu and Yan, Shilin and Pan, Junting and Dong, Hao and Gao, Peng and Li, Hongsheng},
  booktitle=ICLR,
  year={2024}
}

@inproceedings{pdm,
      title={Where's Waldo: Diffusion Features for Personalized Segmentation and Retrieval}, 
      author={Dvir Samuel and Rami Ben-Ari and Matan Levy and Nir Darshan and Gal Chechik},
      year={2024},
      booktitle=NIPS
}

@inproceedings{omnilabel,
      title={OmniLabel: A Challenging Benchmark for Language-Based Object Detection},
      author={Samuel Schulter and Vijay Kumar B G and Yumin Suh and Konstantinos M. Dafnis and Zhixing Zhang and Shiyu Zhao and Dimitris Metaxas},
      year={2023},
      booktitle={ICCV},
}

@inproceedings{rod,
author = {Yin, Heng and Ren, Yuqiang and Yan, Ke and Ding, Shouhong and Hao, Yongtao},
year = {2025},
title = {ROD-MLLM: Towards More Reliable Object Detection in Multimodal Large Language Models},
booktitle=CVPR
}

@InProceedings{mcbench,
    title={MC-Bench: A Benchmark for Multi-Context Visual Grounding in the Era of MLLMs},
    author={Xu, Yunqiu and Zhu, Linchao and Yang, Yi},
    year={2025},
    booktitle=ICCV
}

@inproceedings{hcrefloco,
title={A Large-Scale Human-Centric Benchmark for Referring Expression Comprehension in the {LMM} Era},
author={Fangyun Wei and Jinjing Zhao and Kun Yan and Hongyang Zhang and Chang Xu},
booktitle={The Thirty-eight Conference on Neural Information Processing Systems Datasets and Benchmarks Track},
year={2024},
}

@inproceedings{sorec,
      title={Referring Expression Comprehension for Small Objects}, 
      author={Kanoko Goto and Takumi Hirose and Mahiro Ukai and Shuhei Kurita and Nakamasa Inoue},
      year={2025},
      booktitle=ICCV
}

@inproceedings{knowdr,
      title={KnowDR-REC: Auditing Knowledge-Conditioned Visual Grounding in Referring Expression Comprehension},
      author={Guanghao Jin and Jingpei Wu and Tianpei Guo and Yiyi Niu and Weidong Zhou and Linyi Yang and Guoyang Liu},
      year={2026},
      booktitle={Findings of ACL},
}

@inproceedings{llava,
      title={Visual Instruction Tuning}, 
      author={Liu, Haotian and Li, Chunyuan and Wu, Qingyang and Lee, Yong Jae},
      booktitle={NeurIPS},
      year={2023},
}

@inproceedings{molmo,
  title={Molmo and PixMo: Open Weights and Open Data for State-of-the-Art Vision-Language Models},
  author={{Molmo and PixMo Team}},
  booktitle=CVPR,
  year={2025}
}

@inproceedings{cogvlm,
      title={CogVLM: Visual Expert for Pretrained Language Models},
      author={Weihan Wang and Qingsong Lv and Wenmeng Yu and Wenyi Hong and Ji Qi and Yan Wang and Junhui Ji and Zhuoyi Yang and Lei Zhao and Xixuan Song and Jiazheng Xu and Bin Xu and Juanzi Li and Yuxiao Dong and Ming Ding and Jie Tang},
      year={2024},
      booktitle=NIPS,
}

@misc{llavanext,
    title={LLaVA-NeXT: Improved reasoning, OCR, and world knowledge},
    url={https://llava-vl.github.io/blog/2024-01-30-llava-next/},
    author={Liu, Haotian and Li, Chunyuan and Li, Yuheng and Li, Bo and Zhang, Yuanhan and Shen, Sheng and Lee, Yong Jae},
    year={2024}
}

@inproceedings{improvedllava,
      title={Improved Baselines with Visual Instruction Tuning},
      author={Liu, Haotian and Li, Chunyuan and Li, Yuheng and Lee, Yong Jae},
      booktitle=CVPR,
      year={2024},
}

@inproceedings{ifbench,
      title={Generalizing Verifiable Instruction Following}, 
      author={Valentina Pyatkin and Saumya Malik and Victoria Graf and Hamish Ivison and Shengyi Huang and Pradeep Dasigi and Nathan Lambert and Hannaneh Hajishirzi},
      year={2025},
      booktitle=NIPS
}

@misc{sapkota2026ultralyticsyoloevolutionoverview,
      title={Ultralytics YOLO Evolution: An Overview of YOLO26, YOLO11, YOLOv8 and YOLOv5 Object Detectors for Computer Vision and Pattern Recognition}, 
      author={Ranjan Sapkota and Manoj Karkee},
      year={2026},
      eprint={2510.09653},
      archivePrefix={arXiv},
      primaryClass={cs.CV},
      url={https://arxiv.org/abs/2510.09653}, 
}

@InProceedings{Carion2020DETR,
author="Carion, Nicolas
and Massa, Francisco
and Synnaeve, Gabriel
and Usunier, Nicolas
and Kirillov, Alexander
and Zagoruyko, Sergey",
title="End-to-End Object Detection with Transformers",
booktitle=ECCV,
year={2020}
}

@InProceedings{Robinson2026rf-detr,
    title={RF-DETR: Neural Architecture Search for Real-Time Detection Transformers},
    author={Isaac Robinson and Peter Robicheaux and Matvei Popov and Deva Ramanan and Neehar Peri},
    booktitle=ICLR,
    year={2026}
}

@InProceedings{huang2026ledetrrevisitingrealtimedetection,
      title={Le-DETR: Revisiting Real-Time Detection Transformer with Efficient Encoder Design}, 
      author={Jiannan Huang and Aditya Kane and Fengzhe Zhou and Yunchao Wei and Humphrey Shi},
      year={2026},
      booktitle={CVPR Findings}
}

@InProceedings{huang2024deim,
      title={DEIM: DETR with Improved Matching for Fast Convergence},
      author={Shihua, Huang and Zhichao, Lu and Xiaodong, Cun and Yongjun, Yu and Xiao, Zhou and Xi, Shen},
      booktitle=CVPR,
      year={2025},
}

@inproceedings{zhang2026vlm4vlarevisitingvisionlanguagemodelsvisionlanguageaction,
      title={VLM4VLA: Revisiting Vision-Language-Models in Vision-Language-Action Models},
      author={Jianke Zhang and Xiaoyu Chen and Yanjiang Guo and Yucheng Hu and Jianyu Chen},
      year={2026},
      booktitle=ICLR
}

@INPROCEEDINGS{Liao2025Can,
  author={Liao, Yuan-Hong and Mahmood, Rafid and Fidler, Sanja and Acuna, David},
  booktitle=CVPR, 
  title={Can Large Vision-Language Models Correct Semantic Grounding Errors By Themselves?}, 
  year={2025}
}

@inproceedings{dota,
      title={DOTA: A Large-scale Dataset for Object Detection in Aerial Images}, 
      author={Gui-Song Xia and Xiang Bai and Jian Ding and Zhen Zhu and Serge Belongie and Jiebo Luo and Mihai Datcu and Marcello Pelillo and Liangpei Zhang},
      year={2018},
      booktitle=CVPR,
}
\bibliographystyle{iclr2027_conference}


\appendix

\section{Appendix}

\subsection{Overview}

This appendix collects supplementary material referenced from the main paper.
\Cref{app:sandbox} describes the multi-stage format search that selects each model's bounding-box representation, JSON key, and output mode used in \Cref{sec:results}.
\Cref{app:robustness} tests whether the representation selected on Pascal transfers to other tasks, compares cross-representation performance, and measures robustness to prompt wording and to stricter IoU thresholds.
\Cref{app:cxcywh} probes the center-format failure with an explicit definition of the four numbers, and with a corner-coordinate reinterpretation of the response.
\Cref{app:captions} tests whether the grounded-captioning origin of iGround and Flickr30k Entities biases their use as detection benchmarks.
\Cref{app:refcoco} decomposes the \emph{RefCOCO-Avg} column of \Cref{tab:refexp} into per-split numbers for RefCOCO, RefCOCO+, and RefCOCO-g.
\Cref{app:resizing} documents preprocessing for the HR-InsDet dataset, and per-model evaluation quirks.
\Cref{app:prompts} lists the prompts that \benchmark{} sends to the model, verbatim.

\subsection{Multi-stage Format Search}
\label{app:sandbox}
MLLMs are sensitive to the bounding-box representation and the output format.
A model's score can vary by more than 30 mIoU across formats.
For example, Qwen3-VL Pascal (subset) mIoU drops from 60.9 (\texttt{xyxy}) to 24.4 (\texttt{xywh}) in JSON mode (\Cref{tab:lmms_stage1_pascal}).
We therefore run a per-model probe before the full evaluation to find the format that elicits each model's best score.
The probe runs in three stages.
Stages 1 and 2 run on Pascal.
Stage 1 sweeps the bounding-box representation for each output mode.
Stage 2 fixes the bounding-box representation at Stage 1's JSON-mode winner and sweeps the JSON key.
Stage 3 takes the Stage 1 text winner, the Stage 2 JSON winner, and the unconstrained prompt, runs each at the full 1,000 queries per evaluation split, and selects the configuration with the highest average F1@0.5 per task family.
\Cref{sec:results} reports the Stage 3 winner for each (model, dataset) pair.
We recommend the same probe before evaluating a new model on \benchmark{}, especially when the model's preferred format is not known.
The sweep covers formats observed across the models we tested, and can be extended with new ones.
Each combination in Stages 1 and 2 is run on 50 queries for open-source models and 20 for proprietary models, to limit API cost.

\begin{table*}[!ht]
    \centering
    \resizebox{\textwidth}{!}{
    \footnotesize
    \setlength{\tabcolsep}{3pt}
    \begin{tabular}{ll ccc | ccc ccc ccc ccc ccc ccc ccc ccc}
    \toprule
     & & \multicolumn{3}{c|}{\texttt{--}} & \multicolumn{3}{c}{\texttt{xyxy}} & \multicolumn{3}{c}{\texttt{xywh}} & \multicolumn{3}{c}{\texttt{yxyx}} & \multicolumn{3}{c}{\texttt{yxhw}} & \multicolumn{3}{c}{\texttt{all}} & \multicolumn{3}{c}{\texttt{all-labelled}} & \multicolumn{3}{c}{\texttt{cxcywh}} & \multicolumn{3}{c}{\texttt{cxcywh}+\,\emph{def}} \\
    \cmidrule(lr){3-5} \cmidrule(lr){6-8} \cmidrule(lr){9-11} \cmidrule(lr){12-14} \cmidrule(lr){15-17} \cmidrule(lr){18-20} \cmidrule(lr){21-23} \cmidrule(lr){24-26} \cmidrule(lr){27-29}
    Model & Output & F & m & A & F & m & A & F & m & A & F & m & A & F & m & A & F & m & A & F & m & A & F & m & A & F & m & A \\
    \midrule
    \rowcolor{tablebanner} \multicolumn{29}{l}{\textit{Single-label}} \\
    \multirow{2}{*}{Qwen2.5VL} & Text & 63.9* & 47.2 & 96.0 & 40.6 & 24.4 & 62.0 & 18.0 & 9.8 & 40.0 & 10.8 & 7.6 & 56.0 & 4.9 & 5.2 & 40.0 & 0.0 & 0.0 & 0.0 & 0.0 & 0.0 & 0.0 & 0.0 & 3.5 & 46.0 & 0.0 & 8.5 & 94.0 \\
                                       & JSON & 67.5 & 48.6 & 98.0 & 64.3 & 46.9 & 90.0 & 20.3 & 16.2 & 80.0 & 9.6 & 10.8 & 86.0 & 7.0 & 7.9 & 76.0 & 0.0 & 0.0 & 0.0 & 0.0 & 0.0 & 0.0 & 0.0 & 5.7 & 78.0 & 0.0 & 6.4 & 84.0 \\
    \cmidrule(lr){1-29}
    \multirow{2}{*}{Qwen3VL} & Text & 67.9 & 46.6 & 100.0 & 69.9 & 49.1 & 100.0 & 19.5 & 20.0 & 100.0 & 13.6 & 14.7 & 100.0 & 8.5 & 11.3 & 100.0 & 0.0 & 0.0 & 6.0 & 0.0 & 0.0 & 6.0 & 0.0 & 8.5 & 100.0 & 0.0 & 8.2 & 100.0 \\
                                       & JSON & 75.5 & 66.0 & 98.0 & 74.6 & 60.9 & 96.0 & 17.2 & 24.4 & 98.0 & 12.2 & 17.4 & 96.0 & 6.7 & 13.1 & 98.0 & 0.0 & 0.0 & 0.0 & 0.0 & 0.0 & 0.0 & 0.0 & 10.6 & 96.0 & 0.0 & 11.1 & 98.0 \\
    \cmidrule(lr){1-29}
    \multirow{2}{*}{Qwen3.5} & Text & 77.5* & 70.8 & 100.0 & 64.8 & 57.1 & 100.0 & 16.5 & 24.0 & 100.0 & 8.6 & 17.2 & 100.0 & 8.3 & 13.5 & 98.0 & 0.0 & 0.0 & 2.0 & 0.0 & 0.0 & 2.0 & 0.0 & 11.5 & 100.0 & 0.0 & 11.2 & 100.0 \\
                                       & JSON & 77.1 & 71.4 & 100.0 & 77.7 & 71.1 & 100.0 & 17.4 & 25.5 & 100.0 & 12.6 & 19.3 & 100.0 & 7.3 & 14.2 & 100.0 & 0.0 & 0.0 & 0.0 & 0.0 & 0.0 & 0.0 & 0.0 & 11.3 & 100.0 & 0.0 & 11.5 & 98.0 \\
    \cmidrule(lr){1-29}
    \multirow{2}{*}{\makecell[l]{Qwen3.5 \\ - thinking}} & Text & 73.6 & 57.3 & 98.0 & 77.7 & 62.4 & 100.0 & 21.1 & 23.2 & 94.0 & 13.5 & 17.4 & 100.0 & 10.2 & 12.0 & 88.0 & 0.0 & 0.0 & 2.0 & 0.0 & 0.0 & 0.0 & 0.0 & 9.8 & 92.0 & 20.2 & 20.3 & 84.0 \\
                                       & JSON & 81.3 & 60.9 & 94.0 & 76.7 & 57.6 & 92.0 & 20.8 & 23.7 & 94.0 & 12.0 & 17.3 & 92.0 & 9.3 & 12.8 & 88.0 & 0.0 & 0.0 & 0.0 & 0.0 & 0.0 & 0.0 & 0.0 & 9.7 & 92.0 & 0.0 & 8.6 & 88.0 \\
    \cmidrule(lr){1-29}
    \multirow{2}{*}{InternVL3} & Text & 23.7 & 17.8 & 100.0 & 79.8 & 64.4 & 100.0 & 17.9 & 24.7 & 100.0 & 11.4 & 13.9 & 92.0 & 4.6 & 6.1 & 90.0 & 0.0 & 0.0 & 4.0 & 0.0 & 0.0 & 2.0 & 0.0 & 10.3 & 100.0 & 0.0 & 11.7 & 98.0 \\
                                       & JSON & 0.0 & 0.0 & 2.0 & 80.2 & 71.5 & 98.0 & 20.8 & 26.1 & 100.0 & 7.2 & 15.9 & 98.0 & 8.5 & 14.0 & 100.0 & 0.0 & 0.0 & 0.0 & 0.0 & 0.0 & 20.0 & 0.0 & 12.3 & 100.0 & 0.0 & 12.0 & 100.0 \\
    \cmidrule(lr){1-29}
    \multirow{2}{*}{Gemma-4} & Text & 64.0* & 46.0 & 100.0 & 9.9 & 15.7 & 100.0 & 6.7 & 11.8 & 100.0 & 73.0 & 55.7 & 100.0 & 14.4 & 18.6 & 100.0 & 0.0 & 0.0 & 32.0 & 0.0 & 0.0 & 32.0 & 5.6 & 11.7 & 100.0 & 10.3 & 15.3 & 100.0 \\
                                       & JSON & 64.3 & 45.6 & 100.0 & 5.6 & 6.3 & 74.0 & 4.2 & 6.9 & 88.0 & 43.4 & 26.5 & 68.0 & 7.5 & 9.5 & 82.0 & 0.0 & 0.0 & 26.0 & 0.0 & 0.0 & 20.0 & 0.0 & 4.6 & 98.0 & 0.0 & 4.8 & 100.0 \\
    \cmidrule(lr){1-29}
    \multirow{2}{*}{GLM4.6V} & Text & 79.4 & 67.7 & 98.0 & 78.8 & 66.4 & 98.0 & 16.7 & 23.9 & 100.0 & 14.7 & 18.2 & 100.0 & 7.4 & 12.9 & 98.0 & 0.0 & 0.0 & 0.0 & 0.0 & 0.0 & 2.0 & 0.0 & 10.7 & 98.0 & 0.0 & 12.2 & 100.0 \\
                                       & JSON & 0.0 & 0.0 & 0.0 & 44.1 & 27.3 & 16.0 & 4.0 & 6.7 & 20.0 & 1.4 & 4.3 & 20.0 & 0.0 & 2.8 & 22.0 & 0.0 & 0.0 & 0.0 & 0.0 & 0.0 & 0.0 & 0.0 & 3.0 & 22.0 & 0.0 & 2.0 & 14.0 \\
    \cmidrule(lr){1-29}
    \multirow{2}{*}{GPT5.4} & Text & 36.0 & 38.9 & 100.0 & 42.1 & 41.5 & 100.0 & 35.6 & 42.8 & 100.0 & 34.0 & 42.5 & 100.0 & 32.4 & 40.8 & 100.0 & 38.8 & 38.1 & 100.0 & 43.3 & 41.3 & 100.0 & 28.9 & 35.3 & 100.0 & 41.2 & 40.7 & 100.0 \\
                                       & JSON & 0.0 & 0.0 & 0.0 & 37.6 & 39.1 & 100.0 & 40.4 & 44.7 & 100.0 & 37.5 & 38.2 & 100.0 & 32.3 & 39.9 & 100.0 & 11.4 & 8.9 & 70.0 & 34.3 & 37.3 & 100.0 & 38.8 & 36.6 & 100.0 & 35.0 & 41.3 & 100.0 \\
    \cmidrule(lr){1-29}
    \multirow{2}{*}{Gemini 2.5 Flash} & Text & 76.6* & 73.6 & 95.0 & 5.7 & 14.7 & 100.0 & 22.5 & 37.0 & 100.0 & 37.7 & 55.7 & 100.0 & 24.6 & 40.5 & 100.0 & 5.1 & 6.7 & 85.0 & 8.6 & 17.0 & 100.0 & 31.4 & 38.8 & 95.0 & 20.5 & 31.9 & 95.0 \\
                                       & JSON & 81.1 & 81.3 & 100.0 & 5.9 & 13.5 & 80.0 & 5.8 & 11.3 & 85.0 & 76.1 & 63.2 & 80.0 & 11.9 & 16.5 & 75.0 & 0.0 & 0.0 & 0.0 & 3.3 & 1.7 & 10.0 & 2.2 & 8.4 & 75.0 & 6.0 & 10.8 & 70.0 \\
    \cmidrule(lr){1-29}
    \multirow{2}{*}{Sonnet 4.5} & Text & 7.8 & 9.3 & 55.0 & 30.8 & 28.0 & 100.0 & 16.9 & 27.1 & 100.0 & 11.4 & 13.6 & 100.0 & 14.1 & 14.8 & 100.0 & 0.0 & 0.0 & 0.0 & 17.9 & 15.3 & 80.0 & 1.8 & 2.2 & 100.0 & 1.6 & 1.7 & 100.0 \\
                                       & JSON & 30.6 & 25.0 & 85.0 & 26.7 & 32.6 & 100.0 & 27.4 & 37.6 & 100.0 & 15.5 & 27.4 & 100.0 & 18.3 & 25.0 & 100.0 & 12.0 & 15.1 & 75.0 & 24.0 & 33.4 & 100.0 & 10.8 & 23.8 & 95.0 & 21.4 & 30.4 & 100.0 \\
    \cmidrule(lr){1-29}
    \multirow{2}{*}{GPT-6 Astra} & Text & 75.4 & 78.9 & 100.0 & 75.4 & 78.1 & 100.0 & 76.8 & 78.4 & 100.0 & 75.4 & 78.5 & 100.0 & 66.1 & 76.9 & 100.0 & 71.9 & 75.1 & 95.0 & 62.0 & 58.4 & 95.0 & 75.4 & 78.1 & 100.0 & 76.1 & 76.9 & 100.0 \\
                                       & JSON & 0.0 & 0.0 & 0.0 & 75.2 & 78.1 & 100.0 & 76.9 & 80.8 & 100.0 & 60.6 & 60.5 & 100.0 & 69.0 & 75.1 & 100.0 & 71.3 & 75.6 & 95.0 & 62.4 & 60.0 & 90.0 & 76.5 & 78.7 & 100.0 & 76.5 & 79.3 & 100.0 \\
    \cmidrule(lr){1-29}
    \multirow{2}{*}{Opus 5.5} & Text & 83.9 & 84.8 & 100.0 & 84.7 & 85.5 & 100.0 & 83.2 & 84.6 & 100.0 & 36.5 & 40.4 & 100.0 & 82.1 & 83.4 & 100.0 & 81.0 & 85.5 & 100.0 & 83.2 & 84.1 & 100.0 & 75.4 & 76.7 & 100.0 & 81.7 & 84.1 & 100.0 \\
                                       & JSON & 82.5 & 83.9 & 100.0 & 80.3 & 85.7 & 100.0 & 83.2 & 84.2 & 100.0 & 82.5 & 83.4 & 100.0 & 83.2 & 82.8 & 100.0 & 80.0 & 83.9 & 100.0 & 73.7 & 76.1 & 100.0 & 81.7 & 84.1 & 100.0 & 80.0 & 80.6 & 100.0 \\
    \midrule
    \rowcolor{tablebanner} \multicolumn{29}{l}{\textit{Multi-label}} \\
    \multirow{2}{*}{Qwen2.5VL} & Text & 82.1* & 69.7 & 100.0 & 20.8 & 59.3 & 94.0 & 9.3 & 21.7 & 90.0 & 6.8 & 13.7 & 94.0 & 4.6 & 11.5 & 92.0 & 2.8 & 1.0 & 8.0 & 1.4 & 0.4 & 8.0 & 0.0 & 10.4 & 90.0 & 0.0 & 8.0 & 62.0 \\
                                       & JSON & 81.9 & 72.0 & 100.0 & 66.1 & 51.9 & 86.0 & 19.4 & 21.5 & 86.0 & 11.3 & 12.7 & 86.0 & 8.7 & 10.6 & 88.0 & 0.0 & 0.0 & 0.0 & 0.0 & 0.0 & 0.0 & 0.0 & 8.0 & 86.0 & 0.0 & 8.0 & 88.0 \\
    \cmidrule(lr){1-29}
    \multirow{2}{*}{Qwen3VL} & Text & 79.4* & 68.5 & 88.0 & 68.8 & 49.0 & 100.0 & 24.5 & 22.5 & 100.0 & 20.7 & 16.6 & 100.0 & 13.0 & 13.3 & 100.0 & 0.0 & 0.0 & 0.0 & 0.0 & 0.0 & 0.0 & 0.0 & 7.2 & 100.0 & 0.0 & 7.0 & 100.0 \\
                                       & JSON & 82.0 & 73.5 & 96.0 & 74.1 & 65.5 & 96.0 & 17.8 & 24.3 & 96.0 & 16.1 & 19.5 & 98.0 & 8.8 & 13.6 & 96.0 & 0.0 & 0.0 & 0.0 & 0.0 & 0.0 & 0.0 & 0.0 & 10.3 & 98.0 & 0.7 & 9.6 & 94.0 \\
    \cmidrule(lr){1-29}
    \multirow{2}{*}{Qwen3.5} & Text & 82.8* & 74.6 & 98.0 & 25.3 & 36.6 & 64.0 & 12.6 & 10.9 & 52.0 & 13.1 & 12.1 & 66.0 & 6.6 & 8.9 & 56.0 & 0.0 & 0.0 & 0.0 & 0.0 & 0.0 & 0.0 & 0.0 & 6.2 & 64.0 & 0.0 & 6.7 & 70.0 \\
                                       & JSON & 84.3 & 73.7 & 96.0 & 80.0 & 74.0 & 98.0 & 18.3 & 27.1 & 100.0 & 16.3 & 20.5 & 100.0 & 9.6 & 15.4 & 98.0 & 0.0 & 0.0 & 0.0 & 0.0 & 0.0 & 0.0 & 0.0 & 11.2 & 100.0 & 0.0 & 11.3 & 100.0 \\
    \cmidrule(lr){1-29}
    \multirow{2}{*}{\makecell[l]{Qwen3.5 \\ - thinking}} & Text & 42.1* & 26.6 & 30.0 & 78.0 & 63.6 & 94.0 & 19.7 & 24.3 & 94.0 & 19.5 & 19.0 & 98.0 & 11.3 & 12.7 & 86.0 & 0.0 & 0.0 & 0.0 & 0.0 & 0.0 & 0.0 & 1.7 & 10.0 & 90.0 & 10.9 & 13.0 & 72.0 \\
                                       & JSON & 76.2 & 62.5 & 92.0 & 75.0 & 60.1 & 90.0 & 19.5 & 22.8 & 92.0 & 18.6 & 17.9 & 92.0 & 11.3 & 12.7 & 86.0 & 0.0 & 0.0 & 0.0 & 0.0 & 0.0 & 0.0 & 0.0 & 8.6 & 88.0 & 0.0 & 8.4 & 82.0 \\
    \cmidrule(lr){1-29}
    \multirow{2}{*}{InternVL3} & Text & 18.4 & 19.3 & 52.0 & 77.0 & 60.4 & 100.0 & 22.3 & 23.7 & 100.0 & 18.4 & 17.6 & 100.0 & 12.3 & 13.9 & 100.0 & 0.0 & 0.0 & 0.0 & 0.0 & 0.0 & 0.0 & 0.0 & 9.1 & 100.0 & 0.0 & 9.4 & 100.0 \\
                                       & JSON & 0.0 & 0.0 & 0.0 & 82.5 & 67.8 & 96.0 & 21.6 & 26.0 & 96.0 & 19.1 & 20.5 & 98.0 & 11.3 & 15.6 & 94.0 & 0.0 & 0.0 & 2.0 & 0.0 & 0.0 & 0.0 & 0.7 & 12.0 & 100.0 & 0.0 & 11.7 & 98.0 \\
    \cmidrule(lr){1-29}
    \multirow{2}{*}{Gemma-4} & Text & 74.2* & 62.6 & 100.0 & 18.0 & 19.5 & 100.0 & 13.2 & 16.0 & 100.0 & 71.8 & 55.5 & 100.0 & 21.9 & 24.9 & 100.0 & 0.0 & 0.0 & 0.0 & 0.0 & 0.0 & 0.0 & 9.3 & 12.9 & 100.0 & 14.9 & 16.7 & 100.0 \\
                                       & JSON & 76.6 & 63.4 & 100.0 & 6.5 & 3.5 & 28.0 & 2.7 & 2.1 & 20.0 & 17.0 & 7.8 & 26.0 & 1.4 & 0.8 & 6.0 & 0.0 & 0.0 & 0.0 & 0.0 & 0.0 & 0.0 & 0.0 & 0.0 & 2.0 & 0.0 & 0.3 & 6.0 \\
    \cmidrule(lr){1-29}
    \multirow{2}{*}{GLM4.6V} & Text & 0.0 & 0.0 & 0.0 & 80.3 & 66.2 & 96.0 & 19.0 & 23.4 & 88.0 & 15.9 & 18.2 & 88.0 & 6.8 & 12.8 & 86.0 & 0.0 & 0.0 & 0.0 & 0.0 & 0.0 & 16.0 & 0.0 & 9.8 & 86.0 & 0.8 & 13.7 & 98.0 \\
                                       & JSON & 0.0 & 0.0 & 0.0 & 63.7 & 46.1 & 44.0 & 12.6 & 15.3 & 48.0 & 10.3 & 10.4 & 42.0 & 1.8 & 5.9 & 40.0 & 0.0 & 0.0 & 0.0 & 0.0 & 0.0 & 0.0 & 0.0 & 7.0 & 46.0 & 0.0 & 7.3 & 42.0 \\
    \cmidrule(lr){1-29}
    \multirow{2}{*}{GPT5.4} & Text & 15.4 & 7.9 & 20.0 & 54.5 & 46.5 & 100.0 & 51.1 & 50.3 & 100.0 & 54.5 & 48.7 & 100.0 & 54.9 & 48.5 & 100.0 & 54.1 & 44.0 & 100.0 & 56.8 & 48.6 & 95.0 & 47.7 & 43.8 & 100.0 & 43.5 & 39.6 & 100.0 \\
                                       & JSON & 0.0 & 0.0 & 0.0 & 47.9 & 49.8 & 100.0 & 41.2 & 44.9 & 100.0 & 40.4 & 42.6 & 100.0 & 37.5 & 43.2 & 100.0 & 14.0 & 11.8 & 85.0 & 47.4 & 43.3 & 95.0 & 44.7 & 43.5 & 100.0 & 50.5 & 46.8 & 100.0 \\
    \cmidrule(lr){1-29}
    \multirow{2}{*}{Gemini 2.5 Flash} & Text & 93.6* & 83.6 & 100.0 & 26.2 & 35.6 & 100.0 & 20.3 & 31.1 & 100.0 & 49.6 & 61.9 & 100.0 & 65.3 & 55.3 & 100.0 & 24.2 & 24.5 & 95.0 & 38.0 & 35.0 & 100.0 & 46.8 & 48.2 & 100.0 & 46.2 & 44.7 & 100.0 \\
                                       & JSON & 93.6 & 83.2 & 100.0 & 22.2 & 14.0 & 65.0 & 15.6 & 13.9 & 65.0 & 45.9 & 28.6 & 65.0 & 27.7 & 21.6 & 65.0 & 0.0 & 0.0 & 5.0 & 0.0 & 0.0 & 25.0 & 22.6 & 12.9 & 60.0 & 13.1 & 9.2 & 45.0 \\
    \cmidrule(lr){1-29}
    \multirow{2}{*}{Sonnet 4.5} & Text & 15.4 & 10.2 & 30.0 & 31.3 & 32.4 & 100.0 & 31.3 & 32.9 & 100.0 & 33.3 & 29.6 & 100.0 & 24.4 & 24.1 & 100.0 & 0.0 & 0.0 & 5.0 & 32.1 & 29.2 & 95.0 & 2.4 & 1.7 & 100.0 & 2.3 & 3.3 & 100.0 \\
                                       & JSON & 16.2 & 19.5 & 60.0 & 26.8 & 30.1 & 100.0 & 31.7 & 30.9 & 100.0 & 32.1 & 26.5 & 100.0 & 31.3 & 31.1 & 100.0 & 14.1 & 16.5 & 50.0 & 36.6 & 33.0 & 100.0 & 27.9 & 31.2 & 100.0 & 28.9 & 28.1 & 100.0 \\
    \cmidrule(lr){1-29}
    \multirow{2}{*}{GPT-6 Astra} & Text & 41.2 & 24.5 & 25.0 & 80.4 & 79.4 & 100.0 & 82.7 & 81.5 & 100.0 & 66.0 & 69.2 & 100.0 & 84.6 & 81.7 & 100.0 & 79.6 & 80.0 & 100.0 & 66.0 & 66.3 & 100.0 & 81.9 & 82.4 & 100.0 & 80.0 & 81.2 & 100.0 \\
                                       & JSON & 0.0 & 0.0 & 0.0 & 80.0 & 83.0 & 100.0 & 78.1 & 81.7 & 100.0 & 63.5 & 61.9 & 100.0 & 75.0 & 80.2 & 100.0 & 83.5 & 82.4 & 100.0 & 72.5 & 72.2 & 95.0 & 80.8 & 80.5 & 100.0 & 82.7 & 81.3 & 100.0 \\
    \cmidrule(lr){1-29}
    \multirow{2}{*}{Opus 5.5} & Text & 82.0 & 81.4 & 100.0 & 84.3 & 85.9 & 100.0 & 78.0 & 77.3 & 100.0 & 43.1 & 45.4 & 100.0 & 87.1 & 87.3 & 100.0 & 84.3 & 84.8 & 100.0 & 86.0 & 85.1 & 100.0 & 46.0 & 44.7 & 100.0 & 83.2 & 80.0 & 100.0 \\
                                       & JSON & 86.3 & 86.2 & 100.0 & 83.5 & 84.7 & 100.0 & 55.4 & 60.0 & 100.0 & 68.0 & 66.5 & 100.0 & 88.0 & 87.0 & 100.0 & 65.3 & 66.4 & 100.0 & 72.5 & 72.4 & 100.0 & 84.3 & 81.9 & 100.0 & 80.4 & 79.8 & 100.0 \\
    \bottomrule
    \end{tabular}
    }
    \caption{Stage 1 sweep on Pascal.
    Each representation column groups three sub-columns: F (F1@0.5), m (mIoU), A (FA, \%).
    Single-label rows above the midrule, multi-label rows below.
    The \texttt{--} column shows the model's natural output parsed under its default scheme (\Cref{tab:preferred_format}).
    * marks unconstrained text cells scored with the JSON parser (the parser of each cell is listed in \Cref{tab:preferred_format}). }
    \label{tab:lmms_stage1_pascal}
\end{table*}

\textbf{Stage 1: bounding-box representation.}
We evaluate each bounding-box representation from \Cref{subsec:bbox} for both text and JSON output, plus the \texttt{cxcywh} sub-variants described in \Cref{app:cxcywh}.
The \texttt{--} column of \Cref{tab:lmms_stage1_pascal} reports the unconstrained prompt. 
For unconstrained text, the parser extracts four numbers from the response and assigns them according to the model's preferred bounding box representation (see \Cref{tab:preferred_format}).
An asterisk on the \texttt{--} cell marks cells where we report the JSON parse of the unconstrained text response instead of the text parse (\Cref{tab:preferred_format}).
For example, Qwen2.5-VL and Qwen3.5 output JSON with key \texttt{bbox\_2d} and \texttt{xyxy} coordinates under an unconstrained text prompt for both single-label and multi-label queries, while Qwen3-VL does so only for multi-label.
The reparse matters most in multi-label queries, where the regex pairs each class name with the boxes that follow it in the text.
Any other ordering misaligns labels and boxes.
The boxes still overlap with the ground truth, so mIoU stays high while F1 collapses to near zero.
Reparsing as JSON fixes the misalignment, recovering F1 from 0.0 to 79.4 for Qwen3-VL and from 0.0 to 74.2 for Gemma-4.

For unconstrained JSON, we report the JSON key used to parse in~\cref{tab:preferred_format}.
The Qwen family returns a list of four coordinates with the key \texttt{bbox\_2d}, Gemma-4 and Gemini 2.5 Flash with \texttt{box\_2d}, and Sonnet 4.5 returns a JSON object with four corner keys (\texttt{xmin}, \texttt{ymin}, \texttt{xmax}, \texttt{ymax}).
Opus 5.5 returns \texttt{bbox\_2d} with \texttt{xyxy} pixel coordinates, while GPT-6 Astra, like GPT-5.4, has no parsable default (it returns a \texttt{boxes} key or a plain list).

On \texttt{all} and \texttt{all-labelled}, responses contain four coordinates rather than the eight required, so format adherence is low across the open-source models.
Labelling the corners as $x_{\min}, y_{\min}, \ldots, x_{\min}, y_{\max}$ (\texttt{all-labelled}) raises format adherence for the closed-source models, with Sonnet 4.5 going from 0\% to 80\% in text, GPT-5.4 from 70\% to 100\% in JSON, and Gemini 2.5 Flash from 85\% to 100\% in text.
Among the open-source models, the same effect appears for InternVL3 JSON (from 0\% on \texttt{all} to 20\% on \texttt{all-labelled}) and GLM-4.6V text (from 0\% to 16\% multi-label).
Open-source models stay below 11 F1 on \texttt{all}, \texttt{all-labelled}, and \texttt{cxcywh}, except Qwen3.5-Thinking (20.2) and Gemma-4 (14.9) with the \texttt{cxcywh + definition} text prompt.

\begin{table*}[!ht]
    \centering
    \resizebox{\textwidth}{!}{
    \footnotesize
    \setlength{\tabcolsep}{3pt}
    \begin{tabular}{ll  ccc ccc ccc ccc ccc ccc | l}
    \toprule
     & & \multicolumn{3}{c}{\texttt{bbox\_2d}} & \multicolumn{3}{c}{\texttt{box\_2d}} & \multicolumn{3}{c}{\texttt{bbox}} & \multicolumn{3}{c}{\texttt{bounding\_box}} & \multicolumn{3}{c}{\texttt{coordinates}} & \multicolumn{3}{c}{\texttt{class\_name}} & \\
    \cmidrule(lr){3-5} \cmidrule(lr){6-8} \cmidrule(lr){9-11} \cmidrule(lr){12-14} \cmidrule(lr){15-17} \cmidrule(lr){18-20}
    Model & Bbox & F1@0.5 & mIoU & FA (\%) & F1@0.5 & mIoU & FA (\%) & F1@0.5 & mIoU & FA (\%) & F1@0.5 & mIoU & FA (\%) & F1@0.5 & mIoU & FA (\%) & F1@0.5 & mIoU & FA (\%) & \\

    \rowcolor{tablebanner} \multicolumn{21}{l}{\textit{Single-label}} \\
    Qwen2.5VL & \texttt{xyxy} & 64.3 & 46.9 & 90.0 & 68.4 & 58.8 & 98.0 & 0.0 & 0.0 & 0.0 & 55.3 & 41.1 & 62.0 & 41.7 & 25.1 & 40.0 & -- & -- & -- & \texttt{bbox\_2d} \\

    Qwen3VL & \texttt{xyxy} & 74.6 & 60.9 & 96.0 & 65.0 & 46.3 & 72.0 & 74.0 & 57.6 & 94.0 & 60.0 & 39.0 & 62.0 & 62.4 & 42.9 & 66.0 & -- & -- & -- & \texttt{bbox\_2d} \\

    Qwen3.5 & \texttt{xyxy} & 77.7 & 71.1 & 100.0 & 0.0 & 0.0 & 0.0 & 0.0 & 0.0 & 0.0 & 0.0 & 0.0 & 0.0 & 0.0 & 0.0 & 0.0 & -- & -- & -- & \texttt{bbox\_2d} \\

  Qwen3.5- th & \texttt{xyxy} & 76.7 & 57.6 & 92.0 & 16.2 & 7.8 & 18.0 & 0.0 & 0.0 & 0.0 & 0.0 & 0.0 & 0.0 & 9.3 & 4.0 & 6.0 & -- & -- & -- & \texttt{bbox\_2d} \\

    InternVL3 & \texttt{xyxy} & 80.2 & 71.5 & 98.0 & 83.2 & 71.0 & 98.0 & 83.1 & 74.0 & 100.0 & 72.2 & 65.9 & 100.0 & 83.5 & 75.4 & 100.0 & -- & -- & -- & \texttt{coordinates} \\

    Gemma-4 & \texttt{yxyx} & 43.4 & 26.5 & 68.0 & 55.5 & 34.9 & 100.0 & 61.2 & 42.3 & 86.0 & 68.2 & 50.9 & 100.0 & 68.5 & 50.5 & 100.0 & -- & -- & -- & \texttt{bounding\_box} \\

    GLM4.6V & \texttt{xyxy} & 44.1 & 27.3 & 16.0 & 57.9 & 38.9 & 26.0 & 53.3 & 34.0 & 22.0 & 50.3 & 31.1 & 20.0 & 58.9 & 41.2 & 30.0 & -- & -- & -- & \texttt{coordinates} \\

    GPT5.4 & \texttt{xywh} & 40.4 & 44.7 & 100.0 & 41.2 & 43.3 & 100.0 & 45.5 & 44.4 & 100.0 & 43.8 & 41.6 & 100.0 & 47.2 & 45.0 & 100.0 & -- & -- & -- & \texttt{coordinates} \\

    Gemini 2.5 Flash & \texttt{yxyx} & 76.1 & 63.2 & 80.0 & 76.8 & 69.9 & 85.0 & 75.4 & 74.7 & 100.0 & 67.4 & 54.9 & 70.0 & 65.2 & 50.5 & 85.0 & -- & -- & -- & \texttt{bbox} \\

    Sonnet 4.5 & \texttt{xywh} & 27.4 & 37.6 & 100.0 & 26.0 & 31.7 & 100.0 & 29.8 & 30.6 & 95.0 & 21.0 & 32.5 & 100.0 & 22.2 & 35.1 & 100.0 & -- & -- & -- & \texttt{bbox} \\

    GPT-6 Astra & \texttt{cxcywh} & 76.5 & 78.7 & 100.0 & 74.8 & 77.4 & 100.0 & 71.8 & 77.1 & 100.0 & 77.2 & 78.2 & 100.0 & 77.2 & 77.8 & 100.0 & -- & -- & -- & \texttt{bounding\_box} \\

    Opus 5.5 & \texttt{yxhw} & 83.2 & 82.8 & 100.0 & 7.1 & 8.5 & 100.0 & 81.0 & 83.4 & 100.0 & 83.9 & 83.9 & 100.0 & 82.1 & 84.6 & 100.0 & -- & -- & -- & \texttt{bounding\_box} \\

    \rowcolor{tablebanner} \multicolumn{21}{l}{\textit{Multi-label}} \\
    Qwen2.5VL & \texttt{xyxy} & 66.1 & 51.9 & 86.0 & 83.2 & 67.1 & 96.0 & 47.1 & 29.7 & 54.0 & 73.2 & 53.7 & 90.0 & 68.1 & 49.5 & 86.0 & 0.0 & 0.0 & 0.0 & \texttt{box\_2d} \\

    Qwen3VL & \texttt{xyxy} & 74.1 & 65.5 & 96.0 & 64.8 & 50.9 & 94.0 & 64.7 & 49.7 & 90.0 & 67.3 & 48.3 & 82.0 & 75.4 & 59.3 & 90.0 & 69.5 & 57.7 & 92.0 & \texttt{bbox\_2d} \\

    Qwen3.5 & \texttt{xyxy} & 80.0 & 74.0 & 98.0 & 0.0 & 0.0 & 0.0 & 0.0 & 0.0 & 0.0 & 0.0 & 0.0 & 0.0 & 0.0 & 0.0 & 0.0 & 0.0 & 0.0 & 0.0 & \texttt{bbox\_2d} \\

 Qwen3.5- th & \texttt{xyxy} & 75.0 & 60.1 & 90.0 & 10.7 & 5.1 & 6.0 & 0.0 & 0.0 & 0.0 & 14.4 & 6.9 & 10.0 & 20.0 & 10.1 & 16.0 & 38.0 & 21.2 & 34.0 & \texttt{bbox\_2d} \\

    InternVL3 & \texttt{xyxy} & 82.5 & 67.8 & 96.0 & 80.9 & 67.1 & 96.0 & 83.1 & 68.4 & 96.0 & 82.0 & 66.4 & 94.0 & 83.7 & 70.1 & 98.0 & 47.8 & 29.1 & 64.0 & \texttt{coordinates} \\

    Gemma-4 & \texttt{yxyx} & 17.0 & 7.8 & 26.0 & 75.8 & 63.5 & 100.0 & 2.8 & 1.0 & 4.0 & 19.4 & 9.2 & 30.0 & 62.0 & 42.0 & 76.0 & 0.0 & 0.0 & 0.0 & \texttt{box\_2d} \\

    GLM4.6V & \texttt{xyxy} & 63.7 & 46.1 & 44.0 & 66.4 & 49.2 & 48.0 & 60.4 & 41.0 & 38.0 & 64.6 & 45.7 & 44.0 & 59.5 & 41.7 & 38.0 & 45.0 & 27.7 & 28.0 & \texttt{box\_2d} \\

    GPT5.4 & \texttt{xywh} & 41.2 & 44.9 & 100.0 & 41.3 & 40.6 & 100.0 & 53.2 & 44.1 & 100.0 & 45.8 & 41.9 & 100.0 & 50.5 & 48.7 & 100.0 & 49.0 & 51.0 & 100.0 & \texttt{bbox} \\

    Gemini 2.5 Flash & \texttt{yxyx} & 45.9 & 28.6 & 65.0 & 42.4 & 26.8 & 35.0 & 85.4 & 71.2 & 90.0 & 72.7 & 53.3 & 85.0 & 81.4 & 65.2 & 95.0 & 86.6 & 77.1 & 100.0 & \texttt{class\_name} \\

    Sonnet 4.5 & \texttt{xywh} & 31.7 & 30.9 & 100.0 & 32.1 & 31.6 & 100.0 & 28.6 & 34.8 & 100.0 & 29.6 & 33.2 & 100.0 & 32.6 & 33.8 & 100.0 & 28.2 & 30.8 & 100.0 & \texttt{coordinates} \\

    GPT-6 Astra & \texttt{cxcywh} & 80.8 & 80.5 & 100.0 & 78.8 & 81.0 & 100.0 & 82.7 & 81.4 & 100.0 & 77.8 & 81.6 & 100.0 & 80.4 & 82.1 & 100.0 & 82.7 & 81.8 & 100.0 & \texttt{class\_name} \\

    Opus 5.5 & \texttt{yxhw} & 88.0 & 87.0 & 100.0 & 17.8 & 22.3 & 100.0 & 80.4 & 82.6 & 100.0 & 83.2 & 83.6 & 100.0 & 84.0 & 82.8 & 100.0 & 84.3 & 86.0 & 100.0 & \texttt{bbox\_2d} \\
    \bottomrule
    \end{tabular}
    }
    \caption{Stage 2 sweep on Pascal. Each model's Stage 1 JSON-mode winner is held fixed in the Bbox column. \texttt{class\_name} is multi-label only. The key selected for each model is listed in \Cref{tab:preferred_format}.}
    \label{tab:stage_2_pascal}
\end{table*}

\textbf{Stage 2: JSON key.}
We take the bounding-box representation that scored highest in Stage 1's JSON-mode sweep and test multiple JSON keys for it.
The preferred JSON key can differ between single-label and multi-label.
JSON-key sensitivity differs widely by model.
Qwen3.5 records 0 F1 on every key other than \texttt{bbox\_2d} in both single-label and multi-label.
Qwen3-VL records F1 at or above 60 on every key tested.
InternVL3 F1 ranges from 72.2 (\texttt{bounding\_box}) to 83.5 (\texttt{coordinates}) across keys in single-label, and from 47.8 (\texttt{class\_name}) to 83.7 (\texttt{coordinates}) in multi-label.
GLM-4.6V format adherence sits between 16\% and 48\% across all keys, indicating that its JSON output fails to parse regardless of the key requested.
The GPT and Claude models are flat across keys, except Opus 5.5, whose F1 collapses with \texttt{box\_2d} (\Cref{app:quirks}).
Gemini 2.5 Flash F1 ranges from 65.2 (\texttt{coordinates}) to 76.8 (\texttt{box\_2d}) across keys in single-label, and from 42.4 (\texttt{box\_2d}) to 86.6 (\texttt{class\_name}) in multi-label.

\begin{table}[h]
    \centering
    \resizebox{\textwidth}{!}{
    \setlength{\tabcolsep}{6pt}
    \begin{tabular}{l cc cc cc}
    \toprule
     & \multicolumn{2}{c}{Unconstrained output} & \multicolumn{2}{c}{Bbox representation} & \multicolumn{2}{c}{JSON key} \\
    \cmidrule(lr){2-3} \cmidrule(lr){4-5} \cmidrule(lr){6-7}
    Model & Text & JSON & Text & JSON & Single-label & Multi-label \\
    \midrule
    \rowcolor{tablebanner} \multicolumn{7}{l}{\textit{Open Source Models}} \\
    Qwen2.5VL & xyxy\,+\,bbox\_2d & xyxy\,+\,bbox\_2d & xyxy & xyxy & bbox\_2d & box\_2d \\
    Qwen3VL & xyxy / xyxy\,+\,bbox\_2d & xyxy\,+\,bbox\_2d & xyxy & xyxy & bbox\_2d & bbox\_2d \\
    Qwen3.5 & xyxy\,+\,bbox\_2d & xyxy\,+\,bbox\_2d & xyxy & xyxy & bbox\_2d & bbox\_2d \\
    Qwen3.5-Thinking & xyxy / xyxy\,+\,bbox\_2d & xyxy\,+\,bbox\_2d & xyxy & xyxy & bbox\_2d & bbox\_2d \\
    InternVL3 & xyxy & -- & xyxy & xyxy & coordinates & coordinates \\
    Gemma-4 & yxyx\,+\,box\_2d & yxyx\,+\,box\_2d & yxyx & yxyx & bounding\_box & box\_2d \\
    GLM4.6V & xyxy & -- & xyxy & xyxy & coordinates & box\_2d \\
    \midrule
    \rowcolor{tablebanner} \multicolumn{7}{l}{\textit{Closed Source Models}} \\
    GPT5.4 & xyxy & -- & xyxy & xywh & coordinates & bbox \\
    Gemini 2.5 Flash & yxyx\,+\,box\_2d & yxyx\,+\,box\_2d & yxyx & yxyx & bbox & class\_name \\
    Sonnet 4.5 & xyxy & \{xmin,\,\ldots\} / xyxy+\,bbox & xyxy & xywh & bbox & coordinates \\
    GPT-6 Astra & xyxy & -- & xywh & cxcywh & bounding\_box & class\_name \\
    Opus 5.5 & xyxy & xyxy\,+\,bbox\_2d & yxhw & yxhw & bounding\_box & bbox\_2d \\
    \bottomrule
    \end{tabular}
    }
    \caption{Overall per-model preferred configuration selected via the multi-stage probe and used in \Cref{sec:results}.
    The \emph{Unconstrained output} columns name the parser used for the \texttt{--} column of \Cref{tab:lmms_stage1_pascal}: the bbox representation extracted from the unconstrained text response, and the axis with JSON key for the unconstrained JSON response.
    A dash marks models without a parsable JSON default.
    Sonnet 4.5 emits a corner-keyed JSON object $\{$\texttt{xmin}, \texttt{ymin}, \texttt{xmax}, \texttt{ymax}$\}$ for single-label queries and a single-key list under \texttt{bbox} for multi-label.}
    \label{tab:preferred_format}
\end{table}

\Cref{tab:preferred_format} lists each model's unconstrained output parser, Stage 1 bounding-box representation, and Stage 2 JSON key.

\begin{table*}[h]
    \centering
    \resizebox{1.0\textwidth}{!}{
    \setlength{\tabcolsep}{3pt}
    \begin{tabular}{lccc ccc ccc ccc | ccc}
    \toprule
     & & & & \multicolumn{3}{c}{Pascal} & \multicolumn{3}{c}{OpenImages} & \multicolumn{3}{c}{iGround} & \multicolumn{3}{|c}{Average} \\
    \cmidrule(lr){5-7} \cmidrule(lr){8-10} \cmidrule(lr){11-13} \cmidrule(lr){14-16}
    Model & Bbox & Format & JSON key & F1@0.5 & mIoU & FA (\%) & F1@0.5 & mIoU & FA (\%) & F1@0.5 & mIoU & FA (\%) & F1@0.5 & mIoU & FA (\%) \\
    \rowcolor{tablebanner} \multicolumn{16}{l}{\textit{Open Source Models}} \\
    \multirow{3}{*}{Qwen2.5VL} & \emph{-- (as xyxy)} & Text \emph{(as JSON)} & \emph{-- (as bbox\_2d)} & 70.1 & 52.6 & 98.9 & 49.0 & 37.6 & 98.0 & 42.5 & 40.6 & 90.0 & \textbf{53.9} & 43.6 & 95.6 \\
        & xyxy & Text & -- & 45.8 & 27.5 & 65.5 & 42.1 & 30.1 & 99.4 & 41.5 & 39.7 & 95.1 & 43.2 & 32.4 & 86.7 \\
     & xyxy & JSON & bbox\_2d & 63.0 & 48.3 & 92.4 & 46.8 & 34.3 & 90.7 & 35.3 & 32.8 & 78.4 & \underline{48.4} & 38.5 & 87.2 \\
    \midrule
    \multirow{3}{*}{Qwen3VL} & \emph{-- (as xyxy)} & Text \emph{(as JSON)} & \emph{-- (as bbox\_2d)} & 77.5 & 66.5 & 97.1 & 62.8 & 54.6 & 97.3 & 60.4 & 57.5 & 85.3 & \textbf{66.9} & 59.6 & 93.2 \\
        & xyxy & Text & -- & 63.8 & 51.5 & 100.0 & 48.1 & 37.0 & 100.0 & 58.9 & 55.6 & 94.0 & 56.9 & 48.0 & 98.0 \\
     & xyxy & JSON & bbox\_2d & 76.6 & 66.4 & 97.0 & 62.9 & 56.4 & 98.0 & 58.6 & 57.9 & 88.5 & \underline{66.0} & 60.2 & 94.5 \\
    \midrule
    \multirow{3}{*}{Qwen3.5} & \emph{-- (as xyxy)} & Text \emph{(as JSON)} & \emph{-- (as bbox\_2d)} & 75.4 & 66.3 & 97.7 & 53.5 & 48.9 & 96.3 & 54.8 & 58.6 & 98.4 & \underline{61.2} & 57.9 & 97.5 \\
        & xyxy & Text & -- & 64.8 & 59.4 & 99.7 & 41.3 & 40.1 & 99.6 & 57.9 & 55.8 & 95.8 & 54.7 & 51.8 & 98.4 \\
     & xyxy & JSON & bbox\_2d & 77.3 & 67.2 & 98.4 & 53.9 & 48.9 & 97.4 & 55.3 & 59.3 & 99.5 & \textbf{62.2} & 58.5 & 98.4 \\
    \midrule
    \multirow{3}{*}{\makecell[l]{Qwen3.5 \\ - thinking}} & \emph{-- (as xyxy)} & Text \emph{(as JSON)} & \emph{-- (as bbox\_2d)} & 75.3 & 58.5 & 94.9 & 52.1 & 41.1 & 87.5 & 62.3 & 59.1 & 83.2 & \textbf{63.2} & 52.9 & 88.5 \\
        & xyxy & Text & -- & 76.6 & 61.1 & 97.0 & 51.9 & 43.4 & 91.4 & 59.6 & 59.2 & 85.4 & 62.7 & 54.6 & 91.3 \\
     & xyxy & JSON & bbox\_2d & 74.5 & 57.4 & 94.2 & 52.0 & 41.6 & 87.7 & 62.5 & 59.5 & 81.8 & \underline{63.0} & 52.8 & 87.9 \\
    \midrule
    \multirow{2}{*}{InternVL3} & xyxy & Text & -- & 77.0 & 62.5 & 99.3 & 34.9 & 27.6 & 95.8 & 53.1 & 53.5 & 96.7 & \textbf{55.0} & 47.9 & 97.3 \\
        & xyxy & JSON & coordinates & 77.1 & 65.2 & 97.6 & 34.9 & 28.4 & 99.4 & 52.8 & 54.9 & 99.8 & \underline{55.0} & 49.5 & 98.9 \\
    \midrule
    \multirow{2}{*}{Gemma-4} & yxyx & Text & -- & 68.7 & 56.2 & 99.8 & 34.9 & 26.4 & 94.7 & 53.5 & 42.6 & 92.1 & \textbf{52.4} & 41.7 & 95.5 \\
     & yxyx & JSON & bounding\_box & 58.0 & 41.5 & 99.1 & 31.5 & 20.7 & 99.1 & 55.2 & 42.2 & 99.9 & \underline{48.2} & 34.8 & 99.4 \\
    \midrule
    \multirow{2}{*}{GLM4.6V} & xyxy & Text & -- & 80.5 & 64.9 & 99.4 & 55.9 & 50.0 & 98.2 & 60.3 & 61.1 & 99.5 & \textbf{65.5} & 58.7 & 99.0 \\
        & xyxy & JSON & coordinates & 47.9 & 29.1 & 21.2 & 33.0 & 23.8 & 34.0 & 14.2 & 9.0 & 14.6 & \underline{31.7} & 20.7 & 23.3 \\
    \rowcolor{tablebanner} \multicolumn{16}{l}{\textit{Closed Source Models}} \\
    \multirow{2}{*}{GPT5.4} & xyxy & Text & -- & 48.8 & 43.3 & 100.0 & 41.3 & 36.1 & 100.0 & 37.3 & 39.9 & 100.0 & \textbf{42.5} & 39.7 & 100.0 \\
        & xywh & JSON & coordinates & 47.1 & 43.8 & 100.0 & 38.8 & 37.8 & 99.9 & 27.0 & 36.4 & 100.0 & \underline{37.6} & 39.3 & 100.0 \\
    \midrule
    \multirow{2}{*}{Gemini 2.5 Flash} & yxyx & Text & -- & 43.3 & 56.5 & 99.5 & 29.4 & 46.3 & 97.1 & 41.0 & 53.7 & 79.1 & \underline{37.9} & 52.2 & 91.9 \\
        & yxyx & JSON & bbox & 74.5 & 64.4 & 94.9 & 46.8 & 42.6 & 83.6 & 53.2 & 57.7 & 74.4 & \textbf{58.2} & 54.9 & 84.3 \\
    \midrule
    \multirow{2}{*}{Sonnet 4.5} & xyxy & Text & -- & 27.8 & 27.9 & 97.8 & 21.8 & 20.5 & 76.0 & 21.6 & 27.6 & 76.6 & \underline{23.7} & 25.3 & 83.5 \\
        & xywh & JSON & bbox & 27.3 & 31.7 & 99.9 & 24.1 & 25.6 & 99.8 & 22.2 & 30.2 & 100.0 & \textbf{24.5} & 29.1 & 99.9 \\
    \midrule
    \multirow{2}{*}{GPT-6 Astra} & xywh & Text & -- & 81.4 & 79.6 & 99.9 & 46.2 & 66.2 & 99.9 & 54.7 & 70.5 & 100.0 & \textbf{60.8} & 72.1 & 99.9 \\
        & cxcywh & JSON & bounding\_box & 80.2 & 80.7 & 100.0 & 43.6 & 66.5 & 99.8 & 51.2 & 70.9 & 99.9 & \underline{58.3} & 72.7 & 99.9 \\
    \midrule
    \multirow{2}{*}{Opus 5.5} & yxhw & Text & -- & 86.3 & 82.0 & 100.0 & 56.7 & 66.6 & 99.7 & 60.2 & 71.8 & 96.7 & \textbf{67.7} & 73.5 & 98.8 \\
        & yxhw & JSON & bounding\_box & 85.2 & 82.0 & 100.0 & 55.2 & 65.6 & 100.0 & 51.8 & 66.1 & 100.0 & \underline{64.1} & 71.2 & 100.0 \\
    \bottomrule
    \end{tabular}
    }
    \caption{Object detection results, single-label. The Average columns aggregate across the three datasets.}
    \label{tab:objdet_single}
    \end{table*}

\begin{table*}[h]
    \centering
    \resizebox{1.0\textwidth}{!}{
    \setlength{\tabcolsep}{3pt}
    \begin{tabular}{lccc ccc ccc ccc | ccc}
    \toprule
     & & & & \multicolumn{3}{c}{Pascal} & \multicolumn{3}{c}{OpenImages} & \multicolumn{3}{c}{iGround} & \multicolumn{3}{|c}{Average} \\
    \cmidrule(lr){5-7} \cmidrule(lr){8-10} \cmidrule(lr){11-13} \cmidrule(lr){14-16}
    Model & Bbox & Format & JSON key & F1@0.5 & mIoU & FA (\%) & F1@0.5 & mIoU & FA (\%) & F1@0.5 & mIoU & FA (\%) & F1@0.5 & mIoU & FA (\%) \\
    \rowcolor{tablebanner} \multicolumn{16}{l}{\textit{Open Source Models}} \\
    \multirow{3}{*}{Qwen2.5VL} & \emph{-- (as xyxy)} & Text \emph{(as JSON)} & \emph{-- (as bbox\_2d)} & 80.5 & 66.8 & 99.2 & 41.5 & 30.8 & 96.4 & 37.6 & 38.1 & 95.2 & \textbf{53.2} & 45.3 & 96.9 \\
        & xyxy & Text & -- & 17.7 & 61.1 & 95.3 & 5.4 & 30.1 & 75.9 & 17.7 & 26.3 & 65.5 & 13.6 & 39.2 & 78.9 \\
     & xyxy & JSON & box\_2d & 72.0 & 54.0 & 90.6 & 35.1 & 21.7 & 81.6 & 38.8 & 39.8 & 98.9 & \underline{48.6} & 38.5 & 90.4 \\
    \midrule
    \multirow{3}{*}{Qwen3VL} & \emph{-- (as xyxy)} & Text \emph{(as JSON)} & \emph{-- (as bbox\_2d)} & 81.2 & 70.9 & 96.5 & 51.5 & 37.8 & 89.3 & 58.8 & 58.4 & 93.6 & \textbf{63.8} & 55.7 & 93.1 \\
        & xyxy & Text & -- & 64.4 & 44.8 & 100.0 & 44.3 & 29.6 & 99.9 & 56.9 & 54.4 & 99.3 & 55.2 & 42.9 & 99.7 \\
     & xyxy & JSON & bbox\_2d & 75.1 & 60.6 & 96.2 & 41.8 & 27.2 & 86.0 & 58.3 & 52.1 & 93.8 & \underline{58.4} & 46.6 & 92.0 \\
    \midrule
    \multirow{3}{*}{Qwen3.5} & \emph{-- (as xyxy)} & Text \emph{(as JSON)} & \emph{-- (as bbox\_2d)} & 80.4 & 70.0 & 97.2 & 45.0 & 32.8 & 87.0 & 57.1 & 57.9 & 96.7 & \textbf{60.9} & 53.6 & 93.6 \\
        & xyxy & Text & -- & 19.8 & 27.5 & 48.6 & 10.9 & 40.7 & 97.0 & 5.4 & 47.5 & 79.1 & 12.0 & 38.6 & 74.9 \\
     & xyxy & JSON & bbox\_2d & 80.5 & 69.7 & 97.8 & 44.3 & 32.1 & 86.9 & 57.7 & 58.4 & 98.7 & \underline{60.8} & 53.4 & 94.5 \\
    \midrule
    \multirow{2}{*}{\makecell[l]{Qwen3.5 \\ - thinking}} & xyxy & Text & -- & 78.4 & 60.6 & 94.1 & 23.9 & 13.2 & 58.4 & 60.1 & 49.0 & 83.4 & \textbf{54.1} & 41.0 & 78.6 \\
     & xyxy & JSON & bbox\_2d & 74.6 & 55.6 & 90.6 & 18.9 & 10.3 & 50.4 & 59.4 & 46.5 & 77.6 & \underline{51.0} & 37.4 & 72.9 \\
    \midrule
    \multirow{2}{*}{InternVL3} & xyxy & Text & -- & 78.4 & 60.9 & 99.9 & 33.7 & 23.5 & 99.9 & 56.1 & 52.5 & 99.4 & \underline{56.1} & 45.6 & 99.7 \\
        & xyxy & JSON & coordinates & 82.0 & 69.2 & 97.4 & 34.8 & 25.4 & 95.2 & 54.8 & 52.2 & 97.3 & \textbf{57.2} & 49.0 & 96.6 \\
    \midrule
    \multirow{2}{*}{Gemma-4} & yxyx & Text & -- & 69.5 & 53.3 & 99.9 & 41.6 & 32.8 & 99.7 & 59.7 & 51.7 & 97.4 & \underline{56.9} & 46.0 & 99.0 \\
     & yxyx & JSON & box\_2d & 75.1 & 62.6 & 99.9 & 40.0 & 32.1 & 95.4 & 58.7 & 53.7 & 100.0 & \textbf{58.0} & 49.5 & 98.4 \\
    \midrule
    \multirow{2}{*}{GLM4.6V} & xyxy & Text & -- & 82.2 & 65.3 & 91.0 & 45.8 & 35.5 & 95.8 & 59.7 & 58.3 & 98.4 & \textbf{62.5} & 53.1 & 95.1 \\
     & xyxy & JSON & box\_2d & 71.9 & 51.4 & 54.7 & 35.4 & 23.6 & 64.5 & 59.8 & 57.5 & 94.2 & \underline{55.7} & 44.2 & 71.1 \\
    \rowcolor{tablebanner} \multicolumn{16}{l}{\textit{Closed Source Models}} \\
    \multirow{2}{*}{GPT5.4} & xyxy & Text & -- & 55.6 & 48.1 & 100.0 & 38.5 & 32.7 & 99.9 & 39.5 & 40.2 & 99.7 & \textbf{44.5} & 40.4 & 99.9 \\
     & xywh & JSON & bbox & 53.3 & 47.6 & 99.6 & 36.8 & 32.8 & 96.7 & 32.2 & 37.8 & 100.0 & \underline{40.8} & 39.4 & 98.8 \\
    \midrule
    \multirow{2}{*}{Gemini 2.5 Flash} & yxyx & Text & -- & 58.4 & 60.3 & 100.0 & 35.6 & 33.7 & 99.3 & 52.7 & 54.8 & 98.0 & \underline{48.9} & 49.6 & 99.1 \\
     & yxyx & JSON & class\_name & 78.6 & 66.7 & 94.8 & 42.4 & 31.5 & 78.3 & 59.9 & 60.6 & 92.1 & \textbf{60.3} & 52.9 & 88.4 \\
    \midrule
    \multirow{2}{*}{Sonnet 4.5} & xyxy & Text & -- & 29.7 & 33.6 & 99.9 & 23.5 & 23.1 & 99.6 & 26.3 & 31.6 & 99.6 & \underline{26.5} & 29.4 & 99.7 \\
     & xywh & JSON & coordinates & 32.3 & 34.5 & 99.9 & 22.9 & 21.1 & 96.1 & 25.9 & 31.6 & 100.0 & \textbf{27.0} & 29.1 & 98.7 \\
    \midrule
    \multirow{2}{*}{GPT-6 Astra} & xywh & Text & -- & 82.8 & 81.4 & 100.0 & 47.3 & 62.0 & 100.0 & 64.4 & 72.3 & 98.7 & \textbf{64.8} & 71.9 & 99.6 \\
        & cxcywh & JSON & class\_name & 82.6 & 81.2 & 99.8 & 46.9 & 60.9 & 99.6 & 64.0 & 72.9 & 99.9 & \underline{64.5} & 71.7 & 99.8 \\
    \midrule
    \multirow{2}{*}{Opus 5.5} & yxhw & Text & -- & 85.6 & 81.7 & 100.0 & 50.4 & 54.1 & 99.9 & 65.1 & 71.9 & 97.8 & \textbf{67.0} & 69.2 & 99.2 \\
        & yxhw & JSON & bbox\_2d & 87.2 & 83.1 & 100.0 & 22.5 & 27.6 & 100.0 & 64.2 & 71.5 & 100.0 & \underline{58.0} & 60.7 & 100.0 \\
    \bottomrule
    \end{tabular}
    }
    \caption{Object detection results, multi-label. The Average columns aggregate across the three datasets.}
    \label{tab:objdet_multi}
    \end{table*}

\begin{table*}[h]
    \centering
    \resizebox{1.0\textwidth}{!}{
    \setlength{\tabcolsep}{3pt}
    \begin{tabular}{lccc ccc ccc ccc ccc ccc ccc | ccc}
    \toprule
     & & & & \multicolumn{3}{c}{RefCOCO-Avg} & \multicolumn{3}{c}{RefL4} & \multicolumn{3}{c}{Flickr30k-Entities} & \multicolumn{3}{c}{D3} & \multicolumn{3}{c}{PhraseCut} & \multicolumn{3}{c}{SVG} & \multicolumn{3}{|c}{Average} \\
    \cmidrule(lr){5-7} \cmidrule(lr){8-10} \cmidrule(lr){11-13} \cmidrule(lr){14-16} \cmidrule(lr){17-19} \cmidrule(lr){20-22} \cmidrule(lr){23-25}
    Model & Bbox & Format & JSON key & F1@0.5 & mIoU & FA (\%) & F1@0.5 & mIoU & FA (\%) & F1@0.5 & mIoU & FA (\%) & F1@0.5 & mIoU & FA (\%) & F1@0.5 & mIoU & FA (\%) & F1@0.5 & mIoU & FA (\%) & F1@0.5 & mIoU & FA (\%) \\
    \rowcolor{tablebanner} \multicolumn{25}{l}{\textit{Open Source Models}} \\
    \multirow{3}{*}{Qwen2.5VL} & \emph{-- (as xyxy)} & Text \emph{(as JSON)} & \emph{-- (as bbox\_2d)} & 85.5 & 78.7 & 100.0 & 82.3 & 74.4 & 99.9 & 58.1 & 46.8 & 99.8 & 51.7 & 48.9 & 97.9 & 49.4 & 40.1 & 99.4 & 74.1 & 66.2 & 100.0 & \textbf{66.9} & 59.2 & 99.5 \\
        & xyxy & Text & -- & 81.9 & 71.9 & 91.5 & 81.7 & 74.7 & 99.8 & 53.3 & 39.2 & 92.6 & 47.4 & 49.4 & 100.0 & 43.4 & 31.9 & 94.2 & 74.6 & 65.3 & 98.5 & \underline{63.7} & 55.4 & 96.1 \\
     &  & JSON & bbox\_2d & 74.2 & 69.0 & 92.4 & 76.4 & 71.2 & 98.0 & 50.5 & 39.6 & 96.0 & 46.1 & 39.9 & 98.8 & 35.7 & 27.9 & 87.7 & 65.7 & 63.4 & 97.6 & 58.1 & 51.8 & 95.1 \\
    \midrule
    \multirow{3}{*}{Qwen3VL} & \emph{-- (as xyxy)} & Text \emph{(as JSON)} & \emph{-- (as bbox\_2d)} & 87.5 & 80.0 & 99.9 & 88.6 & 80.4 & 100.0 & 57.1 & 44.8 & 100.0 & 45.9 & 53.3 & 99.9 & 48.7 & 38.8 & 99.4 & 77.8 & 68.6 & 100.0 & \underline{67.6} & 61.0 & 99.9 \\
        & xyxy & Text & -- & 87.3 & 79.9 & 100.0 & 88.5 & 80.1 & 100.0 & 53.2 & 41.0 & 100.0 & 45.3 & 49.6 & 100.0 & 44.1 & 33.1 & 100.0 & 77.6 & 68.5 & 100.0 & 66.0 & 58.7 & 100.0 \\
     &  & JSON & bbox\_2d & 87.3 & 79.9 & 99.9 & 88.4 & 80.1 & 100.0 & 57.5 & 45.0 & 100.0 & 46.9 & 53.3 & 99.8 & 48.4 & 39.7 & 99.2 & 77.4 & 68.6 & 100.0 & \textbf{67.7} & 61.1 & 99.8 \\
    \midrule
    \multirow{3}{*}{Qwen3.5} & \emph{-- (as xyxy)} & Text \emph{(as JSON)} & \emph{-- (as bbox\_2d)} & 86.8 & 77.4 & 100.0 & 82.5 & 72.7 & 100.0 & 59.7 & 45.3 & 99.8 & 35.7 & 54.9 & 100.0 & 43.6 & 35.7 & 98.8 & 74.4 & 64.5 & 100.0 & 63.8 & 58.4 & 99.8 \\
        & xyxy & Text & -- & 86.9 & 77.3 & 100.0 & 84.1 & 73.1 & 100.0 & 58.1 & 42.7 & 99.5 & 37.1 & 50.3 & 99.6 & 43.6 & 33.1 & 99.7 & 75.3 & 65.4 & 100.0 & \underline{64.2} & 57.0 & 99.8 \\
     &  & JSON & bbox\_2d & 86.5 & 77.6 & 100.0 & 82.4 & 72.6 & 100.0 & 62.0 & 47.8 & 99.9 & 36.2 & 53.7 & 100.0 & 43.9 & 36.2 & 98.8 & 75.1 & 65.2 & 100.0 & \textbf{64.4} & 58.8 & 99.8 \\
    \midrule
    \multirow{3}{*}{\makecell[l]{Qwen3.5 \\ - thinking}} & \emph{-- (as xyxy)} & Text \emph{(as JSON)} & \emph{-- (as bbox\_2d)} & 86.8 & 77.2 & 99.0 & 83.6 & 72.6 & 99.1 & 63.7 & 52.2 & 94.7 & 47.0 & 60.1 & 95.9 & 45.7 & 37.1 & 94.5 & 73.4 & 64.5 & 99.4 & \underline{66.7} & 60.6 & 97.1 \\
        & xyxy & Text & -- & 87.2 & 77.4 & 99.3 & 83.2 & 72.4 & 98.8 & 63.1 & 51.8 & 96.9 & 47.9 & 60.4 & 98.1 & 48.5 & 39.2 & 96.7 & 73.3 & 64.4 & 99.3 & \textbf{67.2} & 60.9 & 98.2 \\
     &  & JSON & bbox\_2d & 84.8 & 74.5 & 95.6 & 77.1 & 63.0 & 86.9 & 63.8 & 52.5 & 93.6 & 46.6 & 57.6 & 95.1 & 46.0 & 36.5 & 92.8 & 67.7 & 57.2 & 87.1 & 64.3 & 56.9 & 91.8 \\
    \midrule
    \multirow{2}{*}{InternVL3} & xyxy & Text & -- & 78.1 & 70.5 & 100.0 & 68.8 & 61.4 & 100.0 & 61.6 & 51.1 & 97.9 & 47.1 & 53.8 & 100.0 & 37.8 & 32.8 & 99.6 & 72.3 & 66.5 & 100.0 & \textbf{61.0} & 56.0 & 99.6 \\
        &  & JSON & coordinates & 65.2 & 75.8 & 100.0 & 58.8 & 64.7 & 99.9 & 61.7 & 58.8 & 98.8 & 43.2 & 36.1 & 99.8 & 36.2 & 35.6 & 98.7 & 67.2 & 73.3 & 99.4 & \underline{55.4} & 57.4 & 99.4 \\
    \midrule
    \multirow{2}{*}{Gemma-4} & \multirow{2}{*}{yxyx} & Text & -- & 71.8 & 65.2 & 99.8 & 75.0 & 67.8 & 100.0 & 55.2 & 50.6 & 99.5 & 45.9 & 52.2 & 99.7 & 40.3 & 36.8 & 99.6 & 60.6 & 54.7 & 100.0 & \underline{58.1} & 54.6 & 99.8 \\
     &  & JSON & bounding\_box & 71.2 & 65.9 & 99.8 & 75.8 & 69.0 & 100.0 & 57.9 & 49.9 & 99.7 & 46.8 & 55.0 & 99.9 & 39.5 & 36.8 & 99.9 & 59.8 & 56.9 & 100.0 & \textbf{58.5} & 55.6 & 99.9 \\
    \midrule
    \multirow{2}{*}{GLM4.6V} & xyxy & Text & -- & 84.3 & 78.3 & 99.1 & 86.7 & 79.3 & 98.9 & 74.5 & 64.6 & 99.5 & 39.6 & 59.0 & 98.2 & 49.7 & 41.2 & 98.7 & 77.0 & 68.5 & 99.2 & \textbf{68.6} & 65.2 & 98.9 \\
        &  & JSON & coordinates & 2.5 & 1.2 & 1.5 & 2.0 & 0.9 & 1.1 & 34.3 & 19.9 & 17.2 & 19.1 & 12.5 & 13.6 & 17.5 & 9.6 & 14.4 & 2.7 & 1.2 & 1.6 & \underline{13.0} & 7.5 & 8.2 \\
    \rowcolor{tablebanner} \multicolumn{25}{l}{\textit{Closed Source Models}} \\
    \multirow{2}{*}{GPT5.4} & xyxy & Text & -- & 62.5 & 52.7 & 100.0 & 60.9 & 52.6 & 100.0 & 40.1 & 36.7 & 100.0 & 41.2 & 41.5 & 100.0 & 28.1 & 30.0 & 99.9 & 36.3 & 38.4 & 100.0 & \textbf{44.8} & 42.0 & 100.0 \\
        & xywh & JSON & coordinates & 56.4 & 51.0 & 100.0 & 58.8 & 50.9 & 100.0 & 38.0 & 39.0 & 100.0 & 43.1 & 50.3 & 100.0 & 24.7 & 29.9 & 99.8 & 34.0 & 40.0 & 100.0 & \underline{42.5} & 43.5 & 100.0 \\
    \midrule
    \multirow{2}{*}{Gemini 2.5 Flash} & \multirow{2}{*}{yxyx} & Text & -- & 74.7 & 66.1 & 99.4 & 68.0 & 60.5 & 99.7 & 31.9 & 46.7 & 97.8 & 39.1 & 53.4 & 99.2 & 18.1 & 35.9 & 97.4 & 57.8 & 52.0 & 99.7 & \underline{48.3} & 52.4 & 98.9 \\
        &  & JSON & bbox & 74.7 & 70.7 & 97.7 & 74.4 & 69.0 & 97.8 & 57.4 & 55.3 & 90.3 & 50.8 & 60.4 & 95.6 & 36.9 & 39.7 & 83.2 & 66.8 & 65.1 & 96.4 & \textbf{60.2} & 60.0 & 93.5 \\
    \midrule
    \multirow{2}{*}{Sonnet 4.5} & xyxy & Text & -- & 29.4 & 35.0 & 96.4 & 23.2 & 29.4 & 89.6 & 19.7 & 24.5 & 96.9 & 28.1 & 34.3 & 100.0 & 15.6 & 22.1 & 96.5 & 15.2 & 22.5 & 95.7 & \underline{21.9} & 28.0 & 95.8 \\
        & xywh & JSON & bbox & 31.3 & 37.5 & 99.4 & 26.3 & 31.9 & 92.6 & 21.6 & 24.3 & 100.0 & 25.0 & 35.9 & 99.9 & 16.6 & 23.6 & 99.9 & 18.8 & 27.6 & 100.0 & \textbf{23.3} & 30.1 & 98.6 \\
    \midrule
    \multirow{2}{*}{GPT-6 Astra} & xywh & Text & -- & 92.4 & 85.4 & 100.0 & 93.3 & 85.5 & 100.0 & 59.0 & 68.7 & 99.8 & 53.3 & 81.2 & 99.9 & 45.8 & 55.4 & 99.9 & 77.8 & 70.3 & 100.0 & \textbf{70.3} & 74.4 & 99.9 \\
        & cxcywh & JSON & bounding\_box & 76.4 & 71.5 & 100.0 & 74.0 & 69.4 & 100.0 & 58.8 & 71.2 & 99.8 & 48.7 & 80.6 & 99.8 & 42.8 & 56.4 & 100.0 & 63.3 & 61.3 & 100.0 & \underline{60.7} & 68.4 & 99.9 \\
    \midrule
    \multirow{2}{*}{Opus 5.5} & yxhw & Text & -- & 90.2 & 83.9 & 100.0 & 91.4 & 84.2 & 100.0 & 63.0 & 61.9 & 99.9 & 62.2 & 78.0 & 100.0 & 51.9 & 55.7 & 100.0 & 81.2 & 74.1 & 100.0 & \textbf{73.3} & 73.0 & 100.0 \\
        & yxhw & JSON & bounding\_box & 82.3 & 78.6 & 99.9 & 86.7 & 80.7 & 100.0 & 60.4 & 62.3 & 99.8 & 58.8 & 77.8 & 100.0 & 47.7 & 55.0 & 100.0 & 72.3 & 71.3 & 100.0 & \underline{68.0} & 70.9 & 100.0 \\
    \bottomrule
    \end{tabular}
    }
    \caption{Referring expression results across six datasets. The Average columns aggregate across all six. The RefCOCO/+/g splits are aggregated as RefCOCO-Avg; per-split numbers are in \Cref{tab:refcoco}.}
    \label{tab:refexp}
    \end{table*}

\begin{table*}[h]
    \centering
    \resizebox{1.0\textwidth}{!}{
    \setlength{\tabcolsep}{3pt}
    \begin{tabular}{lccc ccc ccc ccc | ccc}
    \toprule
     & & & & \multicolumn{3}{c}{HR-InsDet {\small\textit{easy}}} & \multicolumn{3}{c}{HR-InsDet {\small\textit{hard}}} & \multicolumn{3}{c}{RoboTools} & \multicolumn{3}{|c}{Average} \\
    \cmidrule(lr){5-7} \cmidrule(lr){8-10} \cmidrule(lr){11-13} \cmidrule(lr){14-16}
    Model & Bbox & Format & JSON key & F1@0.5 & mIoU & FA (\%) & F1@0.5 & mIoU & FA (\%) & F1@0.5 & mIoU & FA (\%) & F1@0.5 & mIoU & FA (\%) \\
    
    \rowcolor{tablebanner} \multicolumn{16}{l}{\textit{Open Source Models}} \\
    \multirow{3}{*}{Qwen2.5VL} & \emph{-- (as xyxy)} & Text \emph{(as JSON)} & \emph{-- (as bbox\_2d)} & 14.5 & 18.8 & 99.8 & 16.1 & 18.7 & 99.8 & 55.3 & 49.0 & 93.8 & \textbf{28.7} & 28.8 & 97.8\\
        & \multirow{2}{*}{xyxy} & Text & -- & 14.5 & 16.9 & 98.6  & 14.2 & 16.1 & 99.3  & 55.6 & 50.5 & 100.0 & \underline{28.1} & 27.8 & 99.3 \\
     &  & JSON & bbox\_2d & 13.3 & 15.7 & 79.1   & 12.6 & 12.5 & 61.8  & 46.8 & 37.2 & 72.0 & 24.2 & 21.8 & 71.0 \\
    \midrule
    \multirow{3}{*}{Qwen3VL} & \emph{-- (as xyxy)} & Text \emph{(as JSON)} & \emph{-- (as bbox\_2d)} & 63.5 &  51.5 &  98.4 & 42.0 &  38.4 &  99.9 & 80.7 &  72.7 &  98.8 & \underline{62.1} & 54.2 & 99.0 \\
        & \multirow{2}{*}{xyxy} & Text & -- & 62.9 &  51.2 &  99.7  & 42.2 &  38.7 &  99.7 & 80.3 &  71.6 &  100.0 & 61.8 & 53.8 & 99.8 \\
  &  & JSON & bbox\_2d & 62.3 &  51.1 &  99.4  &  42.8 &  38.6 &  99.6 &  82.0 &  72.4 &  99.4 & \textbf{62.3} & 54.0 & 99.5 \\

    \midrule
    \multirow{3}{*}{Qwen3.5} & \emph{-- (as xyxy)} & Text \emph{(as JSON)} & \emph{-- (as bbox\_2d)} & 50.3 & 45.6 & 98.8 & 29.9 & 33.6 & 99.0 & 28.3 & 29.8 & 98.1 & \underline{36.2} & 36.4 & 98.6 \\
        & \multirow{2}{*}{xyxy} & Text & -- & 52.8 & 45.3 & 99.6 & 31.8 & 33.6 & 99.1 & 28.1 & 32.8 & 100.0  & \textbf{37.6} & 37.3 & 99.6 \\
     &  & JSON & bbox\_2d & 49.0 & 46.1 & 99.5 & 28.3 & 33.4 & 99.1 & 29.2 & 32.0 & 100.0 & 35.5 & 37.2 & 99.5 \\
    \midrule
    \multirow{3}{*}{\makecell[l]{Qwen3.5 \\ - thinking}} & \emph{-- (as xyxy)} & Text \emph{(as JSON)} & \emph{-- (as bbox\_2d)} & 47.6 & 43.3 & 95.3 & 24.8 & 30.1 & 91.3 & 32.1 & 32.7 & 91.9 & \textbf{34.8} & 35.4 & 92.8 \\
        & \multirow{2}{*}{xyxy} & Text & -- & 38.7 & 24.2 & 44.4 & 19.7 & 13.2 & 29.7 & 29.9 & 30.0 & 88.8 & 29.4 & 22.4 & 54.3 \\
     &  & JSON & bbox\_2d & 42.0 & 36.9 & 84.9 & 22.2 & 25.0 & 81.9 & 24.9 & 18.3 & 57.1 & \underline{29.7} & 26.8 & 74.7 \\
    \midrule
    \multirow{2}{*}{InternVL3} & \multirow{2}{*}{xyxy} & Text & -- & 0.0 & 0.0 & 93.3 & 0.0 & 0.0 & 97.2 & 0.0 & 1.3 & 100.0 & \underline{0.0} & 0.5 & 96.8 \\
        &  & JSON & coordinates & 0.0 & 0.6 & 98.1 & 0.0 & 0.3 & 99.0 & 1.5 & 7.2 & 99.4 & \textbf{0.5} & 2.7 & 98.8 \\
    \midrule
    \multirow{2}{*}{Gemma-4} & \multirow{2}{*}{yxyx} & Text & -- & 0.2 & 0.8 & 100.0 & 0.0 & 0.4 & 100.0 & 0.6 & 6.4 & 100.0 & \textbf{0.3} & 2.5 & 100.0 \\
     &  & JSON & bounding\_box & 0.0 & 0.2 & 20.2 & 0.0 & 0.1 & 24.2 & 0.6 & 6.3 & 98.8 & \underline{0.2} & 2.2 & 47.7 \\
    \midrule
    \multirow{2}{*}{GLM4.6V} & \multirow{2}{*}{xyxy} & Text & -- & 64.1 & 54.5 & 99.3 & 41.1 & 37.6 & 99.7 & 74.8 & 72.0 & 99.4 & \textbf{60.0} & 54.7 & 99.5 \\
        &  & JSON & coordinates & 10.7 & 4.9 & 7.9 & 5.5 & 2.5 & 5.9 & 7.8 & 3.7 & 5.0 & \underline{8.0} & 3.7 & 6.3 \\
    \rowcolor{tablebanner} \multicolumn{16}{l}{\textit{Closed Source Models}} \\
    \multirow{2}{*}{GPT5.4} & xyxy & Text & -- & 19.3 & 24.9 & 100.0 & 8.8 & 13.1 & 100.0 & 44.2 & 42.8 & 100.0 & \underline{24.1} & 26.9 & 100.0 \\
        & xywh & JSON & coordinates & 20.0 & 25.3 & 100.0 & 8.1 & 12.6 & 100.0 & 47.9 & 45.7 & 100.0 & \textbf{25.3} & 27.8 & 100.0 \\
    \midrule
    \multirow{2}{*}{Gemini 2.5 Flash} & \multirow{2}{*}{yxyx} & Text & -- & 9.0 & 21.9 & 98.0 & 3.0 & 8.4 & 97.5 & 26.7 & 31.9 & 93.2 & \underline{12.9} & 20.8 & 96.2 \\
     &  & JSON & box\_2d* & 17.2 & 18.0 & 86.6 & 4.5 & 7.8 & 94.2 & 32.3 & 34.7 & 96.3 & \textbf{18.0} & 20.2 & 92.4 \\
    \midrule
    \multirow{2}{*}{Sonnet 4.5} & xyxy & Text & -- & 12.0 & 13.1 & 74.1 & 6.9 & 7.7 & 69.2 & 14.4 & 17.5 & 92.5 & \underline{11.1} & 12.8 & 78.6 \\
        & xywh & JSON & bbox & 12.6 & 18.3 & 99.8 & 7.8 & 11.3 & 92.0 & 15.0 & 21.1 & 100.0 & \textbf{11.8} & 16.9 & 97.3 \\
    \midrule
    \multirow{2}{*}{GPT-6 Astra} & xywh & Text & -- & 94.4 & 88.8 & 100.0 & 82.7 & 78.2 & 99.8 & 92.4 & 89.7 & 100.0 & \underline{89.8} & 85.6 & 99.9 \\
        & cxcywh & JSON & bounding\_box & 93.3 & 89.6 & 100.0 & 82.8 & 79.4 & 100.0 & 93.8 & 90.5 & 100.0 & \textbf{90.0} & 86.5 & 100.0 \\
    \midrule
    \multirow{2}{*}{Opus 5.5} & yxhw & Text & -- & 81.6 & 76.0 & 99.9 & 70.3 & 66.0 & 99.6 & 60.4 & 57.4 & 95.0 & \textbf{70.8} & 66.4 & 98.2 \\
        & yxhw & JSON & bounding\_box & 39.0 & 35.9 & 99.9 & 26.5 & 25.2 & 100.0 & 31.1 & 29.8 & 100.0 & \underline{32.2} & 30.3 & 100.0 \\
    \bottomrule
    \end{tabular}
    }
    \caption{Instance detection results on HR-InsDet (easy and hard) and RoboTools. The Average columns aggregate across the three datasets. *Despite all other runs being run with bbox JSON key for Gemini, for instance detection Gemini only outputs box\_2d key no matter what it is prompted for.}
    \label{tab:insdet}
    \end{table*}

\begin{table*}[h]
    \centering
    \resizebox{1.0\textwidth}{!}{
    \setlength{\tabcolsep}{3pt}
    \begin{tabular}{lccc ccc ccc ccc ccc | ccc}
    \toprule
     & & & & \multicolumn{6}{c}{RoboTools} & \multicolumn{6}{c}{iGround} & \multicolumn{3}{|c}{\multirow{2}{*}{Average}} \\
    \cmidrule(lr){5-10} \cmidrule(lr){11-16}
     & & & & \multicolumn{3}{c}{2 frames} & \multicolumn{3}{c}{8 frames} & \multicolumn{3}{c}{2 frames} & \multicolumn{3}{c}{8 frames} & & & \\
    \cmidrule(lr){5-7} \cmidrule(lr){8-10} \cmidrule(lr){11-13} \cmidrule(lr){14-16}
    Model & Bbox & Format & JSON key & F1@0.5 & mIoU & FA (\%) & F1@0.5 & mIoU & FA (\%) & F1@0.5 & mIoU & FA (\%) & F1@0.5 & mIoU & FA (\%) & F1@0.5 & mIoU & FA (\%) \\
    
    \rowcolor{tablebanner} \multicolumn{19}{l}{\textit{Open Source Models}} \\
    \multirow{2}{*}{Qwen2.5VL} & \multirow{2}{*}{xyxy} & Text & -- &  22.1 &  22.6 &  100.0 & 0.9  &  3.3  &  56.5 & 14.2 & 8.0 & 35.9 & 1.9 & 0.9 & 15.5 & \underline{9.8} & 8.7 & 52.0 \\
     &  & JSON & bbox\_2d & 9.3  &  5.6  &  14.3  & 2.4  &  6.0  &  99.4 & 31.9 & 33.5 & 94.5 & 24.4 & 25.1 & 74.5 & \textbf{17.0} & 17.5 & 70.7 \\
    \midrule
    \multirow{2}{*}{Qwen3VL} & \multirow{2}{*}{xyxy} & Text & -- &  53.7 &  50.7 &  99.4 & 16.0 &  18.7 &  97.5 & 48.5 & 46.5 & 94.8 & 38.3 & 37.8 & 95.7 & \underline{39.1} & 38.5 & 96.8 \\
     &  & JSON & bbox\_2d &  59.9 &  54.2 &  100.0  & 18.5 &  21.7 &  91.9  & 47.3 & 45.7 & 94.7 & 39.0 & 38.8 & 95.6 & \textbf{41.1} & 40.1 & 95.6 \\
    \midrule
    \multirow{2}{*}{Qwen3.5} & \multirow{2}{*}{xyxy} & Text & -- & 1.8 & 1.2 & 12.4 & 3.0 & 5.1 & 13.7 & 16.2 & 9.1 & 22.0 & 16.7 & 9.7 & 25.4 & \textbf{9.4} & 6.3 & 18.4 \\
     &  & JSON & bbox\_2d & 0.0 & 0.0 & 0.0 & 0.0 & 0.0 & 0.0 & 22.3 & 14.5 & 31.7 & 2.4 & 1.2 & 3.4 & \underline{6.2} & 3.9 & 8.8 \\
    \midrule
    \multirow{2}{*}{\makecell[l]{Qwen3.5 \\ - thinking}} & \multirow{2}{*}{xyxy} & Text & -- & 5.4 & 3.6 & 11.8 & 1.4 & 1.1 & 11.2 & 54.6 & 45.7 & 89.2 & 46.7 & 38.5 & 88.4 & \underline{27.0} & 22.2 & 50.1 \\
     &  & JSON & bbox\_2d & 16.0 & 17.0 & 66.5 & 5.3 & 8.7 & 82.0 & 51.5 & 43.2 & 83.8 & 48.2 & 40.0 & 88.7 & \textbf{30.2} & 27.2 & 80.2 \\
    \midrule
    \multirow{2}{*}{InternVL3} & \multirow{2}{*}{xyxy} & Text & -- & 0.3 & 2.6 & 100.0 & 0.0 & 2.2 & 4.3 & 44.5 & 45.1 & 99.7 & 27.1 & 29.9 & 99.3 & \textbf{18.0} & 20.0 & 75.8 \\
        &  & JSON & coordinates & 0.0 & 2.2 & 100.0 & 0.0 & 3.3 & 37.3 & 37.9 & 39.7 & 98.3 & 24.1 & 27.1 & 99.2 & \underline{15.5} & 18.1 & 83.7 \\
    \midrule
    \multirow{2}{*}{Gemma-4} & \multirow{2}{*}{yxyx} & Text & -- & 0.0 & 2.1 & 10.6 & 0.3 & 2.5 & 98.1 & 37.5 & 36.9 & 88.8 & 27.3 & 27.3 & 81.6 & \underline{16.3} & 17.2 & 69.8 \\
     &  & JSON & bounding\_box & 0.0 & 3.0 & 42.9 & 0.7 & 2.9 & 73.9 & 39.0 & 40.0 & 86.5 & 28.9 & 28.7 & 85.3 & \textbf{17.2} & 18.6 & 72.1 \\
    \midrule
    \multirow{2}{*}{GLM4.6V} & \multirow{2}{*}{xyxy} & Text & -- & 69.9 & 61.2 & 66.5 & 35.4 & 32.5 & 82.6 & 53.7 & 49.4 & 90.1 & 44.7 & 39.2 & 88.1 & \textbf{50.9} & 45.6 & 81.8 \\
        &  & JSON & coordinates & 9.6 & 4.6 & 4.3 & 6.7 & 3.4 & 6.2 & 3.2 & 1.5 & 1.8 & 3.1 & 1.4 & 3.5 & \underline{5.6} & 2.8 & 4.0 \\
    \rowcolor{tablebanner} \multicolumn{19}{l}{\textit{Closed Source Models}} \\
    \multirow{2}{*}{GPT5.4} & xyxy & Text & -- & 31.7 & 34.4 & 100.0 & 17.8 & 22.8 & 100.0 & 39.7 & 38.4 & 99.3 & 36.8 & 37.0 & 98.9 & \textbf{31.5} & 33.2 & 99.6 \\
        & xywh & JSON & coordinates & 30.4 & 34.6 & 99.4 & 16.3 & 21.8 & 97.5 & 30.1 & 32.5 & 83.3 & 33.7 & 32.4 & 69.6 & \underline{27.6} & 30.3 & 87.4 \\
    \midrule
    \multirow{2}{*}{Gemini 2.5 Flash} & \multirow{2}{*}{yxyx} & Text & -- & 18.8 & 20.8 & 85.7 & 6.5 & 10.9 & 95.0 & 53.5 & 48.3 & 96.0 & 36.9 & 36.9 & 98.9 & \textbf{28.9} & 29.2 & 93.9 \\
        &  & JSON & box\_2d*/bbox & 1.9 & 1.4 & 6.2 & 0.8 & 0.7 & 8.7 & 50.3 & 53.5 & 89.5 & 39.0 & 41.1 & 92.4 & \underline{23.0} & 24.1 & 49.2 \\
    \midrule
    \multirow{2}{*}{Sonnet 4.5} & xyxy & Text & -- & 13.6 & 17.8 & 100.0 & 4.2 & 8.0 & 100.0 & 22.8 & 28.2 & 99.7 & 22.9 & 27.3 & 99.5 & \textbf{15.9} & 20.3 & 99.8 \\
        & xywh & JSON & bbox & 11.1 & 14.4 & 94.4 & 3.6 & 8.0 & 98.1 & 24.1 & 30.4 & 97.1 & 22.8 & 29.4 & 82.6 & \underline{15.4} & 20.6 & 93.1 \\
    \midrule
    \multirow{2}{*}{GPT-6 Astra} & xywh & Text & -- & 95.0 & 91.8 & 100.0 & 94.3 & 92.6 & 100.0 & 62.2 & 63.9 & 99.5 & 66.6 & 64.8 & 99.9 & \underline{79.5} & 78.3 & 99.9 \\
        & cxcywh & JSON & bounding\_box & 96.1 & 91.8 & 93.8 & 98.5 & 92.6 & 77.0 & 60.1 & 60.2 & 91.9 & 68.0 & 62.1 & 77.5 & \textbf{80.7} & 76.7 & 85.1 \\
    \midrule
    \multirow{2}{*}{Opus 5.5} & yxhw & Text & -- & 93.2 & 88.2 & 98.8 & 95.8 & 88.6 & 98.8 & 61.1 & 59.0 & 97.9 & 63.1 & 58.4 & 99.9 & \textbf{78.3} & 73.6 & 98.8 \\
        & yxhw & JSON & bounding\_box & 90.5 & 85.7 & 95.0 & 94.0 & 87.7 & 95.7 & 56.6 & 54.7 & 97.9 & 60.4 & 57.2 & 99.1 & \underline{75.4} & 71.3 & 96.9 \\
    \bottomrule
    \end{tabular}
    }
    \caption{Video detection results on RoboTools and iGround at 2 and 8 uniformly sampled frames. The Average column aggregates across both datasets and both frame counts. *For instance detection Gemini only outputs box\_2d key.}
    \label{tab:video} 
    \end{table*}

\textbf{Stage 3: per-task selection.}
Stage 3 runs the Stage 1 text winner, the Stage 2 JSON winner, and the unconstrained prompt for each model on every task family's full datasets.
The unconstrained row appears only for models that emit a parsable shape under the unconstrained prompt (\Cref{tab:lmms_stage1_pascal}).
For each task family, the configuration with the highest average F1@0.5 across the family's datasets is selected.
The per-task tables (\cref{tab:objdet_multi,tab:objdet_single,tab:refexp,tab:insdet,tab:video}) report all configurations per model, with the winner in bold and the runner-up underlined.

\subsection{Validating the evaluation protocol}
\label{app:robustness}

The main tables report each model at the configuration selected by the multi-stage format search (\Cref{app:sandbox}).
This protocol rests on four choices: the format search runs on Pascal only, each model is reported at its best format, each format is tested with a single prompt wording, and F1 is computed at an IoU threshold of 0.5.
This section tests each of these choices.
\Cref{app:transfer} repeats the format search on a referring-expression and an instance-detection dataset.
\Cref{app:crossformat} compares the performance using the best bounding-box representation with the average performance over all bounding-box representations.
\Cref{app:prompt_robustness} paraphrases the prompt, and \Cref{app:f1avg} raises the IoU threshold.

\subsubsection{Transfer of the format search beyond Pascal}
\label{app:transfer}

\begin{table*}[t]
    \centering
    \resizebox{\textwidth}{!}{
    \setlength{\tabcolsep}{4pt}
    \begin{tabular}{l ccc ccc ccc ccc ccc}
    \toprule
     & \multicolumn{3}{c}{GLM4.6V, Text} & \multicolumn{3}{c}{Qwen3VL, Text} & \multicolumn{3}{c}{Qwen3VL, JSON} & \multicolumn{3}{c}{GPT5.4, Text} & \multicolumn{3}{c}{GPT5.4, JSON} \\
    \cmidrule(lr){2-4} \cmidrule(lr){5-7} \cmidrule(lr){8-10} \cmidrule(lr){11-13} \cmidrule(lr){14-16}
    Representation & Pascal & RefL4 & RoboTools & Pascal & RefL4 & RoboTools & Pascal & RefL4 & RoboTools & Pascal & RefL4 & RoboTools & Pascal & RefL4 & RoboTools \\
    \midrule
    \emph{--} & 79.4 & 81.6 & 64.4 & 67.9 & 82.0 & 93.3 & 75.5 & 80.0 & 87.5 & 36.0 & 37.8 & 24.2 & 0.0 & 0.0 & 0.0 \\
    \midrule
    \texttt{xyxy} & \textbf{78.8} & \textbf{82.8} & \textbf{76.3} & \textbf{69.9} & \textbf{80.0} & \textbf{83.3} & \textbf{74.6} & \textbf{78.0} & \textbf{86.6} & 42.1 & \textbf{65.0} & 45.0 & 37.6 & \textbf{53.7} & 35.0 \\
    \texttt{xywh} & 16.7 & 20.0 & 0.0 & 19.5 & 18.0 & 0.0 & 17.2 & 18.0 & 0.0 & 35.6 & 55.0 & \textbf{50.0} & \textbf{40.4} & 45.0 & 43.9 \\
    \texttt{yxyx} & 14.7 & 16.5 & 2.1 & 13.6 & 10.0 & 2.2 & 12.2 & 10.0 & 2.1 & 34.0 & 55.0 & 40.0 & 37.5 & 39.0 & \textbf{58.5} \\
    \texttt{yxhw} & 7.4 & 6.0 & 0.0 & 8.5 & 8.0 & 0.0 & 6.7 & 6.0 & 0.0 & 32.4 & 45.0 & 45.0 & 32.3 & 43.9 & 39.0 \\
    \texttt{all} & 0.0 & 0.0 & 0.0 & 0.0 & 0.0 & 2.6 & 0.0 & 0.0 & 0.0 & 38.8 & 35.0 & 25.0 & 11.4 & 5.6 & 25.6 \\
    \texttt{all-labelled} & 0.0 & 0.0 & 0.0 & 0.0 & 0.0 & 0.0 & 0.0 & 0.0 & 0.0 & \textbf{43.3} & 41.0 & 30.8 & 34.3 & 43.9 & 24.4 \\
    \texttt{cxcywh} & 0.0 & 0.0 & 0.0 & 0.0 & 0.0 & 0.0 & 0.0 & 0.0 & 0.0 & 28.9 & 50.0 & 20.0 & 38.8 & 34.1 & 29.3 \\
    \midrule
    Avg.\ F1 & 16.8 & 17.9 & 11.2 & 15.9 & 16.6 & 12.6 & 15.8 & 16.0 & 12.7 & 36.4 & 49.4 & 36.5 & 33.2 & 37.9 & 36.5 \\
    Avg.\ FA (\%) & 70.9 & 70.3 & 70.0 & 73.1 & 71.7 & 90.9 & 69.1 & 71.4 & 71.4 & 100.0 & 99.3 & 99.3 & 95.7 & 97.1 & 98.6 \\
    \bottomrule
    \end{tabular}
    }
    \caption{Stage~1 sweep repeated on RefL4 (referring expressions) and single-frame RoboTools (instance detection), reported as F1@0.5.
    The Pascal columns are taken from \Cref{tab:lmms_stage1_pascal}.
    Each cell uses 50 queries for GLM4.6V and Qwen3VL and 20 for GPT5.4, and the JSON key is fixed to \texttt{bbox\_2d}.
    Bold marks the best bounding-box representation per column.
    The unconstrained prompt (\texttt{--}) is shown for reference and is not part of the Stage~1 selection.
    Avg.\ F1 and Avg.\ FA average over the seven representations (\Cref{app:crossformat}).}
    \label{tab:transfer}
\end{table*}

We run the format search on Pascal VOC because it is a widely used detection dataset with common object categories.
We therefore expect it to present the easiest conditions for every model, so that score differences between formats reflect the format rather than the difficulty of the task.
To test whether the representation selected on Pascal transfers to other tasks, we repeat the full Stage~1 sweep on RefL4 (referring expressions) and on single-frame RoboTools (instance detection) for GLM-4.6V, Qwen3-VL, and GPT-5.4.
Each cell uses the same number of queries as on Pascal (50 for the open-source models and 20 for GPT-5.4), and the JSON key is fixed to \texttt{bbox\_2d}.

For GLM-4.6V and Qwen3-VL, the representation selected on Pascal (\texttt{xyxy}) is the best representation on every dataset and in every tested output format (\Cref{tab:transfer}).
The remaining representations also behave as on Pascal: \texttt{xywh}, \texttt{yxyx}, and \texttt{yxhw} stay at or below 20.0 F1, and \texttt{all}, \texttt{all-labelled}, and \texttt{cxcywh} stay at or below 2.6 F1 on all three datasets.
In contrast, GPT-5.4 does not specialize to one preferred representation, as already observed on Pascal.
Its F1 varies much less across representations than for the open-source models, and the best-performing representation differs between datasets.
Note that on RefL4 and RoboTools, each query has a single target, so with 20 queries per cell, a single query already changes F1 by about 5 points, and differences of 10 to 15 F1 correspond to only two to three queries.
This indicates that the representation selected on Pascal transfers to other tasks for models that specialize to one format.

\subsubsection{Performance across representations}
\label{app:crossformat}

\begin{table}[t]
    \centering
    \resizebox{\textwidth}{!}{
    \setlength{\tabcolsep}{5pt}
    \begin{tabular}{llc ccc ccc}
    \toprule
     & & & \multicolumn{3}{c}{Single-label} & \multicolumn{3}{c}{Multi-label} \\
    \cmidrule(lr){4-6} \cmidrule(lr){7-9}
    Model & Output & Selected & Sel.\ F1 & Avg.\ F1 & Avg.\ FA (\%) & Sel.\ F1 & Avg.\ F1 & Avg.\ FA (\%) \\
    \midrule
    \rowcolor{tablebanner} \multicolumn{9}{l}{\textit{Open Source Models}} \\
    \multirow{2}{*}{Qwen2.5VL} & Text & \texttt{xyxy} & 40.6 & 10.6 & 34.9 & 20.8 & 6.5 & 68.0 \\
                               & JSON & \texttt{xyxy} & 64.3 & 14.4 & 58.6 & 66.1 & 15.1 & 61.7 \\
    \multirow{2}{*}{Gemma-4} & Text & \texttt{yxyx} & 73.0 & 15.7 & 80.6 & 71.8 & 19.2 & 71.4 \\
                               & JSON & \texttt{yxyx} & 43.4 & 8.6 & 65.1 & 17.0 & 3.9 & 11.7 \\
    \multirow{2}{*}{GLM4.6V} & Text & \texttt{xyxy} & 78.8 & 16.8 & 70.9 & 80.3 & 17.4 & 65.7 \\
                               & JSON & \texttt{xyxy} & 44.1 & 7.1 & 14.3 & 63.7 & 12.6 & 31.4 \\
    \multirow{2}{*}{Qwen3.5} & Text & \texttt{xyxy} & 64.8 & 14.0 & 71.7 & 25.3 & 8.2 & 43.1 \\
                               & JSON & \texttt{xyxy} & 77.7 & 16.4 & 71.4 & 80.0 & 17.7 & 70.9 \\
    \multirow{2}{*}{Qwen3VL} & Text & \texttt{xyxy} & 69.9 & 15.9 & 73.1 & 68.8 & 18.1 & 71.4 \\
                               & JSON & \texttt{xyxy} & 74.6 & 15.8 & 69.1 & 74.1 & 16.7 & 69.1 \\
    \multirow{2}{*}{InternVL3} & Text & \texttt{xyxy} & 79.8 & 16.2 & 69.7 & 77.0 & 18.6 & 71.4 \\
                               & JSON & \texttt{xyxy} & 80.2 & 16.7 & 73.7 & 82.5 & 19.3 & 69.4 \\
    \multirow{2}{*}{Qwen3.5-th.} & Text & \texttt{xyxy} & 77.7 & 17.5 & 68.0 & 78.0 & 18.6 & 66.0 \\
                               & JSON & \texttt{xyxy} & 76.7 & 17.0 & 65.4 & 75.0 & 17.8 & 64.0 \\
    \rowcolor{tablebanner} \multicolumn{9}{l}{\textit{Closed Source Models}} \\
    \multirow{2}{*}{Sonnet 4.5} & Text & \texttt{xyxy} & 30.8 & 13.3 & 82.9 & 31.3 & 22.1 & 85.7 \\
                               & JSON & \texttt{xywh} & 27.4 & 19.2 & 95.7 & 31.7 & 28.6 & 92.9 \\
    \multirow{2}{*}{Gemini 2.5 Flash} & Text & \texttt{yxyx} & 37.7 & 19.4 & 97.1 & 49.6 & 38.6 & 99.3 \\
                               & JSON & \texttt{yxyx} & 76.1 & 15.0 & 57.9 & 45.9 & 19.1 & 50.0 \\
    \multirow{2}{*}{GPT5.4} & Text & \texttt{xyxy} & 42.1 & 36.4 & 100.0 & 54.5 & 53.4 & 99.3 \\
                               & JSON & \texttt{xywh} & 40.4 & 33.2 & 95.7 & 41.2 & 39.0 & 97.1 \\
    \multirow{2}{*}{GPT-6 Astra} & Text & \texttt{xywh} & 76.8 & 71.9 & 98.6 & 82.7 & \textbf{77.3} & 100.0 \\
                               & JSON & \texttt{cxcywh} & 76.5 & 70.3 & 97.9 & 80.8 & 76.2 & 99.3 \\
    \multirow{2}{*}{Opus 5.5} & Text & \texttt{yxhw} & 82.1 & 75.2 & 100.0 & 87.1 & 72.7 & 100.0 \\
                               & JSON & \texttt{yxhw} & 83.2 & \textbf{80.7} & 100.0 & 88.0 & 73.9 & 100.0 \\
    \bottomrule
    \end{tabular}
    }
    \caption{Selected versus average performance across bounding-box representations on the Pascal Stage~1 subset.
    \emph{Sel.\ F1} is the F1@0.5 at the representation selected in Stage~1 (\Cref{tab:preferred_format}).
    \emph{Avg.\ F1} and \emph{Avg.\ FA} average over the seven bounding-box representations of \Cref{subsec:bbox}, excluding the unconstrained prompt.
    All values are computed from \Cref{tab:lmms_stage1_pascal}.
    Bold marks the highest average F1.}
    \label{tab:crossformat}
\end{table}

Reporting each model using its best bounding-box representation measures an upper bound: a user who requests a representation other than the model's preferred one obtains a lower score.
The main tables report this upper bound because the benchmark aims to measure localization separately from instruction following (to output the requested representation).
To quantify this gap, \Cref{tab:crossformat} lists, for every model and output format, the F1 at the representation selected in Stage~1 next to the mean F1 and mean format adherence over all seven representations of \Cref{subsec:bbox}.
The unconstrained prompt is excluded from the average because it does not request a representation.
For GLM-4.6V, Qwen3-VL, and GPT-5.4, the last two rows of \Cref{tab:transfer} report the same averages on RefL4 and RoboTools.

\Cref{tab:crossformat} shows that for all open-source models and Gemini 2.5 Flash, the cross-representation average lies significantly below the selected representation.
This drop is not caused by unparsable outputs alone: when prompted for another four-coordinate representation, these models still return parsable boxes in most cases, but their F1 collapses (\Cref{tab:lmms_stage1_pascal}), indicating that they likely silently keep the coordinate convention they were trained on (\Cref{app:cxcywh}).
In contrast, the GPT and Claude models lose at most 17.5 F1.
GPT-5.4 keeps this advantage over GLM-4.6V and Qwen3-VL on RefL4 and RoboTools (\Cref{tab:transfer}).

Gemini 2.5 Flash, whose documentation specifies a preferred representation, specializes to it like the open-source models.

\subsubsection{Robustness to prompt wording}
\label{app:prompt_robustness}

\begin{table}[t]
    \centering
    \resizebox{\textwidth}{!}{
    \setlength{\tabcolsep}{5pt}
    \begin{tabular}{ll ccccc c cc}
    \toprule
     & & \multicolumn{6}{c}{F1@0.5} & & \\
    \cmidrule(lr){3-8}
    Model & Output & p1 & p2 & p3 & p4 & p5 & Mean $\pm$ std & mIoU & FA (\%) \\
    \midrule
    \multirow{2}{*}{GLM4.6V} & Text$^{\dagger}$ & 78.8 & 79.6 & 81.4 & 81.3 & 80.8 & \textbf{80.4 $\pm$ 1.0} & 68.0 $\pm$ 1.6 & 98.0 $\pm$ 1.3 \\
                             & JSON & 51.0 & 33.1 & 48.2 & 48.3 & 48.6 & 45.8 $\pm$ 6.5 & 28.9 $\pm$ 5.0 & 32.8 $\pm$ 6.1 \\
    \midrule
    \multirow{2}{*}{Qwen3VL} & Text & 68.3 & 53.1 & 58.7 & 48.1 & 62.5 & 58.2 $\pm$ 7.1 & 62.2 $\pm$ 10.1 & 100.0 $\pm$ 0.0 \\
                             & JSON$^{\dagger}$ & 76.7 & 74.6 & 78.0 & 75.4 & 76.3 & \textbf{76.2 $\pm$ 1.2} & 66.8 $\pm$ 1.1 & 97.6 $\pm$ 0.8 \\
    \bottomrule
    \end{tabular}
    }
    \caption{Robustness to prompt wording on the Pascal Stage~1 subset (50 queries, single-label, \texttt{xyxy}, JSON key \texttt{bbox\_2d}).
    p1 is the prompt used throughout the paper and p2--p5 are paraphrases (\Cref{fig:prompts_paraphrase}).
    Means and population standard deviations are taken over the five prompts.
    $^{\dagger}$ marks each model's preferred output format.}
    \label{tab:prompt_robustness}
\end{table}

As shown in \Cref{sec:results}, small changes in the requested output format change the score of most models substantially.
To test whether the wording of the instruction has a similar effect, we paraphrase the single-label object detection prompt four times while keeping the label, the coordinate syntax, and the JSON schema fixed (\Cref{fig:prompts_paraphrase}).
The first variant, p1, is the prompt used throughout the paper.
We evaluate GLM-4.6V and Qwen3-VL on the Pascal Stage~1 subset (50 queries) with \texttt{xyxy} in both output formats.

Both models are robust to the wording in their preferred output format (text for GLM-4.6V, JSON for Qwen3-VL), with a standard deviation of at most 1.2 F1 across the five prompts (\Cref{tab:prompt_robustness}).
Outside the preferred format, the standard deviation rises to 6.5 (GLM-4.6V JSON) and 7.1 F1 (Qwen3-VL text).
Since the format search selects each model's preferred format, a single prompt wording suffices for the main tables.

\subsubsection{Stricter IoU thresholds}
\label{app:f1avg}

\begin{table}[t]
    \centering
    \small
    \setlength{\tabcolsep}{5pt}
    \begin{tabular}{l ccccc c c}
    \toprule
    Model & F1@0.5 & F1@0.6 & F1@0.7 & F1@0.8 & F1@0.9 & F1@avg & Retention (\%) \\
    \midrule
    \rowcolor{tablebanner} \multicolumn{8}{l}{\textit{Open Source Models}} \\
    Gemma-4     & 68.7 & 62.2 & 51.9 & 39.7 & 22.0 & 48.9 & 71 \\
    Qwen2.5VL   & 70.1 & 67.4 & 62.0 & 52.0 & 35.6 & 57.4 & \textbf{82} \\
    Qwen3.5-th. & 75.3 & 71.5 & 64.2 & 51.6 & 31.8 & 58.9 & 78 \\
    Qwen3.5     & 77.3 & 73.1 & 64.6 & 51.2 & 31.8 & 59.6 & 77 \\
    InternVL3   & 77.0 & 72.7 & 67.0 & 55.3 & 37.1 & 61.8 & 80 \\
    Qwen3VL     & 77.5 & 73.8 & 67.5 & 55.3 & 38.1 & 62.4 & 81 \\
    GLM4.6V     & 80.5 & 77.2 & 71.5 & 61.1 & 40.8 & 66.2 & \textbf{82} \\
    \rowcolor{tablebanner} \multicolumn{8}{l}{\textit{Closed Source Models}} \\
    Sonnet 4.5  & 27.3 & 16.7 & 8.2 & 3.3 & 0.7 & 11.3 & 41 \\
    GPT5.4      & 48.8 & 37.0 & 25.1 & 13.6 & 3.6 & 25.6 & 52 \\
    Gemini 2.5 Flash & 74.5 & 69.5 & 61.3 & 48.3 & 26.7 & 56.0 & 75 \\
    GPT-6 Astra & 81.4 & 77.0 & 69.8 & 58.1 & 39.0 & 65.1 & 80 \\
    Opus 5.5    & \textbf{86.3} & \textbf{82.8} & \textbf{76.3} & \textbf{64.3} & \textbf{43.3} & \textbf{70.6} & \textbf{82} \\
    \bottomrule
    \end{tabular}
    \caption{F1 at increasing IoU thresholds on Pascal single-label (1,000 queries), at the configuration of each model in \Cref{tab:objdet_single_sota}.
    F1@avg is the mean over the five thresholds, and retention is F1@avg\,/\,F1@0.5.}
    \label{tab:f1_thresholds}
\end{table}

\begin{table*}[t]
    \centering
    \resizebox{\textwidth}{!}{
    \setlength{\tabcolsep}{4pt}
    \begin{tabular}{l ccc ccc cccccc}
    \toprule
     & \multicolumn{3}{c}{Single-label Obj.\ Det.} & \multicolumn{3}{c}{Multi-label Obj.\ Det.} & \multicolumn{6}{c}{Referring Expressions} \\
    \cmidrule(lr){2-4} \cmidrule(lr){5-7} \cmidrule(lr){8-13}
    Model & Pascal & OpenImages & iGround & Pascal & OpenImages & iGround & RefCOCO-Avg & RefL4 & Flickr30k & D3 & PhraseCut & SVG \\
    \midrule
    \rowcolor{tablebanner} \multicolumn{13}{l}{\textit{Open Source Models}} \\
    Gemma-4     & 48.9 & 25.3 & 41.6 & 55.3 & 30.2 & 46.4 & 59.0 & 60.7 & 40.6 & 37.8 & 24.7 & 41.4 \\
    InternVL3   & 61.8 & 26.1 & 38.6 & 67.4 & 26.3 & 41.0 & 64.0 & 51.7 & 47.3 & 38.9 & 24.1 & 57.8 \\
    Qwen2.5VL   & 57.4 & 39.1 & 25.3 & 64.9 & 33.5 & 22.0 & 73.8 & 68.7 & 45.3 & 45.2 & 34.5 & 56.6 \\
    Qwen3.5-th. & 58.9 & 39.3 & 44.4 & 61.9 & 19.7 & 43.3 & 71.6 & 63.4 & 47.1 & 40.7 & 30.5 & 53.2 \\
    Qwen3.5     & 59.6 & 40.0 & 40.0 & 63.1 & 34.8 & 40.8 & 71.9 & 63.4 & 47.3 & 30.8 & 28.6 & 54.5 \\
    Qwen3VL     & 62.4 & \textbf{51.0} & 45.7 & 65.6 & \textbf{42.4} & 43.9 & 76.1 & 75.2 & 45.2 & 41.6 & 34.6 & 59.6 \\
    GLM4.6V     & 66.2 & 45.3 & 47.0 & 68.9 & 37.5 & 47.4 & 74.9 & 75.7 & \textbf{61.3} & 35.4 & 36.8 & 61.2 \\
    \rowcolor{tablebanner} \multicolumn{13}{l}{\textit{Closed Source Models}} \\
    Sonnet 4.5  & 11.3 & 12.8 & 10.0 & 13.5 & 12.7 & 11.4 & 12.8 & 10.8 & 10.1 & 12.2 & 7.9 & 8.2 \\
    GPT5.4      & 25.6 & 25.2 & 17.8 & 29.0 & 23.7 & 19.1 & 29.9 & 28.2 & 19.5 & 24.7 & 13.3 & 17.0 \\
    Gemini 2.5 Flash & 56.0 & 37.3 & 42.1 & 57.7 & 33.3 & 46.7 & 61.8 & 58.8 & 41.8 & 40.5 & 24.9 & 50.2 \\
    GPT-6 Astra & 65.1 & 36.1 & 45.6 & 68.1 & 35.6 & 53.6 & \textbf{83.7} & \textbf{83.3} & 47.4 & 48.9 & 32.7 & 62.6 \\
    Opus 5.5    & \textbf{70.6} & 44.8 & \textbf{50.9} & \textbf{70.6} & 38.9 & \textbf{54.1} & 81.4 & 81.2 & 49.7 & \textbf{56.2} & \textbf{38.5} & \textbf{68.2} \\
    \bottomrule
    \end{tabular}
    }
    \caption{F1@avg (mean of F1 at IoU thresholds 0.5 to 0.9) for each model at the configuration used in \Cref{tab:objdet_single_sota,tab:objdet_multi_sota,tab:refex_sota_main}.
    RefCOCO-Avg averages the six RefCOCO/+/g splits.
    Bold marks the best model per column.}
    \label{tab:f1avg}
\end{table*}

F1@0.5 counts a prediction as correct once its IoU with the matched ground-truth box reaches 0.5, so it does not distinguish roughly placed boxes from precise boxes.
Detection benchmarks address this with AP averaged over IoU thresholds, which requires a confidence score per box that MLLMs do not produce.
We instead compute F1 at IoU thresholds from 0.5 to 0.9 in steps of 0.1 and report their mean as F1@avg.

Overall, it shows that stricter thresholds reinforce two findings of the main paper.
First, the gap between the leading open-source models and GPT-5.4 widens on all twelve object detection and referring-expression datasets (\Cref{tab:f1avg}).
On Pascal single-label (\Cref{tab:f1_thresholds}), the retention (F1@avg\,/\,F1@0.5) is 71 to 82\% for the open-source models, Gemini 2.5 Flash, and the frontier models, but only 52\% for GPT-5.4 and 41\% for Sonnet 4.5, whose boxes overlap the target but are imprecise (3.6 and 0.7 F1 at an IoU of 0.9).
Second, GLM-4.6V, the best open-source model across tasks (\Cref{fig:cross-task}), produces the most precise boxes among the open-source models: it reaches the highest open-source F1@avg on eight of the twelve datasets, compared to four under F1@0.5. 
Across all models, Opus 5.5 reaches the highest F1@avg on seven of the twelve datasets.
We keep F1@0.5 in the main tables because it is the standard threshold in detection and referring-expression evaluation~\citep{pascal,refcoco_unc}, and because it measures whether a model finds the target separately from how tightly its box fits.

\subsection{The cxcywh representation}
\label{app:cxcywh}

We probe the center-format failure in two ways: prompting with explicit definitions of the four numbers, and reinterpreting a \texttt{cxcywh}-prompted response as corner coordinates.

\textbf{Definition prompt.}
The \texttt{cxcywh + definition} variant appends a one-line description of \texttt{cx}, \texttt{cy}, \texttt{bw}, and \texttt{bh} to the prompt.
The intervention helps three models substantially in text mode: Qwen3.5-Thinking lifts F1 from 0 to 20.2 single-label, GPT-5.4 from 28.9 to 41.2 single-label, and Opus 5.5 from 46.0 to 83.2 multi-label (\Cref{tab:lmms_stage1_pascal}).
Gemma-4 improves modestly (F1 from 5.6 to 10.3 in text, single-label).
The remaining open-source models stay at or near F1 0 with or without definitions.

\begin{table*}[ht]
    \centering
    \setlength{\tabcolsep}{3pt}
    \resizebox{\textwidth}{!}{
    \footnotesize
    \begin{tabular}{l cccc c cccc c}
    \toprule
    \multirow{3}{*}{Model} & \multicolumn{5}{c}{Text} & \multicolumn{5}{c}{JSON} \\
    \cmidrule(lr){2-6} \cmidrule(lr){7-11}
     & \multicolumn{2}{c}{\texttt{cxcywh}} & \multicolumn{2}{c}{corner} & \multirow{2}{*}{\makecell{FA\\(\%)}} & \multicolumn{2}{c}{\texttt{cxcywh}} & \multicolumn{2}{c}{corner} & \multirow{2}{*}{\makecell{FA\\(\%)}} \\
    \cmidrule(lr){2-3} \cmidrule(lr){4-5} \cmidrule(lr){7-8} \cmidrule(lr){9-10}
     & mIoU & F1@0.5 & mIoU & F1@0.5 & & mIoU & F1@0.5 & mIoU & F1@0.5 & \\
    \midrule
    Qwen2.5VL & 3.5 & 0.0 & 17.9 & \textbf{32.0} & 46.0 & 5.7 & 0.0 & 32.2 & \textbf{49.0} & 78.0 \\
    Qwen3VL & 8.5 & 0.0 & 49.4 & \textbf{69.5} & 100.0 & 10.6 & 0.0 & 65.2 & \textbf{75.4} & 96.0 \\
    Qwen3.5 & 11.5 & 0.0 & 62.7 & \textbf{57.8} & 100.0 & 11.3 & 0.0 & 71.8 & \textbf{78.4} & 100.0 \\
{Qwen3.5-th.} & 9.8 & 0.0 & 55.8 & \textbf{74.0} & 92.0 & 9.7 & 0.0 & 54.8 & \textbf{73.7} & 92.0 \\
    InternVL3 & 10.3 & 0.0 & 58.2 & \textbf{73.6} & 100.0 & 12.3 & 0.0 & 69.1 & \textbf{79.0} & 100.0 \\
    Gemma-4$^*$ & 11.7 & 5.6 & 33.4 & \textbf{40.0} & 100.0 & 4.6 & 0.0 & 27.5 & \textbf{44.8} & 98.0 \\
    GLM4.6V & 10.7 & 0.0 & 65.9 & \textbf{78.1} & 98.0 & 3.0 & 0.0 & 23.9 & \textbf{40.9} & 22.0 \\
    \bottomrule
    \end{tabular}
    }
    \caption{\textbf{Reevaluating \texttt{cxcywh} outputs as corner-format on Pascal.}
    The corner parse uses \texttt{xyxy} for every model except Gemma-4$^*$, which uses \texttt{yxyx}.}
    \label{tab:cxcywh}
    \end{table*}

\textbf{Corner reinterpretation.}
In~\cref{tab:cxcywh}, the same \texttt{cxcywh}-prompted response is reparsed as \texttt{xyxy} (or \texttt{yxyx} for Gemma-4) to test whether the model is silently returning corner coordinates regardless of the prompt.
Under the center parser, F1 is 0.0 for every cell except Gemma-4 text (F1 5.6), and mIoU is at most 12.3 (InternVL3 JSON).
The corner reparse lifts F1 to between 32.0 (Qwen2.5-VL text) and 79.0 (InternVL3 JSON), and mIoU to between 17.9 (Qwen2.5-VL text) and 71.8 (Qwen3.5 JSON).

\begin{table*}[ht]
\centering
\resizebox{1.0\textwidth}{!}{
\setlength{\tabcolsep}{3pt}
\begin{tabular}{lccc ccc ccc ccc ccc}
\toprule
 & & & & \multicolumn{3}{c}{iGround} & \multicolumn{3}{c}{iGround {\small\textit{+ captions}}} & \multicolumn{3}{c}{Flickr30k-Entities} & \multicolumn{3}{c}{Flickr30k-Entities {\small\textit{+ captions}}} \\
\cmidrule(lr){5-7} \cmidrule(lr){8-10} \cmidrule(lr){11-13} \cmidrule(lr){14-16}
Model & Bbox & Format & JSON key & F1@0.5 & mIoU & FA (\%) & F1@0.5 & mIoU & FA (\%) & F1@0.5 & mIoU & FA (\%) & F1@0.5 & mIoU & FA (\%) \\
\midrule
\multirow{2}{*}{Qwen2.5VL} & \multirow{2}{*}{xyxy} & Text & -- & 41.5 & 39.7 & 95.1 & 40.8 & 40.8 & 98.8 & 53.3 & 39.2 & 92.6 & 58.7 & 43.0 & 96.6 \\
 & & JSON & bbox\_2d & 35.3 & 32.8 & 78.4 & 35.4 & 33.1 & 79.5 & 50.5 & 39.6 & 96.0 & 54.5 & 41.9 & 94.7 \\
\cmidrule(lr){1-16}
\multirow{2}{*}{Qwen3VL} & \multirow{2}{*}{xyxy} & Text & -- & 58.9 & 55.6 & 94.0 & 61.5 & 57.7 & 91.0 & 53.2 & 41.0 & 100.0 & 56.2 & 43.0 & 100.0 \\
 & & JSON & bbox\_2d & 58.6 & 57.9 & 88.5 & 63.7 & 59.9 & 85.0 & 57.5 & 45.0 & 100.0 & 60.7 & 48.5 & 100.0 \\
\cmidrule(lr){1-16}
\multirow{2}{*}{Qwen3.5} & \multirow{2}{*}{xyxy} & Text & -- & 57.9 & 55.8 & 95.8 & 59.8 & 57.8 & 99.0 & 58.1 & 42.7 & 99.5 & 62.0 & 46.4 & 100.0 \\
 & & JSON & bbox\_2d & 55.3 & 59.3 & 99.5 & 57.3 & 60.4 & 99.7 & 62.0 & 47.8 & 99.9 & 67.0 & 53.2 & 99.6 \\
\cmidrule(lr){1-16}
\multirow{2}{*}{\makecell[l]{Qwen3.5 \\ - thinking}} & \multirow{2}{*}{xyxy} & Text & -- & 59.6 & 59.2 & 85.4 & 60.9 & 59.5 & 88.5 & 63.1 & 51.8 & 96.9 & 66.3 & 54.2 & 97.0 \\
 & & JSON & bbox\_2d & 62.5 & 59.5 & 81.8 & 62.0 & 59.9 & 84.4 & 63.8 & 52.5 & 93.6 & 65.9 & 52.9 & 89.9 \\
\cmidrule(lr){1-16}
\multirow{2}{*}{InternVL3} & \multirow{2}{*}{xyxy} & Text & -- & 53.1 & 53.5 & 96.7 & 52.5 & 53.4 & 98.2 & 61.6 & 51.1 & 97.9 & 64.9 & 52.5 & 96.6 \\
    &  & JSON & coordinates & 52.8 & 54.9 & 99.8 & 51.6 & 53.8 & 100.0 & 61.7 & 58.8 & 98.8 & 65.8 & 60.9 & 99.0 \\
\cmidrule(lr){1-16}
\multirow{2}{*}{Gemma-4} & \multirow{2}{*}{yxyx} & Text & -- & 53.5 & 42.6 & 92.1 & 56.9 & 53.2 & 99.8 & 55.2 & 50.6 & 99.5 & 60.1 & 55.8 & 100.0 \\
 & & JSON & bounding\_box & 55.2 & 42.2 & 99.9 & 60.0 & 55.6 & 100.0 & 57.9 & 49.9 & 99.7 & 59.6 & 50.0 & 99.3 \\
\cmidrule(lr){1-16}
\multirow{2}{*}{GLM4.6V} & \multirow{2}{*}{xyxy} & Text & -- & 60.3 & 61.1 & 99.5 & 61.5 & 61.2 & 99.5 & 74.5 & 64.6 & 99.5 & 77.1 & 67.1 & 99.4 \\
    &  & JSON & coordinates & 14.2 & 9.0 & 14.6 & 11.7 & 6.5 & 11.3 & 34.3 & 19.9 & 17.2 & 23.8 & 12.6 & 12.5 \\
\bottomrule
\end{tabular}
}
\caption{\textbf{Effect of appending the original caption to the prompt.}
}
\label{tab:captions}
\end{table*}

\subsection{Adapting grounded-captioning datasets to detection}
\label{app:captions}

iGround and Flickr30k Entities are originally grounded-captioning datasets.
Each ground-truth box is annotated with respect to a specific caption, and other instances of the same object class in the image may go unannotated.
When we adapt these datasets to object detection by treating each grounded noun as a class label, this introduces a potential ground-truth mismatch.
A model that correctly localizes every instance of the class can be penalized for predictions on instances the caption does not mention.
To check whether this systematically biases our results, we re-run the same models with the original caption appended to the prompt.
\Cref{tab:captions} reports the comparison.
The delta is small for most models (typically below 5 mIoU), suggesting the ground-truth mismatch is not a dominant source of error and that these datasets are usable as detection benchmarks without modification.
The exception is Gemma-4 on iGround, whose mIoU rises by more than 10 points with the caption, while its F1 rises by less than 5.
We use the no-caption variant in the main results.

\begin{table*}[ht]
    \centering
    \resizebox{1.0\textwidth}{!}{
    \setlength{\tabcolsep}{3pt}
    \begin{tabular}{lccc ccccccccc cccccc ccc}
    \toprule
     & & & & \multicolumn{9}{c}{RefCOCO} & \multicolumn{6}{c}{RefCOCO+} & \multicolumn{3}{c}{RefCOCO-g} \\
    \cmidrule(lr){5-13} \cmidrule(lr){14-19} \cmidrule(lr){20-22}
     & & & & \multicolumn{3}{c}{test} & \multicolumn{3}{c}{testA} & \multicolumn{3}{c}{testB} & \multicolumn{3}{c}{testA} & \multicolumn{3}{c}{testB} & \multicolumn{3}{c}{test} \\
    \cmidrule(lr){5-7} \cmidrule(lr){8-10} \cmidrule(lr){11-13} \cmidrule(lr){14-16} \cmidrule(lr){17-19} \cmidrule(lr){20-22}
    Model & Bbox & Format & JSON key & F1@0.5 & mIoU & FA (\%) & F1@0.5 & mIoU & FA (\%) & F1@0.5 & mIoU & FA (\%) & F1@0.5 & mIoU & FA (\%) & F1@0.5 & mIoU & FA (\%) & F1@0.5 & mIoU & FA (\%) \\
    
        \rowcolor{tablebanner} \multicolumn{22}{l}{\textit{Open Source Models}} \\

    \multirow{3}{*}{Qwen2.5VL} & \emph{-- (as xyxy)} & Text \emph{(as JSON)} & \emph{-- (as bbox\_2d)} & 88.6 & 81.1 & 99.9 & 92.0 & 84.9 & 100.0 & 83.6 & 76.1 & 100.0 & 88.0 & 80.9 & 100.0 & 74.5 & 69.7 & 99.9 & 86.1 & 79.4 & 100.0 \\
        & \multirow{2}{*}{xyxy} & Text & -- & 85.3 & 75.5 & 92.9 & 87.5 & 75.9 & 90.0 & 80.3 & 70.0 & 92.1 & 82.6 & 71.3 & 87.9 & 70.6 & 61.7 & 89.7 & 85.2 & 77.1 & 96.5 \\
     & & JSON & bbox\_2d & 77.0 & 71.5 & 91.3 & 80.4 & 72.6 & 87.7 & 72.3 & 67.3 & 91.9 & 75.6 & 69.4 & 92.2 & 60.5 & 59.1 & 93.5 & 79.2 & 73.9 & 97.9 \\
    \cmidrule(lr){1-22}
    \multirow{3}{*}{Qwen3VL} & \emph{-- (as xyxy)} & Text \emph{(as JSON)} & \emph{-- (as bbox\_2d)} & 90.9 & 82.5 & 99.9 & 91.0 & 83.8 & 100.0 & 85.8 & 77.9 & 99.8 & 87.7 & 80.3 & 99.9 & 80.1 & 74.1 & 99.8 & 89.3 & 81.7 & 100.0 \\
        & \multirow{2}{*}{xyxy} & Text & -- & 91.0 & 82.8 & 100.0 & 91.2 & 83.6 & 100.0 & 85.2 & 77.9 & 100.0 & 87.4 & 80.1 & 100.0 & 79.0 & 73.0 & 100.0 & 90.1 & 81.9 & 100.0 \\
     & & JSON & bbox\_2d & 90.0 & 81.9 & 99.9 & 91.4 & 84.0 & 100.0 & 85.2 & 77.7 & 99.9 & 87.5 & 80.2 & 99.9 & 79.4 & 73.8 & 99.9 & 90.0 & 82.1 & 99.9 \\
    \cmidrule(lr){1-22}
    \multirow{3}{*}{Qwen3.5} & \emph{-- (as xyxy)} & Text \emph{(as JSON)} & \emph{-- (as bbox\_2d)} & 89.7 & 79.4 & 100.0 & 91.1 & 82.0 & 100.0 & 84.9 & 75.3 & 100.0 & 87.5 & 78.6 & 100.0 & 79.7 & 71.2 & 100.0 & 88.1 & 78.1 & 100.0 \\
        & \multirow{2}{*}{xyxy} & Text & -- & 89.1 & 79.1 & 100.0 & 90.9 & 81.7 & 100.0 & 85.4 & 75.6 & 100.0 & 87.7 & 78.3 & 100.0 & 79.7 & 71.2 & 100.0 & 88.7 & 78.1 & 100.0 \\
     & & JSON & bbox\_2d & 89.4 & 79.5 & 100.0 & 90.9 & 82.0 & 100.0 & 84.2 & 75.5 & 100.0 & 86.6 & 78.2 & 99.9 & 79.7 & 71.6 & 100.0 & 88.6 & 78.9 & 100.0 \\
    \cmidrule(lr){1-22}
    \multirow{3}{*}{\makecell[l]{Qwen3.5 \\ - thinking}} & \emph{-- (as xyxy)} & Text \emph{(as JSON)} & \emph{-- (as bbox\_2d)} & 89.7 & 79.2 & 99.0 & 91.3 & 81.8 & 99.4 & 84.3 & 74.7 & 98.6 & 87.4 & 78.0 & 99.5 & 80.1 & 71.8 & 98.3 & 87.8 & 77.5 & 99.3 \\
        & \multirow{2}{*}{xyxy} & Text & -- & 89.8 & 79.4 & 99.7 & 92.4 & 82.2 & 99.5 & 84.8 & 74.9 & 98.7 & 88.9 & 79.3 & 99.8 & 80.0 & 71.8 & 98.6 & 87.3 & 77.0 & 99.3 \\
     & & JSON & bbox\_2d & 87.6 & 77.1 & 96.5 & 91.0 & 80.9 & 98.4 & 82.2 & 70.8 & 93.2 & 85.9 & 76.7 & 98.0 & 77.7 & 68.4 & 94.0 & 84.4 & 72.9 & 93.3 \\
    \cmidrule(lr){1-22}
    \multirow{2}{*}{InternVL3} & \multirow{2}{*}{xyxy} & Text & -- & 80.9 & 73.3 & 100.0 & 85.8 & 77.1 & 99.9 & 74.7 & 67.1 & 99.8 & 79.9 & 71.7 & 100.0 & 66.7 & 61.0 & 100.0 & 80.6 & 72.9 & 100.0 \\
        &  & JSON & coordinates & 67.6 & 78.4 & 99.9 & 69.7 & 77.8 & 99.9 & 61.2 & 74.4 & 100.0 & 66.4 & 72.6 & 99.9 & 57.5 & 74.0 & 100.0 & 68.9 & 77.8 & 100.0 \\
    \cmidrule(lr){1-22}
    \multirow{2}{*}{Gemma-4} & \multirow{2}{*}{yxyx} & Text & -- & 76.0 & 68.6 & 99.6 & 75.8 & 68.7 & 99.7 & 74.1 & 67.8 & 99.9 & 67.2 & 60.6 & 99.7 & 59.9 & 55.7 & 99.8 & 77.9 & 69.5 & 100.0 \\
     & & JSON & bounding\_box & 76.1 & 70.2 & 99.8 & 75.1 & 68.7 & 99.1 & 73.1 & 67.1 & 99.9 & 67.6 & 62.5 & 99.8 & 57.3 & 55.6 & 100.0 & 78.0 & 71.2 & 100.0 \\
    \cmidrule(lr){1-22}
    \multirow{2}{*}{GLM4.6V} & \multirow{2}{*}{xyxy} & Text & -- & 88.2 & 81.2 & 99.0 & 87.6 & 81.9 & 99.1 & 84.7 & 77.9 & 98.9 & 81.2 & 75.5 & 99.3 & 76.3 & 71.5 & 99.4 & 87.8 & 81.7 & 98.6 \\
        &  & JSON & coordinates & 2.6 & 1.2 & 1.3 & 2.8 & 1.3 & 1.7 & 1.8 & 0.9 & 1.1 & 3.1 & 1.5 & 2.4 & 2.5 & 1.2 & 1.5 & 2.0 & 0.9 & 1.0 \\
    
    \rowcolor{tablebanner} \multicolumn{22}{l}{\textit{Closed Source Models}} \\
    \multirow{2}{*}{GPT5.4} & xyxy & Text & -- & 65.8 & 54.7 & 100.0 & 64.6 & 54.4 & 100.0 & 64.0 & 53.5 & 100.0 & 58.2 & 50.3 & 100.0 & 56.7 & 47.8 & 100.0 & 65.5 & 55.6 & 100.0 \\
        & xywh & JSON & coordinates & 59.3 & 51.8 & 100.0 & 58.8 & 52.7 & 99.9 & 58.5 & 51.9 & 100.0 & 51.9 & 48.6 & 100.0 & 48.5 & 46.4 & 100.0 & 61.3 & 54.7 & 100.0 \\
    \cmidrule(lr){1-22}
    \multirow{2}{*}{Gemini 2.5 Flash} & \multirow{2}{*}{yxyx} & Text & -- & 78.7 & 69.0 & 99.0 & 77.2 & 68.1 & 99.6 & 76.7 & 68.2 & 99.6 & 68.6 & 61.7 & 99.7 & 68.0 & 61.7 & 99.2 & 78.9 & 68.3 & 99.5 \\
        &  & JSON & bbox & 78.9 & 73.0 & 97.7 & 79.2 & 73.1 & 98.5 & 76.0 & 71.4 & 97.6 & 67.0 & 65.4 & 97.5 & 66.2 & 65.5 & 96.1 & 80.7 & 75.8 & 98.5 \\
    \cmidrule(lr){1-22}
    \multirow{2}{*}{Sonnet 4.5} & xyxy & Text & -- & 32.3 & 37.4 & 96.7 & 31.1 & 36.4 & 95.2 & 30.5 & 35.2 & 95.7 & 25.1 & 33.2 & 97.0 & 29.6 & 32.9 & 96.2 & 27.8 & 34.8 & 97.4 \\
        & xywh & JSON & bbox & 33.1 & 39.3 & 99.3 & 31.3 & 38.6 & 99.4 & 34.8 & 40.0 & 99.5 & 26.1 & 34.8 & 99.3 & 30.6 & 34.7 & 99.0 & 31.9 & 37.7 & 99.9 \\
    \cmidrule(lr){1-22}
    \multirow{2}{*}{GPT-6 Astra} & xywh & Text & -- & 94.3 & 87.1 & 100.0 & 94.8 & 87.8 & 100.0 & 92.0 & 84.6 & 100.0 & 91.4 & 85.3 & 100.0 & 89.0 & 82.1 & 100.0 & 92.7 & 85.9 & 100.0 \\
        & cxcywh & JSON & bounding\_box & 77.8 & 72.4 & 100.0 & 74.8 & 69.7 & 100.0 & 78.2 & 73.2 & 100.0 & 72.8 & 68.4 & 100.0 & 76.4 & 72.0 & 100.0 & 78.4 & 73.2 & 99.9 \\
    \cmidrule(lr){1-22}
    \multirow{2}{*}{Opus 5.5} & yxhw & Text & -- & 91.8 & 85.4 & 100.0 & 91.7 & 85.9 & 100.0 & 91.6 & 83.8 & 100.0 & 86.7 & 81.7 & 99.9 & 87.8 & 81.4 & 100.0 & 91.3 & 85.2 & 100.0 \\
        & yxhw & JSON & bounding\_box & 84.9 & 80.6 & 100.0 & 85.0 & 80.9 & 99.8 & 83.3 & 78.6 & 100.0 & 78.4 & 75.4 & 99.8 & 78.5 & 77.1 & 100.0 & 83.5 & 79.1 & 100.0 \\
    \bottomrule
    \end{tabular}
    }
    \caption{Referring expression results on all splits of RefCOCO, RefCOCO+, and RefCOCO-g.}
    \label{tab:refcoco}
    \end{table*}

\subsection{Per-split RefCOCO/+/g results}
\label{app:refcoco}

\Cref{tab:refex_sota_main,tab:refexp} aggregate RefCOCO, RefCOCO+, and RefCOCO-g into a single \emph{RefCOCO-Avg} column.
\Cref{tab:refcoco} reports the per-split numbers behind this average.
\emph{RefCOCO} and \emph{RefCOCO+} are collected with the ReferItGame two-player protocol, in which one player describes a target object and the other identifies it from the image.
\emph{RefCOCO+} disallows location words at collection time, so players describe the target through other attributes.
\emph{RefCOCO-g} uses non-interactive Mechanical Turk annotation, with longer descriptions on average.
RefCOCO reports three splits (\texttt{test}, \texttt{testA}, \texttt{testB}), RefCOCO+ reports two (\texttt{testA}, \texttt{testB}), and RefCOCO-g reports a single \texttt{test} split.
On RefCOCO and RefCOCO+, \texttt{testA} and \texttt{testB} come from the UNC split~\citep{refcoco_unc}, with \texttt{testA} containing images with multiple people and \texttt{testB} containing images with multiple non-person instances.
RefCOCO's \texttt{test} comes from the Google split~\citep{refcocog}.
For open-source models other than GLM-4.6V, the split ranking is consistent: \texttt{testA} scores higher than \texttt{testB} on RefCOCO and RefCOCO+, and RefCOCO+ scores below RefCOCO at the same split label.
RefCOCO+ \texttt{testB} is the hardest split, with F1 between 57.3 (Gemma-4 JSON) and 80.1 (Qwen3.5-Thinking and Qwen3-VL, unconstrained).
RefCOCO-g \texttt{test} stays within 3.2 F1 of RefCOCO \texttt{test} on every open-source model.

\subsection{Additional Implementation Details}
\label{app:resizing}
\paragraph{HR-InsDet image preprocessing.}
For \emph{HR-InsDet}, the scenes are typically around 8192$\times$6144.
This creates a problem for the proprietary models because the images must pass through OpenRouter's 30 MB request-size limit and Anthropic's 5 MB image-upload limit.
We therefore resize each scene to a maximum side length of 4096 before encoding, so the typical 8192$\times$6144 image becomes 4096$\times$3072, and we send it as a JPEG at quality 95.
The shared proprietary input for GPT, Claude, and Gemini is therefore a 4096$\times$3072 JPEG rather than the original image.

However, each provider applies its own preprocessing, which is not exposed through the API, so the proprietary models see different effective resolutions.
By default, GPT-5.4 fits images within 2048 pixels per side and 2{,}500 patches of 32$\times$32 pixels, which reduces the scene to 1824$\times$1368.
GPT-6 Astra processes images at native resolution, which makes a full-size query about four times as expensive.
For budget reasons, we send it the scene at 1824$\times$1368 and the reference image at 1600$\times$1600, the sizes GPT-5.4 processes.
Anthropic limits images to 1568 pixels per side and 1568 visual tokens of 28$\times$28 pixels for Sonnet 4.5, and to 2576 pixels per side and 4784 visual tokens for Opus 5.5.
This reduces the scene to 1270$\times$952 and 2212$\times$1659, respectively, and the 1920$\times$1080 \emph{RoboTools} frames to 1456$\times$819 for Sonnet 4.5.
Gemini 2.5 Flash uses tiled 768$\times$768 processing with an effective resolution comparable to GPT-5.4.
For GPT and Claude, we map the predicted pixel coordinates from the resized image back to the original resolution.

For the open-source models, we match the effective pixel budget where the processors allow it.
For the Qwen family, we set \texttt{max\_pixels=12,845,056}, which reduces a typical 8192$\times$6144 \textit{HR-InsDet} image to roughly 4116$\times$3080.
InternVL preprocesses images as 448$\times$448 tiles, so we set \texttt{max\_patches=64}, which gives the same budget because $64 \times 448^2 = 12{,}845{,}056$.
GLM-4.6V's processor caps each image at about 4.7\,MP and reduces an 8192$\times$6144 scene to 2520$\times$1876, below this budget.
Gemma-4 instead uses a fixed budget of at most 2520 patches of 16$\times$16 pixels (about 0.65\,MP), pooled into 280 visual tokens, so it operates at substantially lower effective resolution.
Knowing the effective image size is important for Qwen2.5-VL, GPT, and Claude because these models output coordinates directly in pixel space, so the predicted boxes must be remapped to the original ground-truth resolution before evaluation.

\begin{table}[t]
    \centering
    \small
    \begin{minipage}[t]{0.48\linewidth}
    \vspace{0pt}\centering
    \begin{tabular}{lcc}
    \toprule
    Scene longest side & F1@0.5 & mIoU \\
    \midrule
    native (4.7\,MP) & 64.1 & 54.5 \\
    3072 (4.7\,MP) & 62.9 & 53.7 \\
    2048 (3.1\,MP) & 60.9 & 52.2 \\
    1024 (0.8\,MP) & 44.4 & 37.1 \\
    \bottomrule
    \end{tabular}
    \end{minipage}\hfill
    \begin{minipage}[t]{0.48\linewidth}
    \vspace{0pt}\centering
    \begin{tabular}{lcc}
    \toprule
    Subset & 1824\,px & 4096\,px \\
    \midrule
    HR-InsDet \textit{easy} & 94.2 & 99.0 \\
    HR-InsDet \textit{hard} & 83.3 & 88.8 \\
    \bottomrule
    \end{tabular}
    \end{minipage}
    \caption{Effect of input resolution on instance detection.
    Left: GLM-4.6V on HR-InsDet \textit{easy} (1,000 queries), with the scene downscaled to the given longest side before encoding; native is the setting of the main tables. 
    Right: F1@0.5 of GPT-6 Astra on the first 100 queries of each subset, at the resolution of the main tables (scene 1824$\times$1368, reference image 1600$\times$1600) and at full resolution (scene 4096$\times$3072, reference image 3072$\times$3072), both in the configuration of \Cref{tab:insdet_sota}.}
    \label{tab:resolution}
\end{table}

\paragraph{Effect of input resolution.}
Since the effective resolution differs between models, we measure how strongly it affects instance detection (\Cref{tab:resolution}).
For GLM-4.6V, we downscale the scene on \emph{HR-InsDet easy}; for GPT-6 Astra, we send the full-resolution scene and reference image on the first 100 queries of each subset.
Between about 2000 and 4000 pixels on the longest side, resolution changes F1 by only a few points: GLM-4.6V loses 3.2 F1 at 2048 pixels, and GPT-6 Astra gains about 5 F1 at 4096 pixels.
Only stronger downscaling hurts, with GLM-4.6V losing 19.7 F1 at 1024 pixels.
Resolution therefore does not explain the low instance detection scores of GPT-5.4 and Sonnet 4.5: even at 1024 pixels, GLM-4.6V clearly outperforms both, although they process the scene at 1824 and 1270 pixels.

\paragraph{Per-model evaluation quirks.}
\label{app:quirks}
Gemini 2.5 Flash returns coordinates only under the \texttt{box\_2d} key on instance detection, regardless of the requested key.
On RoboTools video instance detection in the JSON format, Gemini 2.5 Flash mostly omits the requested \texttt{frame\_idx} and returns a plain list of boxes (125 of 161 queries with 2 frames, 122 with 8 frames).
\Cref{tab:video} reports these answers as parsed by our protocol, which cannot assign them to frames.
Because each RoboTools query contains a single instance, the $k$-th box can instead be assigned to the $k$-th frame.
This raises F1 from 1.9 to 18.8 and format adherence from 6.2\% to 83.9\% with 2 frames, but F1 only from 0.8 to 2.9 with 8 frames, where most of these answers contain a single box.
The text format therefore remains better for Gemini 2.5 Flash on average (28.9 vs.\ 27.8 F1).

In JSON mode, GLM-4.6V emits multiple \texttt{<|begin\_of\_box|>...<|end\_of\_box|>} blocks where the first contains a plain coordinate list and the rest contain JSON-shaped versions.
Reading only the first block (\Cref{sec:implementation}) hands the JSON parser a plain list, which fails to parse.
Format adherence therefore collapses on every GLM-4.6V JSON cell except multi-label object detection.

Opus 5.5 sometimes returns coordinates on the normalized grid or as 0--1 fractions instead of in pixels, mostly on instance detection.
The response does not state its coordinate space, and these values are also valid pixel coordinates, so we read all answers as pixels.
Reading each answer in the coordinate space in which it matches the most ground-truth boxes would raise its average instance detection F1 from 70.8 to 86.0.
Requesting the JSON key \texttt{box\_2d}, which Gemini documents together with the normalized grid, also changes its coordinate space, so its F1 collapses with this key (\Cref{tab:stage_2_pascal}).
GPT-6 Astra returns some referring-expression answers as 0--1 fractions, but only in JSON mode.
Several models leave out frames although the prompt asks for every frame; as the prompt does not fix the numbering, answers that name neither the first nor the last frame are read as 1-based.


\subsection{Prompts}
\label{app:prompts}

The prompts that \benchmark{} sends to the model are listed verbatim below.
Placeholders in curly braces (\texttt{\{label\}}, \texttt{\{num\_frames\}}, \texttt{\{caption\}}) are filled per query at evaluation time.
Placeholders in angle brackets (\texttt{<bbox>} and \texttt{<key>}) stand for the bounding-box representation and JSON key being evaluated, with values listed in \Cref{subsec:bbox} and \Cref{subsec:format}.
Placeholders inside the JSON schema (\texttt{<label>} and \texttt{<frame\_idx>}) mark where the model writes each instance's class name or frame index.
Single-label object detection is shown across all the variants the prompt can take (\Cref{fig:prompts_demo}).
The remaining tasks are shown in JSON mode with the bounding-box representation specified, in \Cref{fig:prompts_image_other} for image tasks and \Cref{fig:prompts_video} for video tasks.

\begin{figure}[t]
\begin{lstlisting}[title={Text mode, unconstrained representation.}]
Locate all instances of '{label}' in the image. Return only the bounding box coordinates and nothing else.
\end{lstlisting}

\begin{lstlisting}[title={Text mode, structure given.}]
Locate all instances of '{label}' in the image. Return only the bounding box coordinates in the format [<bbox>] and nothing else.
\end{lstlisting}

\begin{lstlisting}[title={JSON mode, unconstrained representation.}]
Locate all instances of '{label}' in the image. Return only the bounding box coordinates in a JSON format and nothing else.
\end{lstlisting}

\begin{lstlisting}[title={JSON mode, structure given.}]
Locate all instances of '{label}' in the image. Return the bounding box coordinates in a JSON format: [{"<key>": [<bbox>]}, ...]
Repeat the "<key>" key for each instance. Do not include any other text or comments.
\end{lstlisting}

\begin{lstlisting}[title={JSON mode, structure given, center-format definition appended.}]
Locate all instances of '{label}' in the image. Return the bounding box coordinates in a JSON format: [{"<key>": [<bbox>]}, ...]
Repeat the "<key>" key for each instance. Do not include any other text or comments. Use the following definitions: cx and cy are the coordinates of the box center; bw is the box width; bh is the box height.
\end{lstlisting}

\begin{lstlisting}[title={JSON mode, structure given, image caption prepended.}]
This image shows: "{caption}". Locate all instances of '{label}' in the image. Return the bounding box coordinates in a JSON format: [{"<key>": [<bbox>]}, ...]
Repeat the "<key>" key for each instance. Do not include any other text or comments.
\end{lstlisting}

\caption{Single-label object detection prompts under each variant \benchmark{} evaluates. }
\label{fig:prompts_demo}
\end{figure}

\begin{figure}[t]
\begin{lstlisting}[title={Multi-label object detection.}]
Locate all instances of the following labels in the image: {label}. Return the bounding box coordinates in a JSON format: [{"label": "<label>", "<key>": [<bbox>]}, ...]
Repeat the object for each instance, with its class as "label" and its coordinates under "<key>". Do not include any other text or comments.
\end{lstlisting}

\begin{lstlisting}[title={Referring expression, single matching instance.}]
Locate the object that matches the description '{label}' in the image. Return the bounding box coordinates in a JSON format: [{"<key>": [<bbox>]}, ...]
Repeat the "<key>" key for each instance. Do not include any other text or comments.
\end{lstlisting}

\begin{lstlisting}[title={Referring expression, all matching instances.}]
Locate every object that matches the description '{label}' in the image. Return the bounding box coordinates in a JSON format: [{"<key>": [<bbox>]}, ...]
Repeat the "<key>" key for each instance. Do not include any other text or comments.
\end{lstlisting}

\begin{lstlisting}[title={Referring expression, all matching instances on datasets that include negative queries.}]
Locate every object that matches the description '{label}' in the image. Return the bounding box coordinates in a JSON format: [{"<key>": [<bbox>]}, ...]
Repeat the "<key>" key for each instance. Do not include any other text or comments. If no matching objects are found, return [].
\end{lstlisting}

\begin{lstlisting}[title={Instance detection.}]
Here are reference images of an object:
<image 1>
Locate all instances of this object in:
<image 2>

Return the bounding box coordinates in a JSON format: [{"<key>": [<bbox>]}, ...]
Repeat the "<key>" key for each instance. Do not include any other text or comments.
\end{lstlisting}

\caption{Image-task prompts other than single-label object detection, in JSON mode with the bounding-box representation specified.}
\label{fig:prompts_image_other}
\end{figure}

\begin{figure}[t]
\begin{lstlisting}[title={Instance detection across video frames.}]
Here are reference images of an object:
<image 1>
Locate all instances of this object in:
<image 2>
<image 3>
...

Return the bounding box coordinates for each of the {num_frames} frames in a JSON format: [{"frame_idx": <frame_idx>, "<key>": [<bbox>]}, ...]
Repeat the "<key>" key for each instance. Do not include any other text or comments.
\end{lstlisting}

\begin{lstlisting}[title={Object detection across video frames.}]
Locate all instances of '{label}' in the video frames. Return the bounding box coordinates for each of the {num_frames} frames in a JSON format: [{"frame_idx": <frame_idx>, "<key>": [<bbox>]}, ...]
Repeat the "<key>" key for each instance. Do not include any other text or comments.
\end{lstlisting}

\caption{Video-task prompts, in JSON mode with the bounding-box representation specified.}
\label{fig:prompts_video}
\end{figure}

\begin{figure}[t]
\begin{lstlisting}[title={Text mode.}]
p1: Locate all instances of '{label}' in the image. Return only the bounding box coordinates in the format [x_min, y_min, x_max, y_max] and nothing else.
p2: Detect every instance of '{label}' in the image. Output only the bounding box coordinates, formatted as [x_min, y_min, x_max, y_max].
p3: Find all objects of the class '{label}' in the image and report each one's bounding box in the format [x_min, y_min, x_max, y_max]. Do not include any other text.
p4: Please provide the bounding box coordinates, given as [x_min, y_min, x_max, y_max], for every '{label}' appearing in the image. Return the coordinates only.
p5: Where is every '{label}' in the image? Respond with only bounding box coordinates in the format [x_min, y_min, x_max, y_max].
\end{lstlisting}

\begin{lstlisting}[title={JSON mode.}]
p1: Locate all instances of '{label}' in the image. Return the bounding box coordinates in a JSON format: [{"bbox_2d": [x_min, y_min, x_max, y_max]}, ...]
    Repeat the "bbox_2d" key for each instance. Do not include any other text or comments.
p2: Detect every instance of '{label}' in the image. Return the results as JSON: [{"bbox_2d": [x_min, y_min, x_max, y_max]}, ...]
    Use one "bbox_2d" entry per instance. Output nothing besides the JSON.
p3: Find all objects of the class '{label}' in the image and report each one's bounding box as JSON in the form [{"bbox_2d": [x_min, y_min, x_max, y_max]}, ...]
    Add one object with the "bbox_2d" key for every instance. Do not include any other text.
p4: Please provide, in JSON format [{"bbox_2d": [x_min, y_min, x_max, y_max]}, ...], a bounding box for every '{label}' appearing in the image, repeating the "bbox_2d" key for each instance. Return the JSON only.
p5: Where is every '{label}' in the image? Respond with a JSON list [{"bbox_2d": [x_min, y_min, x_max, y_max]}, ...] containing one "bbox_2d" entry per instance, and no other text or comments.
\end{lstlisting}

\caption{Prompt paraphrases for single-label object detection used in \Cref{app:prompt_robustness}. p1 is the prompt used throughout the paper. The label, the coordinate syntax, and the JSON schema are the same in all variants.}
\label{fig:prompts_paraphrase}
\end{figure}

\end{document}